\ifdefined\pdfoutput\pdfoutput=1\fi
\documentclass[10pt,journal,compsoc]{IEEEtran}

\usepackage[utf8]{inputenc}
\usepackage[T1]{fontenc}

\usepackage[nocompress,sort]{cite}

\usepackage{graphicx}
\graphicspath{{figures/}}

\usepackage{amsmath}
\usepackage{amssymb}
\usepackage{amsfonts}
\usepackage{amsthm}
\usepackage{newpxtext}
\newif\ifstixitalic
\IfFileExists{t1stix2.fd}{\stixitalictrue}{\PackageWarningNoLine{main}{t1stix2.fd not found; keeping Palatino italics}}
\ifstixitalic
  \begingroup\fontfamily{\rmdefault}\selectfont\endgroup
  \DeclareFontShape{T1}{\rmdefault}{m}{it}{<->ssub * stix2/m/it}{}
  \DeclareFontShape{T1}{\rmdefault}{b}{it}{<->ssub * stix2/b/it}{}
  \DeclareFontShape{T1}{\rmdefault}{bx}{it}{<->ssub * stix2/b/it}{}
\fi
\usepackage{nicefrac}
\usepackage{booktabs}
\usepackage{array}
\usepackage{tabularx}
\usepackage{multirow}
\usepackage{makecell}
\usepackage{longtable}
\usepackage[table]{xcolor}
\usepackage{colortbl}
\usepackage[tableposition=top]{caption}
\usepackage{subcaption}
\usepackage[figuresright]{rotating}
\usepackage[export]{adjustbox}

\usepackage{enumitem}
\usepackage{wrapfig}
\usepackage{ragged2e}
\usepackage{microtype}
\usepackage{url}

\usepackage{fontawesome5}
\usepackage{pifont}

\usepackage{algorithm}
\usepackage{algpseudocode}

\usepackage{tikz}
\usetikzlibrary{arrows.meta,positioning,backgrounds,calc,shapes,fit,decorations.pathreplacing,shadows}
\usepackage[edges]{forest}

\definecolor{linkColor}{RGB}{237,4,140}
\definecolor{citecolor}{HTML}{0071bc}
\usepackage[
  pagebackref,
  breaklinks,
  colorlinks,
  linkcolor=linkColor,
  urlcolor=linkColor,
  citecolor=citecolor,
  bookmarksopen=true,
  bookmarksnumbered=true,
]{hyperref}
\usepackage[capitalize]{cleveref}

\usepackage[most]{tcolorbox}

\usepackage{cuted}

\definecolor{themeblue}{RGB}{44,92,168}
\definecolor{themelight}{RGB}{238,244,252}
\definecolor{themedark}{RGB}{32,52,84}
\definecolor{headergray}{RGB}{243,245,247}
\definecolor{rowlight}{RGB}{250,250,252}
\definecolor{tocgray}{RGB}{95,95,95}
\definecolor{themegreen}{RGB}{80,204,72}
\definecolor{accentred}{RGB}{196,72,72}
\definecolor{themeaccent}{RGB}{220,150,40}

\definecolor{cardgold}{HTML}{B08B57}
\definecolor{cardblue}{HTML}{6C8EBF}
\definecolor{cardteal}{RGB}{73,168,100}
\definecolor{cardpurple}{RGB}{156,45,177}
\definecolor{cardcoral}{HTML}{C26B5C}

\definecolor{themecoral}{RGB}{205,92,76}
\definecolor{themegray}{RGB}{120,120,128}
\definecolor{themeslate}{RGB}{72,92,128}
\definecolor{themeteal}{RGB}{32,160,160}
\definecolor{thememagenta}{RGB}{200,72,142}

\definecolor{themehighlight}{HTML}{E8EEF7}

\newcommand{\cYes}{\textcolor{themeblue}{\ding{51}}}
\newcommand{\cNo}{\textcolor{black!35}{\ding{55}}}
\DeclareRobustCommand{\cPart}{\tikz[baseline=-0.5ex]{\fill[themeblue] (0,0) -- (90:0.62ex) arc (90:270:0.62ex) -- cycle; \draw[themeblue, line width=0.45pt] (0,0) circle (0.62ex);}}

\renewcommand{\arraystretch}{1.12}
\newcolumntype{C}[1]{>{\centering\arraybackslash}p{#1}}
\newcolumntype{L}[1]{>{\arraybackslash}p{#1}}

\crefname{section}{Sec.}{Secs.}
\Crefname{section}{Section}{Sections}
\Crefname{table}{Table}{Tables}
\crefname{table}{Table}{Tables}
\Crefname{figure}{Figure}{Figures}
\crefname{figure}{Figure}{Figures}
\Crefname{theorem}{Theorem}{Theorems}
\crefname{theorem}{Thm.}{Thm.s}
\Crefname{algorithm}{Algorithm}{Algorithms}
\crefname{algorithm}{Algo.}{Algos.}
\crefname{appendix}{App.}{Apps.}
\Crefname{appendix}{Appendix}{Appendices}

\let\stOrig\st
\let\st\relax
\usepackage{soul}
\let\st\stOrig
\usepackage{soulpos}
\colorlet{hltintDefault}{themegreen!8!white}
\colorlet{hltintG}{themegreen!8!white}
\colorlet{hltintB}{cardblue!18!white}
\colorlet{hltintP}{cardpurple!10!white}
\colorlet{hltintY}{cardgold!22!white}
\colorlet{hltintR}{themecoral!16!white}
\colorlet{hltintN}{themegray!14!white}
\colorlet{hltintS}{themeslate!14!white}
\colorlet{hltintT}{themeteal!14!white}
\colorlet{hltintM}{thememagenta!12!white}

\newcommand{\pgraph}[1]{\noindent \textbf{#1.\,}}

\tcbset{
  pghlstyle/.style={
    enhanced, on line,
    colback=themegreen!8!white,
    colframe=themegreen!8!white,
    boxrule=0pt,
    arc=2pt,
    left=1pt, right=1pt, top=1pt, bottom=0.pt,
    nobeforeafter,
  }
}
\newtcbox{\pghl}{pghlstyle}
\newcommand{\hltintcur}{hltintDefault}
\ulposdef{\hlround}{%
  \mbox{}%
  \rlap{%
    \tikz[baseline]\fill[\hltintcur, rounded corners=1.5pt]
      (0pt, -0.30\baselineskip)
      rectangle (\ulwidth, 0.78\baselineskip);%
  }%
}
\newcommand{\pghlblock}[2]{%
  \noindent
  {\renewcommand{\hltintcur}{#1}\bfseries\hlround{#2.}}\ \ignorespaces
}

\tcbset{
  pghlGstyle/.style={enhanced, on line,
    colback=themegreen!8!white,    colframe=themegreen!8!white,
    boxrule=0pt, arc=2pt, left=1pt, right=1pt, top=1pt, bottom=0.pt, nobeforeafter},
  pghlBstyle/.style={enhanced, on line,
    colback=cardblue!18!white,     colframe=cardblue!18!white,
    boxrule=0pt, arc=2pt, left=1pt, right=1pt, top=1pt, bottom=0.pt, nobeforeafter},
  pghlPstyle/.style={enhanced, on line,
    colback=cardpurple!10!white,   colframe=cardpurple!10!white,
    boxrule=0pt, arc=2pt, left=1pt, right=1pt, top=1pt, bottom=0.pt, nobeforeafter},
  pghlYstyle/.style={enhanced, on line,
    colback=cardgold!22!white,     colframe=cardgold!22!white,
    boxrule=0pt, arc=2pt, left=1pt, right=1pt, top=1pt, bottom=0.pt, nobeforeafter},
  pghlRstyle/.style={enhanced, on line,
    colback=themecoral!16!white,   colframe=themecoral!16!white,
    boxrule=0pt, arc=2pt, left=1pt, right=1pt, top=1pt, bottom=0.pt, nobeforeafter},
  pghlNstyle/.style={enhanced, on line,
    colback=themegray!14!white,    colframe=themegray!14!white,
    boxrule=0pt, arc=2pt, left=1pt, right=1pt, top=1pt, bottom=0.pt, nobeforeafter},
  pghlSstyle/.style={enhanced, on line,
    colback=themeslate!14!white,   colframe=themeslate!14!white,
    boxrule=0pt, arc=2pt, left=1pt, right=1pt, top=1pt, bottom=0.pt, nobeforeafter},
  pghlTstyle/.style={enhanced, on line,
    colback=themeteal!14!white,    colframe=themeteal!14!white,
    boxrule=0pt, arc=2pt, left=1pt, right=1pt, top=1pt, bottom=0.pt, nobeforeafter},
  pghlMstyle/.style={enhanced, on line,
    colback=thememagenta!12!white, colframe=thememagenta!12!white,
    boxrule=0pt, arc=2pt, left=1pt, right=1pt, top=1pt, bottom=0.pt, nobeforeafter},
}
\newtcbox{\pghlG}{pghlGstyle}
\newtcbox{\pghlB}{pghlBstyle}
\newtcbox{\pghlP}{pghlPstyle}
\newtcbox{\pghlY}{pghlYstyle}
\newtcbox{\pghlR}{pghlRstyle}
\newtcbox{\pghlN}{pghlNstyle}
\newtcbox{\pghlS}{pghlSstyle}
\newtcbox{\pghlT}{pghlTstyle}
\newtcbox{\pghlM}{pghlMstyle}

\newcommand{\pgraphhlB}[1]{\pghlblock{cardblue!30!white}{#1}}
\newcommand{\pgraphhlP}[1]{\pghlblock{hltintP}{#1}}
\newcommand{\pgraphhlY}[1]{\pghlblock{hltintY}{#1}}

\newcommand{\pgraphhlM}[1]{\pghlblock{hltintM}{#1}}

\definecolor{darkgreen}{rgb}{0.0,0.5,0.0}

\definecolor{NiceViolet}{RGB}{156,45,177}
\definecolor{NiceGray}{HTML}{696969}
\definecolor{NiceGreen}{RGB}{73,168,100}

\newcommand{\ie}{\emph{i.e.}}
\newcommand{\eg}{\emph{e.g.}}
\newcommand{\st}{\emph{s.t.}}
\newcommand{\etal}{\emph{et al.}}

\newcommand{\Rolemat}{\boldsymbol{\Gamma}}

\colorlet{defbg}{cardblue!10!white}
\colorlet{deftitlebg}{cardblue}
\newcounter{definition}
\newenvironment{definition}[1][]{%
  \refstepcounter{definition}%
  \def\defoptarg{#1}%
  \begin{tcolorbox}[
    enhanced,
    colback=defbg,
    colbacktitle=deftitlebg,
    colframe=deftitlebg,
    coltitle=white,
    arc=3pt,
    boxrule=0pt,
    fonttitle=\bfseries\small,
    title={Definition~\thedefinition\ifx\defoptarg\empty\else\ (\defoptarg)\fi},
    toptitle=2pt, bottomtitle=2pt,
    left=10pt, right=10pt, top=6pt, bottom=6pt,
    before skip=10pt plus 2pt,
    after skip=10pt plus 2pt,
    before upper={%
      \setlength{\abovedisplayskip}{4pt plus 1pt minus 1pt}%
      \setlength{\belowdisplayskip}{4pt plus 1pt minus 1pt}%
      \setlength{\abovedisplayshortskip}{2pt}%
      \setlength{\belowdisplayshortskip}{2pt}},
  ]%
}{%
  \end{tcolorbox}%
}

\newenvironment{defpair}{%
  \begin{figure*}[!t]%
  \tcbset{before skip=2pt, after skip=2pt}%
}{%
  \end{figure*}%
}
\newcommand{\defpairgap}{\vspace{-10pt}}

\theoremstyle{definition}

\definecolor{rhline}{RGB}{46,84,130}
\definecolor{rhcat}{RGB}{241,236,223}
\definecolor{rhref}{RGB}{223,235,245}
\definecolor{rhtext}{RGB}{30,30,30}
\definecolor{rhcite}{RGB}{18,56,140}

\definecolor{rlcolor}{RGB}{191,87,0}
\newcommand{\rl}[1]{\textup{\textcolor{blue}{\textsc{\MakeLowercase{#1}}}}}
\newcommand{\rlhead}{\text{\makebox[5pt]{\rl{H}}\hspace{6pt}\makebox[5pt]{\rl{B}}\hspace{6pt}\makebox[5pt]{\rl{E}}\hspace{6pt}\makebox[5pt]{\rl{A}}}}
\definecolor{vportcolor}{RGB}{46,125,50}
\definecolor{vcondcolor}{RGB}{196,128,0}
\definecolor{vblockcolor}{RGB}{178,34,34}
\DeclareRobustCommand{\vPort}{\tikz[baseline=-0.55ex]{\fill[vportcolor] (0,0) circle (0.75ex);}}
\DeclareRobustCommand{\vCond}{\tikz[baseline=-0.55ex]{\fill[vcondcolor] (0,0) -- (90:0.75ex) arc (90:270:0.75ex) -- cycle; \draw[vcondcolor, line width=0.5pt] (0,0) circle (0.75ex);}}
\DeclareRobustCommand{\vBlock}{\textcolor{vblockcolor}{\ding{55}}}
\DeclareRobustCommand{\vRed}{\tikz[baseline=-0.55ex]{\draw[black!45, line width=0.5pt] (0,0) circle (0.75ex);}}

\definecolor{titlebg}{RGB}{232,239,248}
\definecolor{titlefg}{RGB}{28,48,82}
\makeatletter
\def\section{\@startsection{section}{1}{\z@}{-3.5ex plus -2ex minus -1.5ex}%
{0.9ex plus 1ex minus 0ex}{\normalfont\fontsize{12.5}{15}\selectfont\bfseries\color{titlefg}}}
\def\subsection{\@startsection{subsection}{2}{\z@}{-3.0ex plus -1.5ex minus -1.5ex}%
{0.6ex plus .5ex minus 0ex}{\normalfont\fontsize{10.8}{13}\selectfont\bfseries\color{titlefg}}}
\def\subsubsection{\@startsection{subsubsection}{3}{\z@}{-2.5ex plus -1ex minus -1ex}%
{0.5ex plus 0.5ex minus 0ex}{\normalfont\normalsize\selectfont\bfseries\itshape\color{titlefg}}}
\makeatother

\newcommand{\paperauthor}{Jingjing Jiang}
\newcommand{\githuburl}{https://github.com/mm-vl/REPO}
\newcommand{\homepageurl}{https://jingjing12110.github.io/SE-MMAgents}
\newcommand{\contactemail}{jingjingjiang2017@gmail.com}

\hypersetup{pdftitle={A Cross-Scenario Survey and Analysis of Self-Evolving Multimodal Agents},pdfauthor={\paperauthor}}

\begin{document}

\twocolumn[{%
\begin{tcolorbox}[enhanced,width=\textwidth,colback=titlebg,colframe=titlebg,boxrule=0pt,arc=4mm,
  left=8mm,right=8mm,top=7mm,bottom=6mm]
\rmfamily
\begin{center}
{\fontsize{21}{26}\selectfont\bfseries\color{titlefg}A Cross-Scenario Survey and Analysis of\\Self-Evolving Multimodal Agents\par}
\vspace{5mm}
{\fontsize{12}{15}\selectfont\bfseries\paperauthor\par}
\end{center}
\vspace{4mm}
\begin{list}{}{\leftmargin=7mm\rightmargin=7mm\topsep=0pt\partopsep=0pt\parsep=0pt\itemsep=0pt}\item\relax
{\fontsize{10}{13.5}\selectfont\justifying
\renewenvironment{abstract}{}{}%
\justify
\begin{abstract}
Recent work shows that multimodal agents increasingly self-improve without recurring human annotation. 
However, each self-improvement mechanism is designed and evaluated within a single scenario, leaving its preconditions unverified under cross-scenario transfer.
This survey investigates the structural source of asymmetric mechanism transfer across four scenarios: multimodal reasoning, multimodal generation, GUI and Web navigation, and embodied manipulation. 
We first formalize the self-evolving multimodal agent by four criteria and a differentia of self-reference, and adopt the three-stage self-evolving loop of experience generation, feedback acquisition, and update consolidation as our analytical framework to position the scenarios along ceilings on rollout count and feedback fidelity. 
Comparing self-evolving with fixed loops within each scenario separates the limitations that the scenario imposes from those inherent in the self-evolving design.
We then recast each mechanism as a trade-off that raises a scarce factor by presupposing a task-structural resource, such as output equivalence or checkable outcomes, and judge it portable, conditional, blocked, or redundant across scenarios. 
We attribute the asymmetry to divergent combinations of scarce factors and resources across opposite transfer directions, and identify the matched-budget measurements that could overturn these still untested verdicts.

\end{abstract}
\par}
\vspace{4mm}
{\small\color{titlefg}\hypersetup{urlcolor=titlefg}
\makebox[1.2em][l]{\faGlobe}\textbf{Homepage:}\ \url{\homepageurl}\par\vspace{0.8mm}
\vspace{0.8mm}
\makebox[1.2em][l]{\faEnvelope}\textbf{Email:}\ \href{mailto:\contactemail}{\texttt{\contactemail}}\par}
\end{list}
\end{tcolorbox}
\vspace{4mm}}]

\section{Introduction}
\label{sec:introduction}

Recent advances in multimodal foundation models have fueled an explosive proliferation of multimodal agents~\cite{durante2024agent, xie2024large, yao2025survey, hu2026guirl}, spawning a flourishing spectrum of self-evolution techniques across multimodal reasoning~\cite{visplay, mmzero, vzero, activezero}, multimodal generation~\cite{solace, suder, internalgap, silmm, corlulm}, GUI and Web navigation~\cite{digirl, acurl,seagent, mobilerl}, and embodied manipulation~\cite{evovla, siefm, searrobot, findcantdo}. 
Despite strong scenario-specific gains, existing self-evolving methods are designed and evaluated within single scenarios, and their mechanisms do not transfer equally well in both directions of a scenario pair. 
For example, majority voting derives label-free rewards for reasoning tasks from consistency across sampled answers~\cite{evolmm,visplay,mmupt}, yet finds no counterpart in embodied manipulation, where action trajectories admit no equivalence check and physical rollouts incur steep execution and reset costs. 
Conversely, world models circumvent expensive environmental interactions via imagined rollouts~\cite{worldvlaloop,merl}, but raise a factor that is not scarce in reasoning, where sampling directly from the policy costs little. 
Beneath this directional breakdown in mechanism transferability, which suggests a \textit{cross-scenario asymmetry}, also lies a shared self-evolving loop across diverse tasks.
Specifically, across all four scenarios, self-evolving agents share the same cycle, \ie, generating exploratory experience, acquiring evaluative feedback, and consolidating persistent updates. 
This survey investigates the structural source of cross-scenario asymmetry by characterizing this shared loop and explaining why the two directions of a scenario pair meet different obstacles to transfer. 

\begin{table*}[!t]
\centering
\scriptsize
\renewcommand{\arraystretch}{1}
\setlength{\tabcolsep}{3.5pt}
\caption{\textbf{Comparison with related surveys.}
\textit{Scenarios} indicates domains covered in dedicated sections: MR (multimodal reasoning), MG (multimodal generation), GW (GUI and Web navigation), and EM (embodied manipulation), or General if scenario-agnostic.
\textit{Loop decomposition} denotes the stages or components formalized by each survey, where EG, FA, and UC represent experience generation, feedback acquisition, and update consolidation, respectively (-- indicates none).
}
\label{tab:survey_comparison}
\vspace{-2mm}
\rowcolors{2}{themelight}{white}
\renewcommand{\tabularxcolumn}[1]{m{#1}}
\begin{tabularx}{\textwidth}{>{\raggedright\arraybackslash}m{3.1cm}>{\raggedright\arraybackslash}m{4.3cm}>{\raggedright\arraybackslash}m{5.4cm}>{\raggedright\arraybackslash}X}
\toprule
\rowcolor{gray!25}
\rule{0pt}{2.6ex}\textbf{Survey} & \textbf{Scenarios} & \textbf{Loop decomposition} & \textbf{Focus} \\
\midrule
\rowcolor{white}\multicolumn{4}{@{}l}{\;\,\ding{60} \textit{Multimodal agents}} \\
Durante \etal~\cite{durante2024agent} {\tiny arXiv'24} & EM, games, healthcare, video, NLP & -- & Agent AI paradigm and applications \\
Xie \etal~\cite{xie2024large} {\tiny arXiv'24} & MG, GW, EM, games, driving, video & -- & Components and planner types \\
Hu \etal~\cite{hu2025osagents} {\tiny ACL'25} & GW & -- & Construction and evaluation of OS agents \\
Yao \etal~\cite{yao2025survey} {\tiny arXiv'25} & GW, EM, deep research, healthcare, driving, recommendation & -- & Agentic capabilities of MLLMs \\
Huang \etal~\cite{huang2026agentic} {\tiny arXiv'26} & MG & -- & Agentic control of visual generation \\
\midrule
\rowcolor{white}\multicolumn{4}{@{}l}{\;\,\ding{60} \textit{Self-evolving and adapting agents}} \\
Fang \etal~\cite{fang2025comprehensive} {\tiny arXiv'25} & Biomedicine, software, finance, law & Inputs / agent system / environment / optimizer & Prompt, memory, tool, and multi-agent optimization \\
Jiang \etal~\cite{jiang2025adaptation} {\tiny arXiv'25} & GW, deep research, software, drug discovery & Agent {vs.}\ tool $\times$ signal source & Adaptation by post-training, memory, and skills \\
Chen \etal~\cite{chen2026recursive} {\tiny arXiv'26} & General & Improvement target $\times$ loop closure & Recursive self-improvement \\
Gao \etal~\cite{gao2025survey} {\tiny TMLR'26} & GW, software, finance, healthcare, education & What / when / how / where & Self-evolution strategies \\
Pu \etal~\cite{pu2026trustworthy} {\tiny arXiv'26} & General & Diagnosis / proposal / evaluation / update & Trustworthiness of recursive self-improvement \\
Ren \etal~\cite{ren2026selfimprovements} {\tiny arXiv'26} & GW, EM, software, games, science & Model {vs.}\ scaffolding $\times$ signal source & Model and scaffolding self-improvement \\
Xiang \etal~\cite{xiang2025systematic} {\tiny TechRxiv'26} & Science, software, open-world simulation & Model-centric / environment-centric / co-evolution & Environment-driven co-evolution \\
Xu \etal~\cite{xu2026dynamic} {\tiny arXiv'26} & General & Typed graph state / schema-constrained rewrites & Self-evolution as graph transformation \\
Zhang \etal~\cite{zhang2025landscape} {\tiny TMLR'26} & GW, EM, deep research, software, math, vision & State / action / transition / reward & Agentic reinforcement learning \\
\midrule
\rowcolor{white}\multicolumn{4}{@{}l}{\;\,\ding{60} \textit{Self-evolving multimodal models and agents}} \\
Deng \etal~\cite{deng2025selfimprovement} {\tiny EMNLP Findings'25} & General & Collection / organization / optimization & Self-improvement of MLLMs \\
Feng \etal~\cite{feng2026selfevolving} {\tiny arXiv'26} & EM, driving, aerial & Memory / task / environment / embodiment / model & Self-evolving embodied AI \\
Guo \etal~\cite{guo2026memory} {\tiny arXiv'26} & General & Formation $\times$ management tier & Multimodal agent memory \\
\textbf{Ours} & \textbf{MR, MG, GW, EM} & \textbf{EG / FA / UC} & \textbf{Cross-scenario portability} \\
\bottomrule
\end{tabularx}
\end{table*}

However, as summarized in \cref{tab:survey_comparison}, existing work leaves this cross-scenario asymmetry unaddressed. 
Primary studies benchmark exclusively within specific scenarios~\cite{asg,searrobot}, while prior surveys either overlook the self-evolving loop as an analytical object~\cite{durante2024agent,xie2024large,yao2025survey,wang2024survey,xi2025rise} or formalize it as scenario-agnostic~\cite{fang2025comprehensive, gao2025survey, ren2026selfimprovements}.
Even the closest related work~\cite{jiang2025adaptation} attributes verifiable feedback and retraining costs to domain-level paradigms, not to specific mechanisms. 
To fill this gap, we first formalize the surveyed object through \textit{a genus and a differentia}. 
The genus comprises autonomous learning loops satisfying four criteria: (\emph{i}) \emph{autonomy}, where the agent operates without human intervention across goals, actions, feedback, and updates post-initialization; (\emph{ii}) \emph{freshness}, where experience is generated under the current policy; (\emph{iii}) \emph{durability}, where updates persist beyond the episode boundary; and (\emph{iv}) \emph{recurrence}, where the loop iterates across cycles. 
The differentia, \emph{self-reference}, requires the agent to realize at least one component of its own loop: the goal sampler, the feedback function, or the evolution operator. 
Loops in the genus lacking self-reference, termed \textit{fixed loops}, serve as controls across all scenarios to disentangle scenario-imposed constraints from the intrinsic effects of self-evolution.

To locate where a mechanism can transfer, we adopt \textbf{the three-stage self-evolving loop} as our analytical framework, decomposing each cycle into experience generation, feedback acquisition, and update consolidation. 
We formalize each stage's bottleneck through two governing factors: (\emph{i}) rollout count and information gain, (\emph{ii}) feedback fidelity and credit localization, and (\emph{iii}) retention and generalization. 
Each scenario imposes ceilings on the rollouts affordable per cycle ($\bar{K}$) and the feedback fidelity attainable by admissible evaluators ($\bar{F}$), which span a two-dimensional \textit{positioning matrix}. 
In this matrix, multimodal reasoning is situated where both ceilings are high, multimodal generation where $\bar{F}$ is low, GUI and Web navigation where $\bar{K}$ is low, and embodied manipulation where both are low, with simulation raising both ceilings.
Comparing self-evolving and fixed loops within each cell further separates the limits imposed by scenario ceilings from those incurred by self-evolving design, \eg, replacing answer keys with sampled consensus in reasoning~\cite{evolmm, visplay}. 
The matrix thus identifies which ceilings bind in each scenario, but alone cannot determine whether a given mechanism is portable.

Building on this matrix, we formulate each self-evolving mechanism as a \textit{trade-off}, in which the agent raises a scarce factor by replacing a fixed component with its own records, proposals, judgments, or memory writes, at the price of presupposing a task-structural resource, such as output equivalence or checkable outcomes. 
In a target scenario, a mechanism is thus \textit{portable} if the target holds this resource, \textit{conditional} if it holds the resource only for a sub-family or after adaptation, \textit{blocked} if the outcome to be judged is unobservable, and \textit{redundant} if the raised factor is not scarce. 
We attribute the cross-scenario asymmetry structurally to the different combinations of scarce factors and resources that the two directions of a scenario pair encounter: \eg, voting finds no output equivalence in manipulation, whereas world models raise a factor that is not scarce in reasoning. 
These verdicts, inferred from single-scenario evidence and not yet tested, define testable preconditions, not empirical guarantees. 
A gain over a budget-matched ablation without self-evolution would overturn any verdict that predicts no benefit from an unadapted transfer, which turns these predictions into a concrete measurement agenda.

This paper makes four contributions, the first two to the survey and the last two to the analysis.
\begin{enumerate}[label={\ding{60}},leftmargin=1.3em, topsep=0.1em, itemsep=0.2em]
\item \textbf{A condition-based definition and paradigm demarcation.} We operationalize autonomy and the differentia through a provenance matrix that records which party realizes each loop component, and apply the definition to sixteen neighboring paradigms, of which five fall outside the genus, two never satisfy the differentia, five are divided by it, and four are self-evolving by definition. 
\item \textbf{A stage-resolved review and positioning matrix.} We review each scenario through the three-stage self-evolving loop, place the four scenarios in the positioning matrix, contrast self-evolving with fixed loops in each cell, and trace five degradation modes along the cross-cycle paths through which they propagate.
\item \textbf{Directed portability across scenario pairs.} We recast the mechanisms of all four scenarios as trade-offs and assign each a portability verdict for every target scenario, yielding a structural account of the cross-scenario asymmetry.
\item \textbf{A falsifiable measurement agenda.} We review the evaluation of task-solving capability, self-evolving capability, and judges, and identify the measurements whose absence leaves untestable three claims, \ie, that a gain stems from self-evolution, that a scenario occupies its assigned position, and that the verdicts hold. 
\end{enumerate}
The remainder of this paper is organized as follows. \cref{sec:preliminary} defines the surveyed object, and \cref{sec:framework} develops the three-stage loop and the positioning matrix. \cref{sec:mr,sec:mg,sec:gw,sec:em} review the four scenarios, and \cref{sec:cross_scenario} derives the portability verdicts. Finally, \cref{sec:evaluation} treats evaluation, and \cref{sec:open_challenges,sec:conclusion} discuss open challenges and conclude.

\section{Preliminaries}
\label{sec:preliminary}

\begin{defpair}
\begin{definition}[\,Multimodal Agent\,]
\label{def:mm_agent}
A \emph{multimodal agent} over observation, action, and goal spaces $(\mathcal{O}, \mathcal{U}, \mathcal{G})$ is specified by the four-component tuple
\begin{equation*}
\mathcal{A} \;=\;
\bigl(\,\Phi_\theta,\; \mathcal{M},\; \mathcal{T},\; \mathcal{I}\,\bigr),
\end{equation*}
whose components are the following.
\begin{itemize}[nosep, leftmargin=1.em]
\item $\Phi_\theta : (\mathbb{R}^{d})^{*} \to \Delta(\widetilde{\mathcal{V}} \sqcup \mathcal{Z}_{\mathrm{latent}})$ is the \textbf{core foundation model}, a next-step distribution over a discrete multimodal vocabulary $\widetilde{\mathcal{V}}$ and a latent space $\mathcal{Z}_{\mathrm{latent}} \subseteq \mathbb{R}^{d'}$ for diffusion or flow-matching heads, whose iterated sampling defines the sequence-level distribution $\Phi_\theta^{\star}$.
\item $\mathcal{M}\!=\!\mathcal{M}_{\mathrm{short}} \times \mathcal{M}_{\mathrm{long}}$ is the \textbf{memory}, an episode-scoped buffer $\mathcal{M}_{\mathrm{short}} \subseteq (\mathbb{R}^{d})^{*}$ and a persistent key-indexed store $\mathcal{M}_{\mathrm{long}} \subseteq \bigl(\mathbb{R}^{d} \times (\mathbb{R}^{d})^{*}\bigr)^{*}$. A write operator $\Psi : \mathcal{M} \!\times\! \mathcal{O} \!\times\! \mathcal{U} \to \mathcal{M}_{\mathrm{short}}$ updates the buffer within an episode, and only the evolution operator of \cref{def:se_agent} writes $\mathcal{M}_{\mathrm{long}}$.
\item $\mathcal{T}\!=\!\{\eta_j : \mathcal{X}_j \to \Delta(\mathcal{Y}_j)\}_{j=1}^{|\mathcal{T}|}$ is the \textbf{tool set}, a registry of typed callable interfaces. Tool actions $\mathcal{U}_{\mathrm{tool}}$ are calls of a fixed form, a tool name with its arguments, $\mathcal{T}$ fixes which calls execute, and their execution effects enter the transition $P$ of \cref{def:mm_pomdp}.
\item $\mathcal{I}\!=\!(\phi, \rho)$ is the \textbf{interaction interface}, a prompt assembler $\phi : \mathcal{O} \!\times\! \mathcal{M} \!\times\! \mathcal{G} \!\to\! (\mathbb{R}^{d})^{*}$ that embeds heterogeneous context into one sequence and an action parser $\rho : (\widetilde{\mathcal{V}} \sqcup \mathcal{Z}_{\mathrm{latent}})^{*} \!\to \mathcal{U}$ that decodes a generated sequence into an action.
\end{itemize}
The agent is multimodal when $\mathcal{O}$ or $\mathcal{U}$ contains a non-textual channel that $\Phi_\theta$ encodes or decodes natively.
\end{definition}
\defpairgap
\begin{definition}[\,Multimodal Environment\,]
\label{def:mm_pomdp}
Under a goal $g \in \mathcal{G}$, the environment of a multimodal agent is the goal-conditioned POMDP
\begin{equation*}
\mathcal{P}_{g} \;=\;
\bigl(\,\mathcal{S},\;\mathcal{O},\;\mathcal{U},\;\mathcal{G},\;\mu_0,\;P,\;Z,\;R,\;\gamma\,\bigr),
\end{equation*}
with latent state space $\mathcal{S}$, initial-state distribution $\mu_0(\cdot \mid g) \in \Delta(\mathcal{S})$, transition kernel $P : \mathcal{S} \times \mathcal{U} \to \Delta(\mathcal{S})$ that includes tool-execution effects, observation function $Z : \mathcal{S} \to \Delta(\mathcal{O})$, goal-conditioned reward $R : \mathcal{S}\times\mathcal{U}\times\mathcal{G} \to \mathbb{R}$, which we treat as the idealized task objective, and discount $\gamma \in [0, 1]$, which the horizon cap $T_{\max}$ keeps well defined at $\gamma = 1$. The interface spaces are multimodal.
\begin{itemize}[nosep, leftmargin=1.0em]
\item $\mathcal{O} = \mathcal{O}_{\mathrm{vis}}^{\bot} \times \mathcal{O}_{\mathrm{txt}}^{\bot} \times \mathcal{O}_{\mathrm{aud}}^{\bot} \times \mathcal{O}_{\mathrm{prop}}^{\bot}$ ranges over visual, textual, auditory, and proprioceptive or tactile channels, where $\mathcal{X}^{\bot} := \mathcal{X} \cup \{\bot\}$ and $\bot$ marks a channel absent at a step.
\item $\mathcal{U} = \mathcal{U}_{\mathrm{text}} \sqcup \mathcal{U}_{\mathrm{gen}} \sqcup \mathcal{U}_{\mathrm{act}} \sqcup \{\textsc{stop}\}$ admits one category per step, text generation, artifact synthesis, or digital and physical intervention $\mathcal{U}_{\mathrm{act}} = \mathcal{U}_{\mathrm{tool}} \sqcup \mathcal{U}_{\mathrm{gui}} \sqcup \mathcal{U}_{\mathrm{motor}}$.
\item $\mathcal{G}$ contains multi-channel goals, such as a language constraint paired with a target image or pose, together with any task parameter, such as an initial scene, that a goal fixes.
\end{itemize}
\end{definition}
\vspace{-5mm}
\end{defpair}

Conceptually, a \emph{multimodal agent} is a goal-directed decision-making system that interacts with an external environment through a closed perception-cognition-action loop over heterogeneous observation and scenario-specific action spaces~\cite{russell2009artificial}. 
We restrict the scope to agents whose cognitive core is a large (multimodal) foundation model~\cite{qwen25vl,bai2025qwen3,chen2024internvl,team2024chameleon,xie2024showo,chen2025janus,wang2024emu3,brohan2023rt} extended via interface~\cite{willard2023outlines,geng2023grammar,dong2024xgrammar}, memory~\cite{packer2024memgpt}, and tools~\cite{yao2023react,schick2023toolformer,qin2024toolllm,patil2024gorilla}, excluding architectures lacking such a core, such as world-model planners and task-specific visuomotor policies. 
Since the environment supplies perceptions, accepts actions, and specifies goals, we adopt \emph{{multimodal agent-environment interaction}} as our formal scaffold. 
To systematically diagnose the cross-scenario asymmetry, we unfold this loop across two temporal scales: a \emph{micro-scale} capturing step-by-step behaviors within a single episode, and a \emph{macro-scale} tracking the cycle-by-cycle evolution of the agent's own components over long-term experiences. 
The resulting dual-scale dynamics are formalized under goal-conditioned partially observable Markov decision processes (POMDPs)~\cite{kaelbling1998planning,schaul2015uvfa}.
On this scaffold, we define the surveyed object across four levels, from the static agent and environment through the learning loop and the autonomous learning loop to the self-evolving loop.

\subsection{Multimodal Agent}\label{sec:mm_agent}

We formulate the static architecture of a multimodal agent as a four-component tuple $\mathcal{A}$ (\cref{def:mm_agent}) and its environment as a goal-conditioned, modality-decomposed POMDP $\mathcal{P}_{g}$ (\cref{def:mm_pomdp}). 
The multimodality condition of \cref{def:mm_agent} requires $\Phi_\theta$ to directly encode or decode a non-textual channel without intermediate text conversion, such as screen pixels, camera frames, generated images or videos, or low-level motor commands. 
Under this criterion, an agent perceiving a web page solely through its HTML or accessibility tree is text-only despite rendered visual content.

\begin{defpair}
\begin{definition}[\,Self-Evolving Multimodal Agent\,]\label{def:se_agent}
Let $\mathcal{A}_k = (\Phi_{\theta_k}, \mathcal{M}_k, \mathcal{T}_k, \mathcal{I}_k)$ denote the agent at cycle $k \in \mathbb{N}$. A \emph{self-evolving multimodal agent} is specified by the triple
\begin{equation*}
\mathcal{E}_{\mathcal{A}} \;=\;
\bigl(\,\mathcal{A}_0,\; V_\nu,\; f_\zeta\,\bigr),
\end{equation*}
whose components are the following.
\begin{itemize}[nosep, leftmargin=1.em]
\item $\mathcal{A}_0 \in \mathbb{A}^{\mathrm{mm}}$ is the \textbf{initial agent}, with $\mathbb{A}^{\mathrm{mm}}$ the set of configurations of \cref{def:mm_agent}.
\item $V_\nu : (\mathbb{D}^{\mathrm{mm}})^{*} \times (\mathbb{A}^{\mathrm{mm}})^{*} \to (\mathbb{Y}^{\mathrm{mm}})^{*}$ is the \textbf{feedback function} with configuration $\nu$. It maps a batch of goal-trajectory pairs $(g,\tau) \in \mathbb{D}^{\mathrm{mm}}$, with $\tau$ distributed as in \cref{eq:traj-dist}, together with the agent history $\mathcal{A}_{0:k}$, to feedback in $\mathbb{Y}^{\mathrm{mm}}$, which subsumes scalar rewards, preferences, and step-level, token-level, or structured critiques, and we write $V_\nu(\mathcal{D}_k)$ for $V_\nu(\mathcal{D}_k, \mathcal{A}_{0:k})$. It is \emph{agent-independent} when the preference it induces over trajectories that share a goal depends only on those trajectories and on artifacts fixed before the loop other than copies of the agent's model, and it \emph{reads the current agent} when that preference depends on $\mathcal{A}_k$, through the agent's model or through other samples of its current policy.
\item $f_\zeta : \mathbb{A}^{\mathrm{mm}} \times \mathbb{H} \times (\mathbb{D}^{\mathrm{mm}})^{*} \times (\mathbb{Y}^{\mathrm{mm}})^{*} \to \mathbb{A}^{\mathrm{mm}}$ is the \textbf{evolution operator} with configuration $\zeta$, which writes the update of a cycle into the agent tuple and reads the history $\mathcal{H}_k$ of \cref{def:se_env} when it replays earlier experience.
\end{itemize}
A proposer or evaluator that the agent system trains belongs to $\Phi_\theta$, which places it under the updates of $f_\zeta$ and makes $\sigma_\omega$ and $V_\nu$ depend on it through $\mathcal{A}_k$. The criteria $\mathcal{C}_1$ to $\mathcal{C}_4$ and the differentia below decide whether the loop of the triple with the environment of \cref{def:se_env} is an autonomous learning loop and whether that loop is self-evolving, and only a triple whose loop passes all five tests is a self-evolving multimodal agent.
\end{definition}
\defpairgap
\begin{definition}[\,Multimodal Macro-Environment\,]\label{def:se_env}
Let $\mathcal{H}_k \!=\! (\mathcal{A}_{0:k}, \mathcal{D}_{0:k-1}, V_\nu(\mathcal{D}_{0:k-1})) \in \mathbb{H}$ denote the history of agents, trajectories, and realized scores at the start of cycle $k$. The \textbf{multimodal macro-environment} is the triple
\begin{equation*}
\mathcal{E}_{\mathcal{E}} \;=\; \bigl(\,\{\mathcal{P}_{g}\}_{g\in\mathcal{G}},\; \sigma_\omega,\; \mu_{\mathcal{G}}\,\bigr),
\end{equation*}
whose components are the following.
\begin{itemize}[nosep, leftmargin=1.em]
\item $\{\mathcal{P}_{g}\}$ is a task-indexed family of POMDPs of \cref{def:mm_pomdp} that share $\mathcal{S}$, $\mathcal{O}$, $\mathcal{U}$, $P$, $Z$, and $R$, in which the task argument $g$ fixes the goal and may fix the initial-state distribution or a generated scene. Environment synthesis therefore acts through $g$, whereas drift of $P$ acts on the loop from outside.
\item $\sigma_\omega : \mathbb{N} \times \mathbb{H} \to \Delta(\mathcal{G})$ is the \textbf{goal sampler} with configuration $\omega$, which draws the goals of cycle $k$ from $\sigma_\omega(k, \mathcal{H}_k)$. It is \emph{stationary} when it ignores $(k, \mathcal{H}_k)$, \emph{scheduled} when it depends on $k$ alone, as a difficulty order fixed before the loop does, and \emph{adaptive} when it reads $\mathcal{H}_k$, as a success-conditioned curriculum does. Placing $\sigma_\omega$ in the macro-environment locates where goals enter the loop, whereas the provenance matrix of \cref{sec:defining_conditions} records who realizes it.
\item $\mu_{\mathcal{G}} \in \Delta(\mathcal{G})$ is a fixed \textbf{evaluation distribution} outside the control of $\sigma_\omega$, which measures performance through $J(\mathcal{A};\mu_{\mathcal{G}}) := \mathbb{E}_{g\sim\mu_{\mathcal{G}}} J(\mathcal{A};g)$.
\end{itemize}
\end{definition}
\vspace{-5mm}
\end{defpair}

\pgraphhlB{Micro-Scale Interaction Loop}\,
Within one episode, shown in the micro-scale panel of \cref{fig:2_loop}, the agent acts under a configuration $\mathcal{A}$ with its persistent components $(\Phi_\theta, \mathcal{M}_{\mathrm{long}}, \mathcal{T}, \mathcal{I})$ held fixed, leaving only the short-term buffer $\mathcal{M}_{\mathrm{short}}$ to evolve from initial state $m_0$. 
The episode starts from $s_0 \sim \mu_0(\cdot \mid g)$, runs over $t \in \{0, \dots, T-1\}$, and terminates upon a \textsc{stop} action or horizon limit $T_{\max}$, yielding a random stopping time $T \le T_{\max}$. 
Each step passes through the environment kernels and agent components:
\begin{equation}
\label{eq:micro_execution_flow}
s_t \xrightarrow{Z} o_t \xrightarrow{\phi(\cdot, m_t, g)} x_t \xrightarrow{\Phi_\theta^\star} y_t \xrightarrow{\rho} u_t \xrightarrow{P} s_{t+1},
\end{equation}
where $\phi$ and $\rho$ are deterministic, $Z$, $\Phi_\theta^\star$, and $P$ are stochastic, and the buffer updates as $m_{t+1} = \Psi(m_t, o_t, u_t)$. 
The fixed argument $\mathcal{M}_{\mathrm{long}}$ in $\phi$ is suppressed per episode. 
This synchronous discrete-time chain abstracts away the gap between decision and control frequencies, which is immaterial for text actions but adds per-rollout cost under real-time embodied control. 
Over the micro-history $h_{t} = (o_{0:t}, u_{0:t-1})$, the agent induces the policy $\pi_{\mathcal{A}} = \rho_{\sharp}\Phi_\theta^{\star}$,
\begin{equation}
\label{eq:induced-policy}
\pi_{\mathcal{A}}(u_{t}| h_{t}, g) \!=\!
\Phi_\theta^{\star}\!\bigl(\rho^{-1}(u_{t}) \,\big|\, \phi(o_{t}, m_{t}, g)\bigr),
\end{equation}
where conditioning on $h_t$ suffices, fixed $(\mathcal{A}, m_0)$ and update $\Psi$ uniquely determining $m_t$ from $h_t$. 
Compounded with environment transitions, this policy induces a trajectory distribution $p^{\pi_{\mathcal{A}}}_{g}(\tau)$ over $\tau = (s_{0:T},\, o_{0:T-1},\, u_{0:T-1})$:
\begin{equation}
\label{eq:traj-dist}
p^{\pi_{\mathcal{A}}}_{g}(\tau) \!=\!
\mu_{0}(s_{0} \mid g)\!\! \prod_{t=0}^{T-1}\!\! Z(o_{t}| s_{t}) \pi_{\mathcal{A}}(u_{t}| h_{t}, g) P(s_{t+1}| s_{t}, u_{t}).
\end{equation}

\pgraph{Goal-Conditioned Expected Return}
The loop induces the goal-conditioned expected return
\begin{equation}
\label{eq:objective}
J(\mathcal{A};\,g) \!=\!
\mathbb{E}_{\tau \,\sim\, p^{\pi_{\mathcal{A}}}_{g}} \!\left[\,\sum_{t=0}^{T-1} \gamma^{t}\, R(s_t,\, u_t,\, g)\,\right],
\end{equation}
a performance functional of $\mathcal{A}$, not a quantity optimized within an episode. 
Self-evolution modifies $\mathcal{A}$ between episodes, as the macro-scale loop below formalizes.

\subsection{Self-Evolving Multimodal Agent}\label{sec:self_evolving_mm_agent}

Informally, a self-evolving multimodal agent improves its persistent components across cycles without human input post-initialization, learns from experience generated under its own policy, and internally realizes at least one improvement driver: selecting its goals, deriving its feedback, or authoring its updates. 
We formalize the agent and its environment as the triples $\mathcal{E}_{\mathcal{A}}$ (\cref{def:se_agent}) and $\mathcal{E}_{\mathcal{E}}$ (\cref{def:se_env}).


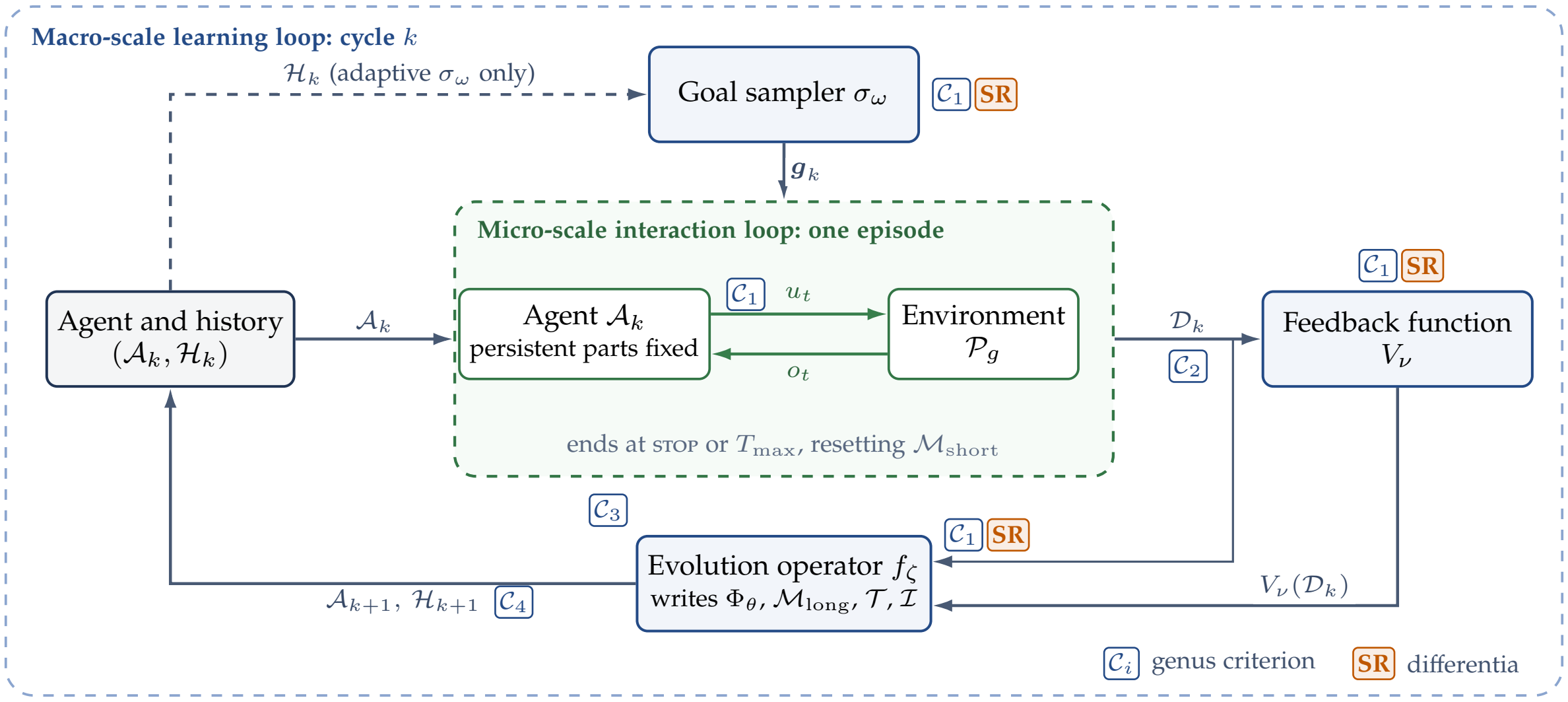
\begin{figure*}[!t]
\centering
\begin{tikzpicture}[
  font=\footnotesize,
  >={Latex[length=2mm,width=1.25mm]},
  op/.style={rectangle, rounded corners=3pt, draw=themeblue!80!black, line width=0.75pt,
    fill=themeblue!6, minimum width=2.7cm, minimum height=0.95cm, align=center, inner sep=3pt},
  state/.style={rectangle, rounded corners=3pt, draw=themedark, line width=0.75pt,
    fill=themedark!5, minimum width=2.1cm, minimum height=0.95cm, align=center, inner sep=3pt},
  inner/.style={rectangle, rounded corners=3pt, draw=cardteal!70!black, line width=0.7pt,
    fill=white, minimum width=1.9cm, minimum height=0.9cm, align=center, inner sep=3pt},
  flow/.style={->, draw=themedark!82, line width=0.85pt},
  micro/.style={->, draw=cardteal!75!black, line width=0.85pt},
  lbl/.style={font=\scriptsize, color=themedark!90, fill=white, inner sep=1.5pt, align=center},
  crit/.style={draw=themeblue!80!black, fill=white, text=themeblue!80!black, rounded corners=1.4pt,
    line width=0.5pt, font=\scriptsize\bfseries, inner sep=1.2pt, minimum height=9pt},
  sr/.style={draw=rlcolor, fill=rlcolor!10, text=rlcolor, rounded corners=1.4pt,
    line width=0.5pt, font=\scriptsize\bfseries, inner sep=1.2pt, minimum height=9pt}
]
\node[rectangle, rounded corners=6pt, draw=cardteal!70!black, dashed, line width=0.8pt,
  fill=cardteal!5, minimum width=6.6cm, minimum height=2.75cm] (panel) at (0,0) {};
\node[anchor=north west, font=\scriptsize\bfseries, color=cardteal!60!black]
  at ([shift={(0.12,-0.08)}]panel.north west) {Micro-scale interaction loop: one episode};
\node[inner] (agent) at (-2.0,0.05) {Agent $\mathcal{A}_k$\\\scriptsize persistent parts fixed};
\node[inner] (env) at (2.0,0.05) {Environment\\$\mathcal{P}_g$};
\draw[micro] ([yshift=0.2cm]agent.east) -- node[above, font=\scriptsize, color=cardteal!60!black] (ut) {$u_t$} ([yshift=0.2cm]env.west);
\draw[micro] ([yshift=-0.2cm]env.west) -- node[below, font=\scriptsize, color=cardteal!60!black] {$o_t$} ([yshift=-0.2cm]agent.east);
\node[anchor=south, font=\scriptsize, color=themedark!75] at ([yshift=0.07cm]panel.south)
  {ends at \textsc{stop} or $T_{\max}$, resetting $\mathcal{M}_{\mathrm{short}}$};

\node[state] (state) at (-6.15,0) {Agent and history\\$(\mathcal{A}_k,\mathcal{H}_k)$};
\node[op] (sigma) at (0,2.45) {Goal sampler $\sigma_\omega$};
\node[op] (V) at (6.15,0) {Feedback function\\$V_\nu$};
\node[op] (f) at (0,-2.45) {Evolution operator $f_\zeta$\\\scriptsize writes $\Phi_\theta$, $\mathcal{M}_{\mathrm{long}}$, $\mathcal{T}$, $\mathcal{I}$};

\draw[flow, dashed] (state.north) |- node[lbl, pos=0.75, above] {$\mathcal{H}_k$ (adaptive $\sigma_\omega$ only)} (sigma.west);
\draw[flow] (state.east) -- node[lbl, above] {$\mathcal{A}_k$} (panel.west);
\draw[flow] (sigma.south) -- node[lbl, right] {$\boldsymbol{g}_k$} (panel.north);
\draw[flow] (panel.east) -- node[lbl, above] (dk) {$\mathcal{D}_k$} (V.west);
\draw[flow] (V.south) |- node[lbl, pos=0.6, above] {$V_\nu(\mathcal{D}_k)$} ([yshift=-0.22cm]f.east);
\draw[flow, line width=0.6pt] (4.5,0) |- ([yshift=0.22cm]f.east);
\draw[flow] (f.west) -| node[lbl, pos=0.25, below] (ret) {$\mathcal{A}_{k+1},\,\mathcal{H}_{k+1}$} (state.south);

\begin{scope}[on background layer]
\node[rectangle, rounded corners=8pt, draw=themeblue!55, dashed, line width=0.7pt,
  minimum width=15.6cm, minimum height=6.9cm] (frame) at (0,-0.12) {};
\end{scope}
\node[anchor=north west, font=\scriptsize\bfseries, color=themeblue!80!black]
  at ([shift={(0.15,-0.1)}]frame.north west) {Macro-scale learning loop: cycle $k$};

\node[crit, anchor=west] (c1s) at ([xshift=0.12cm]sigma.east) {$\mathcal{C}_1$};
\node[sr, right=1pt of c1s] {SR};
\node[crit, anchor=south east] (c1v) at ([yshift=0.08cm]V.north) {$\mathcal{C}_1$};
\node[sr, right=1pt of c1v] {SR};
\node[crit, anchor=west] (c1f) at ([xshift=0.12cm]f.east |- f.north) {$\mathcal{C}_1$};
\node[sr, right=1pt of c1f] {SR};
\node[crit, anchor=south east] at ([shift={(-0.08,0.08)}]f.north west) {$\mathcal{C}_3$};
\node[crit, left=2pt of ut] {$\mathcal{C}_1$};
\node[crit, below=3pt of dk] {$\mathcal{C}_2$};
\node[crit, right=2pt of ret] {$\mathcal{C}_4$};

\node[crit, anchor=west] (lg1) at (3.2,-3.25) {$\mathcal{C}_i$};
\node[anchor=west, font=\scriptsize, color=themedark!85] (lt1) at (lg1.east) {genus criterion};
\node[sr, anchor=west] (lg2) at ([xshift=0.25cm]lt1.east) {SR};
\node[anchor=west, font=\scriptsize, color=themedark!85] at (lg2.east) {differentia};
\end{tikzpicture}
\caption{\textbf{Dual-scale learning loop and its defining conditions.}
Within an episode (green), $\mathcal{A}_k$ acts in $\mathcal{P}_g$ with its persistent components fixed.
Across episodes (blue), each cycle samples goals with $\sigma_\omega$, scores the resulting batch $\mathcal{D}_k$ with $V_\nu$, and updates the agent with $f_\zeta$ (\cref{eq:macro_evolution_flow}).
Blue tags mark where each genus criterion applies, and orange SR tags mark the meta-operators, at least one of which a self-evolving agent must realize.
}
\label{fig:2_loop}
\end{figure*}

\pgraphhlB{Macro-Scale Learning Loop}\,
Across episodes, the macro-scale frame (\cref{fig:2_loop}) wraps interactions into agent-updating cycles. 
Starting from $\mathcal{A}_0$, each cycle $k$ passes through goal sampling, rollout, evaluation, and update:
\begin{equation}
\label{eq:macro_evolution_flow}
\mathcal{H}_k \xrightarrow{\sigma_\omega} \boldsymbol{g}_k \xrightarrow{p^{\pi_{\mathcal{A}_k}}_{\boldsymbol{g}_k}} \mathcal{D}_k \xrightarrow{V_\nu} V_\nu(\mathcal{D}_k) \xrightarrow{f_\zeta} \mathcal{A}_{k+1},
\end{equation}
where $\boldsymbol{g}_k$ denotes the goal batch drawn from $\sigma_\omega(k,\mathcal{H}_k)$ with support $\mathcal{G}_k \!\subseteq\! \mathcal{G}$, rollout law $p^{\pi_{\mathcal{A}_k}}_{\boldsymbol{g}_k}$ applies $p^{\pi_{\mathcal{A}_k}}_{g}$ across each $g \in \boldsymbol{g}_k$, and $\mathcal{D}_k$ aggregates the resulting goal-trajectory pairs (with single goals corresponding to unit batch size). 
The history then extends to $\mathcal{H}_{k+1} = (\mathcal{A}_{0:k+1}, \mathcal{D}_{0:k}, V_\nu(\mathcal{D}_{0:k}))$, recording realized scores, stochastic evaluator outputs being non-recoverable from trajectories alone and failure-conditioned samplers conditioning directly on realized values. 
We designate this recursive mapping a \emph{learning loop}. 
The term denotes the cross-cycle structure without presupposing monotonic gains in $J(\mathcal{A}_k;\mu_{\mathcal{G}})$, and excludes within-episode adaptation (\eg, in-context learning), which remains confined to $\mathcal{M}_{\mathrm{short}}$.
A \emph{cycle} is one iteration of this mapping. 
When training proceeds in explicit rounds (each regenerating tasks, evaluator states, or rollouts via the antecedent agent before optimization), a cycle corresponds to one round, whose internal optimizer steps form \emph{within-cycle optimization} under $f_\zeta$. 
In round-free regimes, such as online reinforcement learning drawing fresh rollouts at every optimizer step, each step executes the full mapping as a \emph{degenerate cycle}.

\pgraph{Meta-Operators and Substrate}
We designate $\sigma_\omega$, $V_\nu$, and $f_\zeta$ the three meta-operators of the loop, each acting on whole episodes at the macro scale. 
Across multimodal scenarios, the goal sampler draws from a fixed task pool~\cite{simplevlarl}, poses questions about unlabeled images~\cite{visplay}, reselects GUI tasks by historical success~\cite{mobilerl}, or synthesizes simulated scenes~\cite{wang2024robogen}. 
The feedback function verifies discrete answers~\cite{pvse}, inspects generated images via the internal understanding pathway~\cite{suder}, queries a vision-language judge on the final screen~\cite{digirl}, or evaluates simulator success predicates~\cite{simplevlarl}. 
The evolution operator updates $\Phi_\theta$ via policy gradient~\cite{shao2024deepseekmath} or writes lessons, skills, and memory entries~\cite{zhao2024expel,wang2023voyager,wu2026co}. Auxiliary agents enter either as self-play opponents sampled via the task argument (\cref{def:se_env})~\cite{vzero} or as learned proposer-solver roles instantiated within $\Phi_\theta$~\cite{evolmm}. 
The \emph{substrate} denotes the persistent components that $f_\zeta$ modifies across cycles~\cite{fang2025comprehensive, gao2025survey}. Specifically, a \emph{parametric} loop updates $\Phi_\theta$ alone, a \emph{non-parametric} loop modifies only $\mathcal{M}_{\mathrm{long}}$, $\mathcal{T}$, or $\mathcal{I}$ under fixed $\theta$ (training-free), and a \emph{mixed} loop alters both~\cite{agenticvla,perfguard}.

\pgraph{Expected Evaluation Return}
We evaluate a learning loop at each cycle via its expected return under the evaluation distribution $\mu_{\mathcal{G}}$, which lies outside the control of $\sigma_\omega$ to prevent samplers from inflating performance through easier goals. 
With cycles counted in rounds by some methods and in optimizer steps by others, $n$ cycles span horizons of different lengths, and the return is therefore reported per cycle. With $\mathcal{A}_k$ itself random, depending on the goals, rollouts, and feedback recorded in $\mathcal{H}_k$, the expectation at cycle $k$ also ranges over this history, 
\begin{equation}\label{eq:eval_seq}
\mathcal{J}_k \!=\!
\mathbb{E}_{\mathcal{H}_k,\, g \,\sim\, \mu_{\mathcal{G}}} \!\left[\,J(\mathcal{A}_k;\,g)\,\right], \quad k \in \mathbb{N}.
\end{equation}
This sequence evaluates against the idealized reward $R$ (\cref{eq:objective}), whereas the loop ascends the proxy $V_\nu$.

\subsection{Defining Conditions of Self-Evolving Loops}\label{sec:defining_conditions}

To distinguish it from general learning loops, we formalize the self-evolving loop through a genus of four operational criteria and a differentia of self-reference.

\pgraphhlM{The Genus: Autonomous Learning Loops}
We adopt four criteria to bound the genus of autonomous learning loops. 
Autonomy bars human input across all three meta-operators and rollout actions, whereas freshness, durability, and recurrence respectively exclude static datasets, episodic buffer writes, and uniterated single-shot updates:
\begin{enumerate}[label=$\mathcal{C}_\arabic*$,ref=(C\arabic*),leftmargin=1.6em,topsep=0.em,itemsep=0.3em]
\item \textbf{Autonomy.\;}
Autonomy requires the loop to operate without human intervention across its meta-operators and action stream post-initialization. 
Formally, at every cycle $k\!\ge\!0$, the goals $\boldsymbol{g}_k$, the signal $V_\nu(\mathcal{D}_k)$, and the update of $f_\zeta$ consume no annotation, verdict, or approval added by a human while the loop runs, and no human-issued action, such as a teleoperated correction, enters $\mathcal{D}_k$. 
Artifacts fixed prior to cycle~$0$ (task sets, verification checkers, answer keys, external evaluators and judges, or optimizers) constitute the bootstrap. Human resets modifying $\mathcal{P}_{g}$ and not the learning loop itself, autonomy does not imply reset-free execution~\cite{sharma2022autonomous}, and reset overhead is documented separately. 
\item \textbf{Freshness.\;}
Freshness requires per-cycle experience to be drawn from the agent's current policy, not from a fixed dataset or an external stream.
Formally, the cycle-$k$ distribution depends on $\mathcal{A}_k$ through $\pi_{\mathcal{A}_k}$, admitting on-policy sampling $\mathcal{D}_k \!\sim\! p^{\pi_{\mathcal{A}_k}}_{\boldsymbol{g}_k}$ and asynchronous variants whose behavior policy lags $\mathcal{A}_k$ by a bounded number of updates, but excluding any agent-independent $p_{\mathrm{static}}$~\cite{gao2025survey}. 
Experience collected once by $\mathcal{A}_0$ and reused in later cycles fails the criterion. Generating on-policy rollouts, however, satisfies freshness without constituting self-reference. 
\item \textbf{Durability.\;}
Durability requires the update to alter the agent beyond the episode boundary.
Formally, $\mathcal{A}_{k+1}\!=\!f_\zeta(\mathcal{A}_k, \mathcal{H}_k, \mathcal{D}_k, V_\nu(\mathcal{D}_k))$ must modify persistent components ($\Phi_{\theta_k}$, long-term memory $\mathcal{M}_{\mathrm{long}} \subseteq \mathcal{M}_k$, toolset $\mathcal{T}_k$, or interface $\mathcal{I}_k$), thereby altering induced policy $\pi_{\mathcal{A}_{k+1}}$ on at least one goal. 
This excludes adaptation confined to the within-episode buffer $\mathcal{M}_{\mathrm{short}}$ and updates restricted to auxiliary proposers or evaluators, which belong to $\Phi_\theta$ yet occupy parameters that $\pi_{\mathcal{A}}$ does not read, while admitting memory writes, tool additions, and prompt mutations even without any gradient descent~\cite{gao2025survey, xiang2025systematic}. 
\item \textbf{Recurrence.\;}
Recurrence requires the loop to close across cycles instead of terminating after a single update.
Formally, the map $\mathcal{A}_k \!\mapsto\! \mathcal{A}_{k+1}$ of $\mathcal{C}_3$ is iterated over more than one cycle, whether rounds or degenerate cycles, and each output becomes the next cycle's agent, not a branch discarded after evaluation. 
This excludes single-pass optimization pipelines ($\mathcal{A}_0 \!\mapsto\! \mathcal{A}_1$), even if compliant with $\mathcal{C}_1$ through $\mathcal{C}_3$. 
The criterion presupposes $\mathcal{C}_3$ and, being satisfied by every multi-step online method, discriminates loops only in round-based regimes. 
\end{enumerate}

\begin{table*}[!t]
\centering
\footnotesize
{\renewcommand\tabularxcolumn[1]{m{#1}}%
\caption{\textbf{Demarcation of self-evolution from established paradigms.}
C1 to C4 test autonomy, freshness, durability, and recurrence, with parentheses locating the runtime human input under C1. 
SR marks the meta-operators the agent realizes, \ie, the goal sampler ($\sigma$), the feedback function ($V$), or the evolution operator ($f$). 
In the instance-dependent group, $^{\circ}$ and $^{\bullet}$ mark fixed-loop and self-evolving instances, respectively.
$^{\dagger}$\,Acts on classifiers, not on the foundation-model agents that the genus presupposes, and its foundation-model counterpart, label-free reinforcement learning on a fixed question pool, falls under Self-Rewarding and Label-Free RL.
(\cYes\ satisfied, \cNo\ unsatisfied, \cPart\ satisfied by some variants, -- inapplicable.) 
}
\vspace{-2mm}
\label{tab:s2_paradigm}
\rowcolors{3}{themelight}{white}
\begin{tabularx}{\textwidth}{%
>{\raggedright\arraybackslash}m{2.75cm}
>{\centering\arraybackslash}m{0.72cm}
>{\centering\arraybackslash}m{0.4cm}
>{\centering\arraybackslash}m{0.4cm}
>{\centering\arraybackslash}m{0.4cm}
>{\centering\arraybackslash}m{1.55cm}
>{\raggedright\arraybackslash}X
>{\raggedright\arraybackslash}m{3.9cm}}
\toprule
\rowcolor{gray!25}
\rule{0pt}{2.6ex}& \multicolumn{4}{c}{\textbf{Genus criteria}} & \textbf{Differentia} & & \\
\noalign{\global\aboverulesep=0pt\global\belowrulesep=0pt}
\arrayrulecolor{gray!25}\cmidrule{1-1}\cmidrule{7-8}\arrayrulecolor{black}\cmidrule(lr){2-5}\cmidrule(lr){6-6}
\noalign{\global\aboverulesep=0.4ex\global\belowrulesep=0.65ex}
\rowcolor{gray!25}
\multirow{-2}{*}{\textbf{Paradigm}} & \rule[-1ex]{0pt}{2.8ex}\textbf{C1} & \textbf{C2} & \textbf{C3} & \textbf{C4} & \textbf{SR} & \multirow{-2}{*}{\textbf{Defining Mechanism}} & \multirow{-2}{*}{\textbf{Representative Works}} \\
\midrule
\rowcolor{white}
\multicolumn{8}{@{}l}{\;\,\ding{60} \textit{Outside the genus}} \\
Self-Training$^{\dagger}$
& \cYes & \cYes & \cYes & \cYes & $V$
& Iterative training of a classifier on its own pseudo-labels over a static unlabeled pool.
& Noisy Student~\cite{xie2020self}; FixMatch~\cite{sohn2020fixmatch}; Meta Pseudo Labels~\cite{pham2021meta}
\\
Test-Time Adaptation$^{\dagger}$
& \cYes & \cYes & \cPart & \cPart & \cPart\,($V$)
& Inference-time adaptation of a classifier to an external test stream under a self-supervised or entropy objective.
& TTT~\cite{sun2020test}; TENT~\cite{wang2021tent}; MEMO~\cite{zhang2022memo}; CoTTA~\cite{wang2022continual}; EATA~\cite{niu2022efficient}
\\
Test-Time Scaling
& \cYes & \cYes & \cNo & -- & \cPart\,($V$)
& Additional inference compute per query by repeated sampling, voting, or search, with nothing persisting past the query.
& Self-Consistency~\cite{wang2023selfconsistency}; Snell \etal~\cite{snell2025scaling}; Corvid~\cite{corvid}
\\
Self-Correction
& \cYes & \cYes & \cNo & -- & \cPart\,($V$)
& Refinement of an output through critique within one episode.
& Self-Refine~\cite{madaan2023self}; CRITIC~\cite{gou2024critic}; Self-Debug~\cite{chen2023teaching}; CoVe~\cite{dhuliawala2023chain}
\\
Interactive Imitation Learning
& \cNo\,(act) & \cYes & \cYes & \cYes & \cNo
& Training on the learner's own state distribution with expert actions or interventions supplied while learning runs.
& DAgger~\cite{ross2011reduction}; $\pi^{*}_{0.6}$~\cite{pistar06}
\\
\midrule
\rowcolor{white}
\multicolumn{8}{@{}l}{\;\,\ding{60} \textit{Instance-dependent, divided by a genus criterion or the differentia}} \\
Self-Improvement
& \cYes & \cYes & \cYes & \cPart & \cPart\,($\sigma$, $V$, $f$)
& Training on self-generated outputs retained by a filter, once or over rounds.
& Self-Instruct~\cite{wang2023selfinstruct}; ReST$^{\circ}$~\cite{rest}; ReST$^{EM}$$^{\circ}$~\cite{singh2023rest}; STaR$^{\bullet}$~\cite{zelikman2022star}
\\
Continual Learning
& \cYes & \cPart & \cYes & \cYes & \cNo
& Sequential learning over a task or data stream fixed in advance, with mechanisms against catastrophic forgetting.
& EWC~\cite{kirkpatrick2017overcoming}; GEM~\cite{lopez2017gradient}; CRL-VLA$^{\circ}$~\cite{crlvla}
\\
RLHF
& \cPart\,($V$) & \cYes & \cYes & \cYes & \cNo
& Policy optimization against a reward model fit to human preferences.
& InstructGPT~\cite{ouyang2022training}; Bai \etal~\cite{bai2022training}; Llama~2~\cite{touvron2023llama}
\\
RLVR
& \cYes & \cYes & \cYes & \cYes & \cPart\,($\sigma$)
& Policy optimization against rewards computed by a programmatic check of the output.
& T\"{u}lu~3$^{\circ}$~\cite{lambert2024tulu}; Kimi k1.5$^{\bullet}$~\cite{kimi2025k15}
\\
Agentic RL
& \cYes & \cYes & \cYes & \cYes & \cPart\,($\sigma$, $V$)
& Policy optimization of a multi-turn agent from rewards earned in an interactive environment.
& SimpleVLA-RL$^{\circ}$~\cite{simplevlarl}; ZeroGUI$^{\circ}$~\cite{zerogui}; DigiRL$^{\bullet}$~\cite{digirl}
\\
Prompt and Workflow Optimization
& \cYes & \cYes & \cYes & \cYes & \cPart\,($f$)
& Iterative rewriting of an agent's instructions or workflow from the scores of its executions.
& TextGrad$^{\circ}$~\cite{yuksekgonul2024textgrad}; SymbOmni$^{\bullet}$~\cite{symbomni}
\\
Experiential Learning
& \cYes & \cYes & \cYes & \cYes & \cPart\,($f$)
& Distillation of an agent's trajectories into lessons, memory entries, or skills that later tasks retrieve.
& UI-Mem$^{\circ}$~\cite{uimem}; ExpeL$^{\bullet}$~\cite{zhao2024expel}; Voyager$^{\bullet}$~\cite{wang2023voyager}
\\
\midrule
\rowcolor{white}
\multicolumn{8}{@{}l}{\;\,\ding{60} \textit{Self-evolving by definition}} \\
Self-Play
& \cYes & \cYes & \cYes & \cYes & $\sigma$
& Training against tasks or opponents drawn from the agent itself or its history.
& Absolute Zero~\cite{azr}; Vision-Zero~\cite{vzero}
\\
Automatic Curriculum Learning
& \cYes & \cYes & \cYes & \cYes & $\sigma$
& Adaptation of the goal distribution to the agent's measured competence.
& Portelas \etal~\cite{portelas2020automatic}; SEAgent~\cite{seagent}
\\
Self-Rewarding and Label-Free RL
& \cYes & \cYes & \cYes & \cYes & $V$
& Training on rewards or preferences that the trained model produces, by judgment or by consensus over its samples.
& Self-Rewarding~\cite{yuan2024self}; TTRL~\cite{zuo2025ttrl}; EvoLMM~\cite{evolmm}
\\
Recursive Self-Improvement
& \cYes & \cYes & \cYes & \cYes & $f$
& Rewriting of the system's own improvement procedure by the system.
& Promptbreeder~\cite{fernando2023promptbreeder}; STOP~\cite{zelikman2023stop}; G\"{o}del Agent~\cite{yin2025godel}
\\
\bottomrule
\end{tabularx}
}
\end{table*}

\pgraphhlM{The Differentia: Self-Reference}
The four criteria constrain how the loop runs, not who realizes it, a loop being able to satisfy $\mathcal{C}_1$--$\mathcal{C}_4$ while every meta-operator is realized outside the agent. 
The differentia, \emph{self-reference} (SR), separates self-evolving loops by requiring the agent to internally realize at least one evolutionary meta-operator, marked by SR tags in \cref{fig:2_loop}.\footnote{%
Our criteria adapt those of~\cite{gao2025survey, fang2025comprehensive, xiang2025systematic} and treat updated components as a descriptive substrate, but enforce a narrower scope. 
While prior work categorizes learning from environmental or rule-based rewards as reward-based self-evolution~\cite{gao2025survey} or subsumes reinforcement learning with verifiable rewards on fixed datasets~\cite{fang2025comprehensive}, self-reference excludes these regimes whenever the agent realizes zero meta-operators. 
It does not require co-evolving environment dynamics~\cite{xiang2025systematic}, evolving inter-agent topologies~\cite{fang2025comprehensive}, or lifelong operation ($k \to \infty$).
}
Self-reference and the meta-operator clause of autonomy both ask who realizes each meta-operator, and we therefore record their evaluation for a method $m$ in the \emph{provenance matrix} $\Rolemat(m) \in \{0,1\}^{3\times 4}$. 
Its rows index $(\sigma_\omega, V_\nu, f_\zeta)$ and columns span four provenance classes: runtime human input (\rl{H}), bootstrap artifacts fixed prior to initialization (\rl{B}), environment/task generation (\rl{E}), and the agent (\rl{A}), with $\Gamma_{ij}(m) = 1$ when class $j$ participates in realizing meta-operator $i$ (rows admit multiple ones). Both conditions evaluate over single columns:
\begin{equation}\label{eq:role_conditions}
\mathcal{C}_1:\ \Rolemat(m)\,\mathbf{e}_{\text{\rl{H}}} = \mathbf{0},
\qquad
\mathrm{SR}:\ \Rolemat(m)\,\mathbf{e}_{\text{\rl{A}}} \neq \mathbf{0},
\end{equation}
where $\mathbf{e}_{\text{\rl{H}}}$ and $\mathbf{e}_{\text{\rl{A}}}$ select the \rl{H} and \rl{A} columns, rendering the split between \rl{B} and \rl{E} descriptive. 
For example, SimpleVLA-RL~\cite{simplevlarl} realizes all three meta-operators externally, whereas VisPlay~\cite{visplay} samples goals via an internal Questioner within $\Phi_\theta$ and derives feedback from votes over its own answers, the two matrices differing in the $\sigma_\omega$ and $V_\nu$ rows: 
{\setlength{\arraycolsep}{3pt}%
\begin{equation}\label{eq:role_example}
\Rolemat_{\text{SimpleVLA-RL}} =
\overset{\rlhead}{\begin{pmatrix} 0&1&0&0\\ 0&1&0&0\\ 0&1&0&0 \end{pmatrix}}
\quad
\Rolemat_{\text{VisPlay}} =
\overset{\rlhead}{\begin{pmatrix} 0&1&0&1\\ 0&0&0&1\\ 0&1&0&0 \end{pmatrix}}.
\end{equation}}%

The \rl{A} entry is tested under row-specific criteria. 
The goal-sampler row is marked when $\sigma_\omega(k,\mathcal{H}_k)$ conditions on the agent or its history, as with learned proposers, success-dependent curriculum rules, or self-play opponents drawn via the task argument (\cref{def:se_env}); post-hoc filtering of tied goals leaves subsequent draws unconditioned and is excluded. 
The feedback row is marked when $V_\nu$ depends on the agent, through the active policy (self-judges, consensus voting over rollouts, or reward models refitted on trajectories) or frozen snapshots (including the initial agent $\mathcal{A}_0$); group-relative normalization and per-goal monotone transforms add no dependence. 
Snapshots count because shared model lineage preserves priors, and model-based evaluators reportedly favor their own generations~\cite{panickssery2024llm}, a bias plausibly transmitted to descendant policies. 
The evolution-operator row is marked when the agent model, its replica, or an internally trained component writes update content beyond raw rollouts, such as a lesson, skill, memory entry, or relabeled goal. This row tests writer provenance, not functional dependence, every $f_\zeta$ depending trivially on the updated $\mathcal{A}_k$. 
Promptbreeder~\cite{fernando2023promptbreeder}, which rewrites its own improvement mechanism, is a special case of an \rl{A} entry in this row. 
Beyond membership, an \rl{A} entry tracking the current agent couples successive cycles through shared model parameters or history-conditioned goal rules, whereas a frozen snapshot couples none.

\pgraph{Fixed Loops and Scope}
An autonomous learning loop with self-reference is \emph{self-evolving}, whereas one lacking it is \emph{fixed-loop}, completing the membership test in \cref{def:se_agent}. 
Because the subsequent three-stage decomposition, its six factors, the two ceilings, and cross-cycle coupling are defined over autonomous learning loops, fixed-loop methods serve there as evidence characterizing a scenario's operating point, whereas scenario chapters review self-evolving loops, reporting their provenance rows and design variables. 
A method can have \rl{A} entries in several rows (\eg, proposer-solver methods realize both the goal sampler and the feedback function~\cite{evolmm,visplay}).

\begin{figure*}[!t]
\centering
\resizebox{\textwidth}{!}{%
\begin{tikzpicture}[
  font=\footnotesize,
  >={Latex[length=2mm,width=1.25mm]},
  agent/.style={rectangle, rounded corners=3pt, draw=themedark, line width=0.75pt,
    fill=themedark!5, minimum width=1.6cm, minimum height=0.95cm, align=center, inner sep=3pt},
  panel/.style={rectangle, rounded corners=4pt, line width=0.8pt,
    minimum width=4.9cm, minimum height=3.05cm},
  chip/.style={rectangle, rounded corners=2pt, line width=0.7pt, minimum width=1.55cm,
    minimum height=0.55cm, align=center, inner sep=2pt},
  micro/.style={solid},
  macro/.style={dash pattern=on 2.2pt off 1.4pt},
  aux/.style={rectangle, rounded corners=2pt, draw=themedark!52, line width=0.6pt,
    fill=white, minimum height=0.66cm, text width=2.5cm, align=center, inner sep=3pt},
  flow/.style={->, draw=themedark!82, line width=0.85pt},
  agentedge/.style={->, draw=themeblue!85!themedark, line width=1pt},
  optional/.style={->, dashed, draw=themedark!62, line width=0.7pt},
  historyflow/.style={->, draw=themedark!52, line width=0.65pt},
  lbl/.style={font=\scriptsize, color=themedark!90, fill=white, inner sep=1.5pt, align=center},
  tag/.style={font=\scriptsize, color=themedark!75}
]
\node[agent] (Ak) at (0,0) {Agent\\$\mathcal{A}_k$};
\node[agent] (Akp1) at (20.6,0) {Updated agent\\$\mathcal{A}_{k+1}$};

\newcommand{\stagepanel}[8]{%
  \node[panel, draw=#3, fill=#3!6] (#1) at (#2,0) {};
  \node[font=\footnotesize\bfseries, color=#3!60!black] at ([yshift=-0.32cm]#1.north) {#4};
  \node[font=\scriptsize, color=themedark!85] at ([yshift=-0.72cm]#1.north) {#5};
  \node[font=\small] at ([yshift=-1.25cm]#1.north) {#6};
  \node[font=\scriptsize\itshape, color=#3!60!black] at ([yshift=0.28cm]#1.south) {#7};
}
\stagepanel{EG}{4.1}{cardteal}{Stage 1 \textperiodcentered\ EG}{goal sampler $\sigma_\omega$ and rollout}{$\mathcal{Q}^{\mathrm{EG}}_k = K_k \cdot \mathbb{E}[I_{\mathcal{A}_k}]$}{speed and breadth}{}
\stagepanel{FA}{10.35}{cardpurple}{Stage 2 \textperiodcentered\ FA}{feedback function $V_\nu$}{$\mathcal{Q}^{\mathrm{FA}}_k = (2F_\nu{-}1) \cdot \mathbb{E}[L_\nu]$}{$R$-aligned direction}{}
\stagepanel{UC}{16.6}{cardcoral}{Stage 3 \textperiodcentered\ UC}{evolution operator $f_\zeta$}{$\mathcal{Q}^{\mathrm{UC}}_k = S_\zeta \cdot G_\zeta$}{retention and generalization}{}

\node[chip, micro, draw=cardteal!70!black, fill=cardteal!35] (cK) at ([xshift=-1.05cm,yshift=-0.42cm]EG.center) {$\bar{K}$ \scriptsize\textbf{x-axis}};
\node[chip, macro, draw=cardteal!70!black, fill=white] (cI) at ([xshift=1.05cm,yshift=-0.42cm]EG.center) {$I_{\mathcal{A}_k}$};
\node[chip, macro, draw=cardpurple!80!black, fill=cardpurple!30] (cF) at ([xshift=-1.05cm,yshift=-0.42cm]FA.center) {$\bar{F}$ \scriptsize\textbf{y-axis}};
\node[chip, micro, draw=cardpurple!80!black, fill=white] (cL) at ([xshift=1.05cm,yshift=-0.42cm]FA.center) {$L_\nu$};
\node[chip, macro, draw=cardcoral!80!black, fill=white] (cS) at ([xshift=-1.05cm,yshift=-0.42cm]UC.center) {$S_\zeta$};
\node[chip, macro, draw=cardcoral!80!black, fill=white] (cG) at ([xshift=1.05cm,yshift=-0.42cm]UC.center) {$G_\zeta$};

\draw[flow] (Ak.east) -- (EG.west);
\draw[flow] (EG.east) -- node[lbl, above] {$\mathcal{D}_k$} (FA.west);
\draw[flow] (FA.east) -- node[lbl, above] {$V_\nu(\mathcal{D}_k)$} (UC.west);
\draw[flow, draw=cardcoral!72!black, line width=1pt] (UC.east) -- (Akp1.west);

\draw[agentedge] (Akp1.north) -- (20.6,2.2) -- (4.1,2.2) -- (EG.north);
\node[font=\scriptsize\bfseries, color=themedark, fill=white, inner sep=2pt] at (12.35,2.2) {agent edge: the updated agent drives the next rollout};

\node[aux] (Sampler) at (4.1,-2.75) {Goal sampler $\sigma_\omega$};
\node[aux] (History) at (16.6,-2.75) {Updated history $\mathcal{H}_{k+1}$};
\draw[historyflow] (7.25,0) -- (7.25,-1.95) -- (16.6,-1.95) -- (History.north);
\draw[draw=themedark!52, line width=0.65pt] (13.45,0) -- (13.45,-1.95);
\fill[themedark!52] (13.45,-1.95) circle (1.6pt);
\draw[historyflow] (Akp1.south) |- (History.east);
\draw[optional] (History.west) -- node[lbl, above] {goal edge, adaptive sampler only} (Sampler.east);
\draw[flow, draw=cardteal!80!black] (Sampler.north) -- node[lbl, left] {$\boldsymbol{g}_{k+1}$} (EG.south);

\node[tag, font=\scriptsize\bfseries, anchor=west] (h1) at (4.6,-3.85) {Fill};
\node[chip, micro, draw=themedark!70, fill=themedark!25, minimum width=0.7cm, minimum height=0.38cm, anchor=west] (lg1) at ([xshift=0.15cm]h1.east) {};
\node[tag, anchor=west] (lt1) at (lg1.east) {ceiling used as a positioning axis};
\node[chip, micro, draw=themedark!70, fill=white, minimum width=0.7cm, minimum height=0.38cm, anchor=west] (lg2) at ([xshift=0.25cm]lt1.east) {};
\node[tag, anchor=west] (lt2) at (lg2.east) {within-regime factor};
\node[tag, font=\scriptsize\bfseries, anchor=west] (h2) at ([xshift=0.9cm]lt2.east) {Border};
\node[chip, micro, draw=themedark!70, fill=white, minimum width=0.7cm, minimum height=0.38cm, anchor=west] (lg3) at ([xshift=0.15cm]h2.east) {};
\node[tag, anchor=west] (lt3) at (lg3.east) {micro-scale};
\node[chip, macro, draw=themedark!70, fill=white, minimum width=0.7cm, minimum height=0.38cm, anchor=west] (lg4) at ([xshift=0.25cm]lt3.east) {};
\node[tag, anchor=west] at (lg4.east) {macro-scale};
\end{tikzpicture}%
}%
\caption{\textbf{The three-stage self-evolving loop and its six factors.} 
Each stage centers on one meta-operator, and its effectiveness $\mathcal{Q}_k$ is the product of two factors.
The filled ceilings $\bar{K}$ and $\bar{F}$ serve as the axes of the positioning matrix (\cref{fig:positioning_matrix}).
The goal edge, fed by the trajectories and scores of cycle $k$, reaches $\sigma_\omega$ only when it is adaptive, and extensions of the agent edge into $\sigma_\omega$, FA, or UC are not drawn.
}
\label{fig:framework_overview}
\end{figure*}
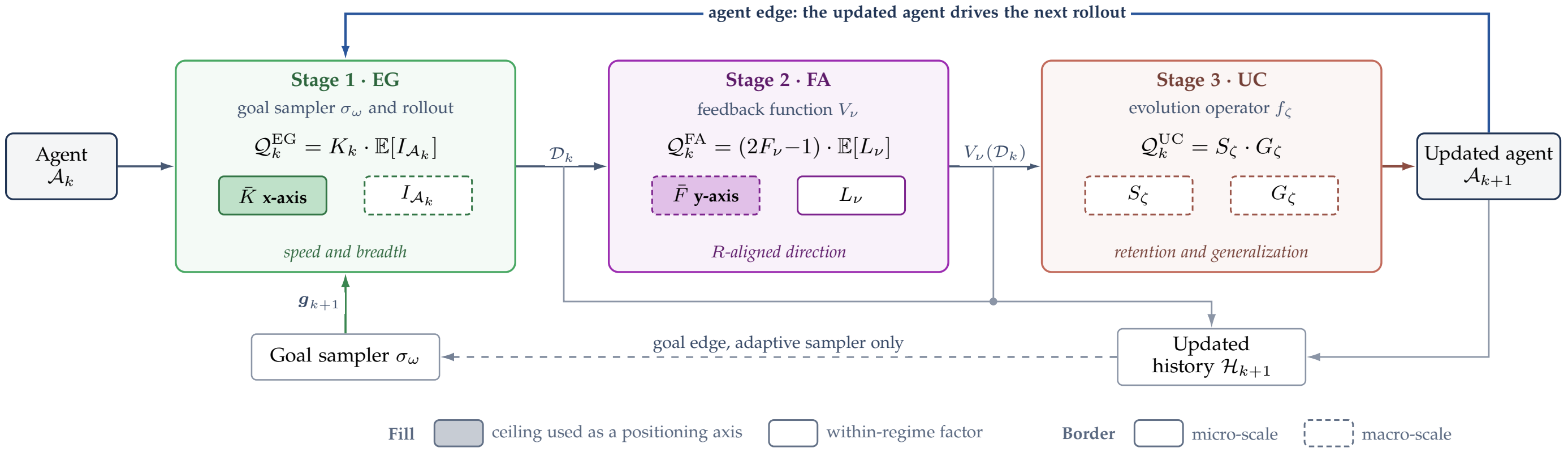

\subsection{Demarcation from Established Paradigms}
\label{sec:distinguish}

The term \emph{self-evolving} shares vocabulary with several established paradigms, and recent surveys draw its boundary at different points~\cite{fang2025comprehensive,gao2025survey,xiang2025systematic,cheng2026comprehensive,liu2025metacognitive}. 
\Cref{tab:s2_paradigm} compares adjacent paradigms; contested boundaries hinge on the \rl{A} entry. 
In the goal-sampler row, discarding tied groups within a batch adds no \rl{A} entry~\cite{simplevlarl}, whereas allocating subsequent rollouts by task success qualifies~\cite{dartgui}. 
In the evolution-operator row, the test is who writes the update: a separate fixed model (\eg, GPT-4o guiding GPT-3.5-Turbo in TextGrad) keeps the loop fixed~\cite{uimem,yuksekgonul2024textgrad}, whereas the acting model itself, as with ExpeL's insights~\cite{zhao2024expel} or STaR's rationales for failed instances~\cite{zelikman2022star}, establishes self-evolution. 
Reflexion~\cite{shinn2023reflexion}, where the acting model preserves self-reflections across episodic trials, qualifies as self-evolving, although its reflections, written for one task, improve no other task. 
Anti-forgetting mechanisms of continual learning map onto the evolution operator $f_\zeta$, whereas classical curriculum samplers such as POET~\cite{wang2019paired}, PLR and other regret-based environment designers~\cite{jiang2021prioritized,dennis2020emergent,parkerholder2022evolving}, goal-proposing teachers~\cite{sukhbaatar2018intrinsic,campero2021learning}, and expert-iteration and AlphaZero self-play~\cite{anthony2017thinking,silver2018general} remain outside our foundation-model scope.

\section{The Three-Stage Self-Evolving Loop}
\label{sec:framework}

We adopt \textit{the three-stage self-evolving loop} as our analytical framework, decomposing each cycle of an autonomous learning loop into experience generation, feedback acquisition, and update consolidation.
Fixed and self-evolving loops across all four scenarios sharing this structure, we differentiate them by where the bottleneck binds instead of by topology, treating rollout cost and feedback fidelity as bounds on viable mechanisms. 
We first formalize the three stages and their governing factors, then position the four scenarios within a positioning matrix spanned by these bounds, and trace how the stages couple across cycles.

\subsection{Loop Decomposition}\label{sec:three_stages}

As shown in \cref{fig:framework_overview}, we factor each cycle into three scenario-invariant stages, each organized around one meta-operator ($\sigma_\omega$, $V_\nu$, $f_\zeta$). 
This decomposition applies across the entire genus of autonomous learning loops and treats auxiliary operations, such as credit assignment or data filtering, as refinements within a stage, not as additional stages.
\begin{enumerate}[label=\textbf{Stage~\arabic*},labelsep=0.6em,leftmargin=3.9em,topsep=0.2em,itemsep=0.2em]
\item \textbf{Experience Generation (EG):}\; the goal sampler $\sigma_\omega$ draws the goal batch $\boldsymbol{g}_k$, and the current agent $\mathcal{A}_k$ rolls out in $\mathcal{P}_{g}$ for each goal $g \in \boldsymbol{g}_k$, producing the trajectory batch $\mathcal{D}_k$.
Through the quantity and informativeness of $\mathcal{D}_k$, the only trajectories FA and UC act on, EG bounds how fast and how broadly the loop can evolve (\emph{speed and breadth}).
\item \textbf{Feedback Acquisition (FA):}\; the feedback function $V_\nu$ maps $\mathcal{D}_k$, together with the agent history $\mathcal{A}_{0:k}$ when $V_\nu$ reads the agent, to scores, preferences, or critiques sourced from the environment, the bootstrap, or the agent itself.
FA is the loop's only source of information about the idealized task reward $R$ (\emph{the $R$-aligned direction}).
\item \textbf{Update Consolidation (UC):}\; the evolution operator $f_\zeta$ computes an update from $(\mathcal{A}_k, \mathcal{H}_k, \mathcal{D}_k, V_\nu(\mathcal{D}_k))$ and writes it into at least one persistent component of $\mathcal{A}_k$, producing the next-cycle agent $\mathcal{A}_{k+1}$.
UC determines whether an update persists across later cycles and generalizes beyond the goals that produced it (\emph{retention and generalization}).
\end{enumerate}
The six factors originate at the two scales of \cref{fig:2_loop}.
The per-episode cost $c(\tau)$ governing $\bar{K}$ and action localization $L_\nu$ arise within the micro-scale interaction loop, whereas information gain, evaluator fidelity $F_\nu$, and consolidation factors ($S_\zeta$, $G_\zeta$) emerge from the macro-scale learning loop, the positioning matrix thereby pairing a micro-scale axis with a macro-scale one.

\begin{table*}[!t]
\centering
\footnotesize
{\renewcommand\tabularxcolumn[1]{m{#1}}%
\caption{\textbf{Asymmetric substitutability of $\mathcal{Q}_k^{\mathrm{EG}}$ across four scenarios.}
Profiling experience generation across the quantity and quality factors from \cref{eq:effective_experience}. 
The bottom row highlights the binding bottleneck for each scenario, showing where cheap rollouts saturate and where costly environments constrain scale. 
}
\label{tab:stage1_dims}
\vspace{-2mm}
\renewcommand{\arraystretch}{1}
\rowcolors{3}{themelight}{white}
\begin{tabularx}{\textwidth}{%
>{\raggedright\arraybackslash}m{4.4cm}
>{\raggedright\arraybackslash}X
>{\raggedright\arraybackslash}X
>{\raggedright\arraybackslash}X
>{\raggedright\arraybackslash}X}
\toprule
\rowcolor{gray!25}
& \textbf{MR} & \textbf{MG} & \textbf{GW} & \textbf{EM} \\
\midrule
\rowcolor{white}
\multicolumn{5}{@{}l}{\ding{172} \textbf{Quantity factor $K_k$:}\; \textit{how many rollouts can a cycle afford?}} \\
\;\,\ding{60} Cost of one rollout $c(\tau)$
& one decoding pass
& one sampling pass or tool chain
& per-step environment latency
& per-step latency plus episode reset \\
\;\,\ding{60} Action recoverability
& none needed (no side effects)
& none needed (no side effects)
& partial (DOM and account-level writes)
& largely irreversible (contact, wear) \\
\;\,\ding{60} Reported wall-clock cost
& not reported per rollout (one pass)
& not reported per rollout (one pass)
& 2.5--4.0\,s per step~\cite{uitars2}, episodes of tens of minutes~\cite{dartgui}, 5--10\,min environment re-initialization per training step~\cite{acurl}
& about 20\,h per real-world experiment with human resets~\cite{siefm}, 2.6--16.9\,s per judge query~\cite{findcantdo} \\
\;\,\ding{60} Reported group size $K_k$
& 4$-$10~\cite{vzero,evolmm,activezero,rise,evovid}
& 3$-$16~\cite{unirl,iris,solace}
& 8$-$64~\cite{mobileguirl,dartgui,mobilerl,zerogui}
& 8~\cite{simplevlarl} \\
\multicolumn{5}{@{}l}{\ding{173} \textbf{Quality factor $\mathbb{E}[I_{\mathcal{A}_k}]$:}\; \textit{how informative is each rollout?}} \\
\;\,\ding{60} Goal-sampling strategy
& task synthesis by a co-trained proposer (adaptive)
& random sampling from a fixed prompt pool (stationary)
& fixed task sets reweighted by success records (adaptive)
& fixed task suites, success-conditioned or self-proposed practice (adaptive) \\
\;\,\ding{60} Interaction mode
& single action, sometimes tool-integrated
& single output or a short tool chain
& multi-step screen states, partially observable
& multi-step physical states, partially observable
\\
\midrule
\rowcolor{cardpurple!10!white}
\textbf{Primary bottleneck of $\mathcal{Q}_k^{\mathrm{EG}}$}
& \textbf{Quality, at saturation}
& \textbf{Quality}
& \textbf{Quantity}
& \textbf{Both, on hardware} \\
\bottomrule
\end{tabularx}
}
\end{table*}

\subsection{Stage 1: Experience Generation (EG)}\label{sec:stage1}

EG produces the trajectory batch $\mathcal{D}_k$ of each cycle by rolling out the current agent $\mathcal{A}_k$ in the POMDPs $\mathcal{P}_g$ of \cref{def:se_env}. 
For each goal $g$ in the batch $\boldsymbol{g}_k$ drawn from $\sigma_\omega(k,\mathcal{H}_k)$, $\mathcal{A}_k$ draws $K_k$ rollouts from $p^{\pi_{\mathcal{A}_k}}_{g}$ and pools them into $\mathcal{D}_k$.
EG must supply experience of sufficient \emph{scale} and \emph{informativeness} under finite compute, wall-clock, and safety budgets, a constraint on quantity and one on quality.
We summarize this bottleneck through an effective-experience decomposition,
\begin{equation}
\mathcal{Q}_k^{\mathrm{EG}}
\;:=\;
K_k
\cdot
\mathbb{E}_{g \sim \sigma_\omega(k,\mathcal{H}_k)}
\!\left[
I_{\mathcal{A}_k}(g)
\right],
\label{eq:effective_experience}
\end{equation}
where $K_k$ is the number of per-goal rollouts in cycle $k$, and $I_{\mathcal{A}_k}(g) \!:=\! \mathrm{Var}_{\tau \sim p^{\pi}_{g}}[\mathrm{ret}_g(\tau)]$ is the outcome variance of $g$. Here, $\mathrm{ret}_g(\tau) \!=\! \sum_{t} \gamma^{t} R(s_t,u_t,g)$ denotes the return of $\tau$, and $p^{\pi}_{g}$ abbreviates $p^{\pi_{\mathcal{A}_k}}_{g}$. 
This variance, a goal-level proxy for the learnable information gain~\cite{wu2025selfplay}, vanishes on goals that $\mathcal{A}_k$ has mastered or never solves, where GRPO-style group-relative updates receive zero advantage under an $R$-aligned evaluator. 
Under a binary terminal reward with $\gamma\!=\!1$, it reduces to $p_g(1-p_g)$, which difficulty-targeting rewards and filters track via an estimated success probability $p_g$. 
Defined on the idealized reward $R$ of \cref{eq:objective}, the proxy renders $\mathcal{Q}_k^{\mathrm{EG}}$ diagnostic, and it exposes two failure modes. As $K_k$ shrinks, too few per-goal rollouts remain to contrast success with failure. As $\mathbb{E}[I_{\mathcal{A}_k}]$ approaches zero, the sampled goals are mastered or unreachable. 
Concretely, $K_k$ is bounded by per-rollout \emph{interaction cost} and \emph{action recoverability}, while $\mathbb{E}[I_{\mathcal{A}_k}]$ is governed by the \emph{diversity and information gain} targeted by $\sigma_\omega$ and by the \emph{interaction mode} realizing $p^{\pi_{\mathcal{A}_k}}_{g}$ (\cref{tab:stage1_dims}).

\pgraphhlY{Quantity Constraint}\; Interaction cost and action recoverability govern $K_k$ via a single per-rollout cost $c(\tau)$, eliminating the need for a separate term for actions that resist reverting, resetting, or safe repetition. 
Specifically, $c(\tau)$ absorbs the time, compute, synchronization, and reset cost of one rollout, with the reset term penalizing both environment restoration and expected damage from failed trajectories. 
Under a fixed per-goal budget $B$, the rollout count satisfies the affordability ceiling $K_k \!\le\! \bar{K} \!:=\! B/\mathbb{E}_\tau[c(\tau)]$, where the expectation runs over episodes of the target task family at the horizon cap $T_{\max}$, not over the rollouts of a particular policy, which lets $\bar{K}$ reflect task structure and not policy choice. 
When $c(\tau)$ is small, group-sampling operators such as self-consistency~\cite{wang2023selfconsistency,chen2023universal} or GRPO~\cite{shao2024deepseekmath,guo2025deepseekr1,huang2025vision} sample a group of rollouts per goal at negligible marginal cost.
As $c(\tau)$ grows, the bottleneck shifts from theoretical execution feasibility to the budget of exploratory failures the system can absorb.

\pgraphhlY{Quality Constraint}\; Even with a large $K_k$, a trajectory set is unproductive if its goals are already mastered or still unreachable for $\mathcal{A}_k$, leaving EG viable only while $\mathbb{E}[I_{\mathcal{A}_k}] \!>\! 0$. For instance, self-play~\cite{wu2025selfplay} stops improving once its synthetic pipeline no longer supplies learnable information gain.
\emph{Diversity and information gain} determines whether $\sigma_\omega$ draws goals with nonzero $I_{\mathcal{A}_k}$. Sampling schemes escalate in intervention from random sampling~\cite{shao2024deepseekmath} through curriculum mining over past failures~\cite{uigenie,mobilerl} and active task selection~\cite{diao2023active,du2023guiding} to environment synthesis~\cite{wang2024robogen,wang2024gensim,guigenesis,webgym,agtworld}. 
Random sampling keeps $\sigma_\omega$ stationary, whereas curriculum mining and active selection make it adaptive. Environment synthesis acts through the task argument $g$ of \cref{def:se_env} and is adaptive only when the generator reads the agent's record. 
Only an adaptive sampler conditions the next goal draw on the agent's record and places an \rl{A} entry in the goal-sampler row. 
\emph{Interaction mode} fixes what evidence each trajectory can expose. Single-pass generation and tool-augmented reasoning yield single-action episodes, whereas digital and embodied environments expose multi-step state transitions whose informativeness depends on platform dynamics, reset conditions, and observability.

\pgraphhlP{Asymmetric Substitutability}\; The product in \cref{eq:effective_experience} enables mutual compensation between its two factors, locating the cross-scenario asymmetry in the cost of raising each factor. 
Raising $K_k$ offsets low information gain only when rollouts are cheap, parallelizable, and recoverable and feedback retains contrast as group size grows. Consensus rewards losing contrast beyond small groups~\cite{evolmm}, the loop elsewhere must instead raise $\mathbb{E}[I_{\mathcal{A}_k}]$ via curriculum mining~\cite{dartgui,mobilerl}, active exploration~\cite{visplay,evovla}, demonstrations at initialization~\cite{siefm}, or synthetic environments~\cite{mmzero,wang2024robogen,wang2024gensim}. 
Equal values of $K_k$ therefore do not imply identical bottlenecks. 
Multimodal generation and GUI/Web navigation report overlapping group sizes (\cref{tab:stage1_dims}), yet generation samples cheap, reversible passes below $\bar{K}$ and is bounded by $\mathbb{E}[I_{\mathcal{A}_k}]$ over a fixed prompt pool, whereas navigation incurs latency-bound episodes and is bounded by $\bar{K}$ itself. 
Cheap but uninformative rollouts saturate the loop, whereas informative but costly rollouts throttle its scaling.

\begin{table*}[!t]
\centering
\footnotesize
{\renewcommand\tabularxcolumn[1]{m{#1}}%
\caption{\textbf{Asymmetric vulnerability of $\mathcal{Q}_k^{\mathrm{FA}}$ across four scenarios.}
Profiling feedback acquisition across the fidelity and localization factors from \cref{eq:effective_feedback}, with signal sources tagged by provenance.
The bottom row identifies the binding bottleneck for each scenario, showing where gameable signals misdirect the loop and where coarse signals under-constrain it.
}
\label{tab:stage2_dims}
\vspace{-2mm}
\renewcommand{\arraystretch}{1}
\renewcommand{\tabularxcolumn}[1]{m{#1}}
\rowcolors{3}{themelight}{white}
\begin{tabularx}{\textwidth}{%
>{\raggedright\arraybackslash}m{4.4cm}
>{\raggedright\arraybackslash}X
>{\raggedright\arraybackslash}X
>{\raggedright\arraybackslash}X
>{\raggedright\arraybackslash}X}
\toprule
\rowcolor{gray!25}
& \textbf{MR} & \textbf{MG} & \textbf{GW} & \textbf{EM} \\
\midrule
\rowcolor{white}
\multicolumn{5}{@{}l}{\ding{172} \textbf{Fidelity factor $F_\nu$:}\; \textit{how faithfully does $V_\nu(\mathcal{D}_k)$ order trajectories by $\mathrm{ret}_g$?}} \\
\;\,\ding{60} Signal source (provenance)
& majority vote \rl{A}, self-judgment \rl{A}, computed answers \rl{E}
& self-rewards \rl{A}, same-model judges \rl{A}, external judges \rl{B}
& task checkers \rl{B}, VLM judges \rl{B}, co-evolving RMs \rl{A}, same-model judges \rl{A}
& episode outcomes \rl{E}, progress heads and same-model judges \rl{A}, VLM judges \rl{B} \\
\;\,\ding{60} Channel coverage
& full (discrete answer)
& often perceptual only
& full on screen, partial for backend state
& full in simulation, visual only on hardware \\
\;\,\ding{60} Gaming resistance under $f_\zeta$
& high for computed answers, untested for votes
& low~\cite{solace,iris}
& untested beyond downstream success~\cite{mobilejudgebench}
& low without judge oversight~\cite{dern2026vlmjudges} \\
\rowcolor{white}
\multicolumn{5}{@{}l}{\ding{173} \textbf{Localization factor $L_\nu$:}\; \textit{how finely can credit be apportioned along $\tau$?}} \\
\;\,\ding{60} Temporal granularity
& episode $=$ step ($T\!=\!1$ without tool calls)
& episode ($T\!=\!1$ if single-pass)
& episode (judges), step (co-evolving RMs)
& episode (outcomes), step (progress heads) \\
\;\,\ding{60} Credit-assignment cost
& none ($T\!=\!1$)
& none if single-pass, low otherwise
& co-trained step RMs~\cite{uigenie,magicguirms}
& demonstration-trained progress heads~\cite{siefm,scvla} \\
\midrule
\rowcolor{cardpurple!10!white}
\textbf{Primary bottleneck of $\mathcal{Q}_k^{\mathrm{FA}}$}
& \textbf{Realized $F_\nu^{(k)}$}
& \textbf{Fidelity ceiling $\bar{F}$}
& \textbf{Localization}
& \textbf{Both, on hardware} \\
\bottomrule
\end{tabularx}
}
\end{table*}

\subsection{Stage 2: Feedback Acquisition (FA)}
\label{sec:stage2}

FA maps the trajectory batch $\mathcal{D}_k$ of each cycle to the feedback $V_\nu(\mathcal{D}_k) \!\in\! (\mathbb{Y}^{\mathrm{mm}})^{*}$, the scores, preferences, critiques, or structured judgments that direct $f_\zeta$.
Any scoring, ranking, filtering, or best-of-$N$ selection by $V_\nu$ belongs to FA even when interleaved with generation, whereas $V_\nu$-independent filtering such as deduplication remains in EG.
FA must supply feedback that is \emph{aligned} with the intended objective $R$ and \emph{localized} at an actionable grain without recurring human supervision ($\mathcal{C}_1$), a constraint on fidelity and one on localization.
We summarize this bottleneck through an effective-feedback decomposition,
\begin{equation}
\mathcal{Q}_k^{\mathrm{FA}}
\;:=\;
\bigl(2F_\nu(\mathcal{D}_k;R,\boldsymbol{g}_k) - 1\bigr)
\cdot
\mathbb{E}_{\tau \sim \widehat{p}_{\mathcal{D}_k}}
\!\left[
L_\nu(\tau)
\right],
\label{eq:effective_feedback}
\end{equation}
where $\widehat{p}_{\mathcal{D}_k}$ is the empirical measure on $\mathcal{D}_k$, $F_\nu \!\in\! [0,1]$ is the rate at which the ordering $V_\nu(\mathcal{D}_k)$ induces on same-goal trajectory pairs agrees with their ordering by $\mathrm{ret}_g$, and $L_\nu(\tau) \!\in\! (0,1]$ is the fraction of environment actions along $\tau$ to which $V_\nu$ can assign credit. 
Here, pairs tied under either ordering are excluded, and $F_\nu \!:=\! 1/2$ if none remains.
This pairwise form covers scalar, pairwise, and ranked feedback alike, and $2F_\nu - 1$ is the Goodman-Kruskal rank correlation of the two orderings, zero for a random evaluator and negative for a reversing one.
Like $I_{\mathcal{A}_k}$, $F_\nu$ is defined on $R$, which makes $\mathcal{Q}_k^{\mathrm{FA}}$ diagnostic, and it exposes two failure modes. As $F_\nu$ falls toward $1/2$, $f_\zeta$ follows an uninformative proxy, and below $1/2$ it optimizes against $R$. As $L_\nu$ shrinks, a correct signal becomes too coarse to guide improvement.
Concretely, $F_\nu$ is bounded by the \emph{signal source} and \emph{gaming resistance} of $V_\nu$, while $L_\nu$ is governed by the \emph{temporal granularity} and \emph{credit-assignment cost} of its scores (\cref{tab:stage2_dims}).

\pgraphhlY{Fidelity Constraint}\; Signal source and gaming resistance act on $F_\nu$ at different levels: the signal source fixes the evaluator configurations $\Xi_V$ a scenario admits, hence the attainable-fidelity ceiling $\bar{F} \!:=\! \sup_{\nu \in \Xi_V} F_\nu$, whereas gaming resistance governs how much of it survives repeated $f_\zeta$ updates.
Where a closed-form success condition or a discrete answer space exists, deterministic instruments such as answers computed by task generation~\cite{pvse,cwi}, unit tests~\cite{coderl,rltf}, and physical-constraint rewards~\cite{evovla,scvla,rmrl} sustain high $F_\nu$ at modest cost.
Lacking both, evaluators rely on self-rewards~\cite{solace,suder,ccselfreward,srum}, reward models~\cite{uigenie,magicguirms,mstar,wang2024mathshepherd,cui2025prime,lrm,sarm2}, or VLM judges~\cite{digirl,evovla,evolvevla,smartsnap}, and gaming resistance binds~\cite{solace,pan2024feedback}.
\emph{Channel coverage} further caps $F_\nu$, an instrument leaving unobserved channels of $\mathcal{U}$ unconstrained~\cite{solace,evovla,scvla}.
Self-rewards and co-trained reward models read $\mathcal{A}_k$ and place an \rl{A} entry in the feedback row, and their realized fidelity $F_\nu^{(k)}$ therefore drifts with the policy they score; all others enter as \rl{B} or \rl{E}.
Reported audits measure label or verdict accuracy $a$, not the pairwise agreement that defines $F_\nu$. Under a binary terminal reward with verdict errors independent across trajectories and symmetric across outcomes, a success-failure pair is ordered correctly with probability $a^2$ and reversed with probability $(1-a)^2$, giving $F_\nu = a^2/(a^2+(1-a)^2)$, which exceeds $a$ whenever $a>1/2$. A reported accuracy thus bounds $F_\nu$ from below only under these assumptions, and errors correlated within a goal void the bound.
In multimodal reasoning, program-computed answers stay correct on 94.4\% of retained questions~\cite{pvse}, yet self-generated pseudo-labels degrade across rounds~\cite{visplay,videozero}, consistent with self-referential evaluators drifting below a high $\bar{F}$, although rising question difficulty confounds these declines.

\begin{table*}[!t]
\centering
\footnotesize
{\renewcommand\tabularxcolumn[1]{m{#1}}%
\caption{\textbf{Within-regime specialization of $\mathcal{Q}_k^{\mathrm{UC}}$ across four scenarios.}
Profiling update consolidation across the retention and generalization factors from \cref{eq:effective_persistence}, both lacking a scenario-fixed ceiling. 
The bottom row identifies the binding bottleneck for each scenario, showing where later cycles erode gains and where updates stay within sampled goals. 
}
\label{tab:stage3_dims}
\vspace{-2mm}
\renewcommand{\arraystretch}{1}
\renewcommand{\tabularxcolumn}[1]{m{#1}}
\rowcolors{3}{themelight}{white}
\begin{tabularx}{\textwidth}{%
>{\raggedright\arraybackslash}m{4.4cm}
>{\raggedright\arraybackslash}X
>{\raggedright\arraybackslash}X
>{\raggedright\arraybackslash}X
>{\raggedright\arraybackslash}X}
\toprule
\rowcolor{gray!25}
& \textbf{MR} & \textbf{MG} & \textbf{GW} & \textbf{EM} \\
\midrule
\rowcolor{white}
\multicolumn{5}{@{}l}{\ding{172} \textbf{Retention factor $S_\zeta$:}\; \textit{does the cycle-$k$ gain survive subsequent updates?}} \\
\;\,\ding{60} Dominant substrate
& $\Phi_\theta$, except training-free methods
& $\Phi_\theta$, or $\mathcal{M}$, $\mathcal{T}$, $\mathcal{I}$ if training-free
& $\Phi_\theta$, or $\mathcal{M}$, $\mathcal{T}$ if training-free
& $\Phi_\theta$, or $\mathcal{M}$, $\mathcal{T}$, $\mathcal{I}$ \\
\;\,\ding{60} Iteration robustness under $f_\zeta$
& untested beyond a few cycles
& compounding hacking within single runs~\cite{solace,iris}
& sustained by trajectory replay~\cite{digirl,mobilerl}, no forgetting on held-out environments~\cite{acurl}
& fixed-loop continual RL only~\cite{crlvla,simplerecipe} \\
\;\,\ding{60} Dominant failure mode
& long-cycle drift
& compounding reward hacking
& distribution shift
& forgetting and distribution shift \\
\rowcolor{white}
\multicolumn{5}{@{}l}{\ding{173} \textbf{Generalization factor $G_\zeta$:}\; \textit{does the gain transfer beyond $\mathcal{G}_k$?}} \\
\;\,\ding{60} Transfer locus
& parameters (any goal)
& parameters, or memory and skills (on use)
& parameters, or memory and skills (on use)
& parameters, or skills and harnesses (on use) \\
\;\,\ding{60} Experience coverage in $\mathcal{D}_k$
& tasks synthesized over unlabeled media
& a fixed prompt pool
& the evaluation task set or stream~\cite{mobileagente,webcoach,uimem,traces2skills}
& one simulator suite \\
\;\,\ding{60} Achieved generalization scope
& cross-task~\cite{vzero,visplay,evolmm}
& in-distribution, partly cross-benchmark~\cite{iris,sidiffagent}
& in-distribution, one platform
& in-distribution, rare sim-to-real~\cite{embodiedrsi,evovla} \\
\midrule
\rowcolor{cardpurple!10!white}
\textbf{Primary bottleneck of $\mathcal{Q}_k^{\mathrm{UC}}$}
& \textbf{Retention}
& \textbf{Retention}
& \textbf{Generalization}
& \textbf{Generalization} \\
\bottomrule
\end{tabularx}
}
\end{table*}

\pgraphhlY{Localization Constraint}\; Even with high $F_\nu$, a trajectory set is directionally unproductive if $L_\nu$ covers a vanishing fraction of environment actions, and FA accordingly remains viable only while $\mathbb{E}_\tau[L_\nu(\tau)]$ suffices for the grain $f_\zeta$ requires.
\emph{Temporal granularity} fixes the scoring unit along $\tau$: the terminal outcome for episode-level scores, the action $u_t$ for step-level scores, and the output $y_t$ decoded into $u_t$ for token-level scores.
Because $L_\nu$ counts environment actions, a single-action episode has $L_\nu \!=\! 1$, and credit within a long reasoning chain $y_t$ lies outside $L_\nu$ even where process reward models assign it~\cite{wang2024mathshepherd}.
Episode-level feedback suffices when the task reduces to one discrete outcome, step-level process supervision becomes necessary as horizons lengthen and terminal success grows too sparse~\cite{wang2024mathshepherd,lai2024step,webagentr1,evovla,densegrpo}, and token-level credit remains rare outside independently checkable reasoning chains~\cite{ireasoner}.
\emph{Credit-assignment cost} determines how much of that grain a cycle can afford, step-level scores requiring a process model or per-step estimates whose cost grows with horizon length~\cite{wang2024mathshepherd}.
Granularity is thus an FA choice that reaches $f_\zeta$ only through its signal.

\pgraphhlP{Asymmetric Vulnerability}\; The product in \cref{eq:effective_feedback} admits only conditional, one-directional compensation between its two factors.
Higher $F_\nu$ from cross-model verification or understanding-generation duality offsets moderate localization loss only when an unbiased anchor exists~\cite{suder,unirl}, whereas finer $L_\nu$ yields no directional gain once $F_\nu$ is gameable~\cite{solace,pan2024feedback}.
Equal ceilings $\bar{F}$ therefore do not imply identical bottlenecks.
Multimodal reasoning and GUI/Web navigation both admit a high $\bar{F}$ (\cref{tab:stage2_dims}), through program-computed answers~\cite{pvse} and per-task state scripts in sandboxed environments~\cite{dartgui,mobilerl}, yet reasoning fixes $L_\nu \!=\! 1$ and is bounded by the realized $F_\nu^{(k)}$ of its consensus and self-judged signals~\cite{visplay,videozero,ttrv}, while navigation scores whole long-horizon episodes and is bounded by $L_\nu$~\cite{seagent,uigenie,magicguirms}.
Multimodal generation instead faces a low $\bar{F}$ on open-ended prompts, where self-rewards and perception-only instruments are readily gamed~\cite{solace,suder}, and embodied manipulation loses both factors on hardware~\cite{videoagent,soar,siefm,scvla}.
Fidelity-bounded signals thus misdirect the loop and localization-bounded ones under-constrain it, making $\bar{F}$ a positioning axis and $L_\nu$ a within-regime variable.

\subsection{Stage 3: Update Consolidation (UC)}
\label{sec:stage3}

UC writes the update that $f_\zeta$ computes from $\mathcal{D}_k$ and $V_\nu(\mathcal{D}_k)$ into one or more persistent components of $\mathcal{A}_k \!=\! (\Phi_{\theta_k}, \mathcal{M}_k, \mathcal{T}_k, \mathcal{I}_k)$, as $\mathcal{C}_3$ requires.
Writes to the episode-scoped buffer $\mathcal{M}_{\mathrm{short}}$ through $\Psi$ fall outside UC, and $V_\nu$-based selection of what to write remains in FA.
UC must write updates that \emph{persist} undegraded across subsequent cycles and \emph{generalize} beyond the cycle-$k$ goals, a constraint on retention and one on generalization.
We summarize this bottleneck through an effective-consolidation decomposition,
\begin{equation}
\mathcal{Q}_k^{\mathrm{UC}}
\;:=\;
S_\zeta\bigl(\mathcal{A}_{k:k+j+1}\bigr)
\cdot
G_\zeta\bigl(\mathcal{A}_{k+1};\,\mu_{\mathcal{G}}\bigr),
\label{eq:effective_persistence}
\end{equation}
where $\Delta_k(g) \!=\! J(\mathcal{A}_{k+1};g) \!-\! J(\mathcal{A}_k;g)$ is the cycle-$k$ gain on goal $g$, $\mathcal{G}_k$ is the support of $\boldsymbol{g}_k$, and $\mu_{\mathcal{G}}$ is the fixed evaluation distribution of \cref{eq:eval_seq}.
Here, $S_\zeta \!\in\! [0,1]$ is the fraction of the cycle-$k$ gain on $\mathcal{G}_k$ that survives $j$ further cycles, $\min\{1,\, \max\{0,\, \mathbb{E}_{g\sim\mu_{\mathcal{G}}|\mathcal{G}_k}[J(\mathcal{A}_{k+1+j};g) - J(\mathcal{A}_k;g)] \,/\, \mathbb{E}_{g\sim\mu_{\mathcal{G}}|\mathcal{G}_k}[\Delta_k(g)]\}\}$ for a positive denominator, and $G_\zeta \!\in\! [0,1]$ is the $\mu_{\mathcal{G}}$-measure of goals outside $\mathcal{G}_k$ with $\Delta_k(g) \!>\! 0$, renormalized to the complement of $\mathcal{G}_k$.
The gain and the two factors correspond to the adaptivity, retention, and generalization goals proposed for self-evolving agents~\cite{gao2025survey}, and $S_\zeta$ normalizes by the cycle-$k$ gain the change on earlier goals that backward transfer records as a signed difference~\cite{lopez2017gradient}.
Like $F_\nu$, both factors are defined through $J$ and hence on $R$, leaving $\mathcal{Q}_k^{\mathrm{UC}}$ diagnostic as well, and it exposes two failure modes. As $S_\zeta$ falls toward zero, later cycles erase or distort the cycle-$k$ gain. As $G_\zeta$ falls toward zero, the update stays confined to $\mathcal{G}_k$.
Concretely, $S_\zeta$ is bounded by the \emph{substrate} that $f_\zeta$ writes and its \emph{iteration robustness}, while $G_\zeta$ is governed by the \emph{transfer locus} of that substrate and the \emph{experience coverage} of $\mathcal{D}_k$ (\cref{tab:stage3_dims}).

\pgraphhlY{Retention Constraint}\; Substrate and iteration robustness act on $S_\zeta$ at different levels: the substrate fixes how far a single write can interfere with earlier gains, whereas iteration robustness governs whether those gains survive repeated $f_\zeta$ applications.
The substrate is either $\Phi_\theta$, written by GRPO~\cite{shao2024deepseekmath}, PPO~\cite{schulman2017proximal}, or DPO~\cite{rafailov2023direct}, or a non-parametric component such as memory~\cite{packer2024memgpt,park2023generative,shinn2023reflexion,m2web,memw}, tools~\cite{wang2023voyager,cai2023large}, or prompts~\cite{zhou2023large,yang2024opro,khattab2024dspy}.
Three documented failure modes threaten iteration robustness: compounding reward hacking through accumulated Stage~2 bias within single runs~\cite{solace,iris}, distribution shift as the rollout law leaves the calibration region of $V_\nu$~\cite{digirl}, and catastrophic forgetting through interference in $\Phi_\theta$~\cite{crlvla,simplerecipe}.
Hacking and shift enter across the cycle boundary, whereas forgetting arises inside $f_\zeta$ and yields to experience replay, parameter isolation, and regularized updates~\cite{crlvla,simplerecipe,rebuffi2017icarl,chaudhry2019efficient,buzzega2020dark,li2018learning}.
Within one task family, the three modes can compound into \emph{long-cycle drift}, a slow multi-source degradation that terminal-only reporting leaves undetected and per-round scores on a fixed benchmark suite reveal only as non-monotone curves.

\pgraphhlY{Generalization Constraint}\; Even with high $S_\zeta$, a persisted update confers no broader benefit if it improves only the goals in $\mathcal{G}_k$, UC thus remaining useful only while $\mathcal{A}_{k+1}$ improves on a non-trivial share of the goals outside $\mathcal{G}_k$.
\emph{Transfer locus} fixes where a write takes effect: on any goal that activates the affected weights for parameters, upon retrieval for memory, upon invocation for tools, and within the templating scope for prompts.
\emph{Experience coverage} determines the subspace of $\mathcal{G}$ to which the update can generalize, and reported gains stay mostly within the support of $\mathcal{D}_k$, as the fixed loops UI-Voyager~\cite{uivoyager}, UI-Mem~\cite{uimem}, and an online skill library~\cite{traces2skills} train and evaluate on the same task sets.
Achieved scope rises from in-distribution transfer to unsampled goals of the same regime, through cross-task transfer to neighboring goal subspaces, to cross-domain transfer to other GUI platforms or real hardware, the last reached by few methods on small trial counts~\cite{embodiedrsi,evovla}.
The substrate thus upper-bounds transfer, and the coverage of $\mathcal{D}_k$ sets how much of it a cycle realizes.

\pgraphhlP{Within-Regime Specialization}\; Neither factor of \cref{eq:effective_persistence} has a scenario-fixed ceiling, and UC therefore specializes within each quadrant, not along an axis.
$S_\zeta$ binds when FA is gameable or EG is non-recoverable, and $G_\zeta$ binds when $\mathcal{D}_k$ covers too little of $\mathcal{G}$.
Shared substrates therefore do not imply identical bottlenecks.
Multimodal reasoning and GUI/Web navigation share $\Phi_\theta$ as the dominant substrate (\cref{tab:stage3_dims}), yet reasoning sustains intensive parametric writes under low-cost EG and rule-checkable FA and is bounded by long-cycle drift~\cite{visplay,evolmm,pvse}, while navigation stabilizes its parametric writes by trajectory replay~\cite{digirl,mobilerl} and, in both parametric and training-free forms, is bounded by transfer that stays within one platform.
Multimodal generation shares the retention bottleneck through compounding hacking of its parametric writes~\cite{solace,iris}, and embodied manipulation shares the generalization bottleneck, writing to $\Phi_\theta$ and to skills, memory, and harnesses~\cite{physmem,zetta,skillharness,embodiedrsi} within single simulator suites.
Few methods measure either factor across cycles, and retention on earlier-round goals is measured in a single self-evolving loop, RSI in UMMs~\cite{rsiumm}, besides fixed-loop continual reinforcement learning~\cite{crlvla,simplerecipe}.
UC thus inherits its bottleneck from the first two stages and refines a method within its quadrant without moving it.

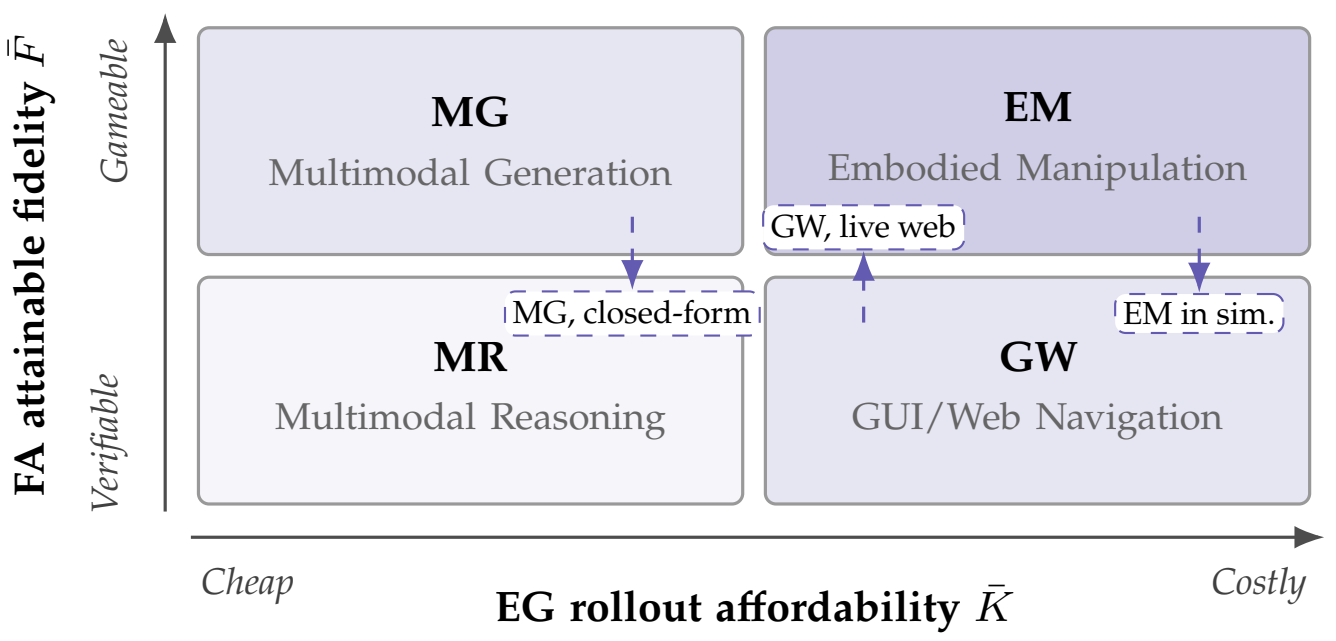
\begin{figure}[!t]
\centering
\begin{tikzpicture}[
  font=\footnotesize, >=Latex,
  cell/.style={minimum width=3.60cm, minimum height=1.50cm,
               text width=3.30cm, align=center, inner sep=3pt,
               rounded corners=2pt, fill=none,
               draw=black!40, line width=0.6pt},
  axlab/.style={font=\footnotesize\bfseries},
  pole/.style={font=\fontsize{7}{8}\selectfont, text=black!75}
]
\definecolor{posaccent}{RGB}{99,91,177}

\fill[posaccent,opacity=0.06,rounded corners=2pt] (0.10,0.10) rectangle (3.70,1.60);
\fill[posaccent,opacity=0.16,rounded corners=2pt] (3.85,0.10) rectangle (7.45,1.60);
\fill[posaccent,opacity=0.16,rounded corners=2pt] (0.10,1.75) rectangle (3.70,3.25);
\fill[posaccent,opacity=0.30,rounded corners=2pt] (3.85,1.75) rectangle (7.45,3.25);

\node[cell] at (1.90,0.85) {{\bfseries MR}\\[2pt]{\scriptsize\color{black!60} Multimodal Reasoning}};
\node[cell] at (5.65,0.85) {{\bfseries GW}\\[2pt]{\scriptsize\color{black!60} GUI/Web Navigation}};
\node[cell] at (1.90,2.50) {{\bfseries MG}\\[2pt]{\scriptsize\color{black!60} Multimodal Generation}};
\node[cell] at (5.65,2.50) {{\bfseries EM}\\[2pt]{\scriptsize\color{black!60} Embodied Manipulation}};
\node[draw=posaccent,dashed,rounded corners=2pt,fill=white,inner sep=1.5pt,font=\fontsize{6}{7}\selectfont] (emsim) at (6.72,1.36) {EM in sim.};
\draw[->,dashed,posaccent,line width=0.6pt] (6.72,2.0) -- (emsim.north);
\node[draw=posaccent,dashed,rounded corners=2pt,fill=white,inner sep=1.5pt,font=\fontsize{6}{7}\selectfont] (mgcf) at (2.97,1.36) {MG, closed-form};
\draw[->,dashed,posaccent,line width=0.6pt] (2.97,2.0) -- (mgcf.north);
\node[draw=posaccent,dashed,rounded corners=2pt,fill=white,inner sep=1.5pt,font=\fontsize{6}{7}\selectfont] (gwlive) at (4.50,1.93) {GW, live web};
\draw[->,dashed,posaccent,line width=0.6pt] (4.50,1.30) -- (gwlive.south);

\draw[->,line width=0.6pt,black!70] (0.05,-0.12) -- (7.55,-0.12);
\draw[->,line width=0.6pt,black!70] (-0.12,0.05) -- (-0.12,3.35);

\node[axlab,anchor=north] at (3.78,-0.32) {EG rollout affordability $\bar{K}$};
\node[pole,anchor=north west] at (0.00,-0.20) {\emph{Cheap}};
\node[pole,anchor=north east] at (7.55,-0.20) {\emph{Costly}};

\node[axlab,rotate=90,anchor=south] at (-0.73,1.68) {FA attainable fidelity $\bar{F}$};
\node[pole,rotate=90,anchor=south] at (-0.25,0.50) {\emph{Verifiable}};
\node[pole,rotate=90,anchor=south] at (-0.25,2.70) {\emph{Gameable}};
\end{tikzpicture}
\vspace{-5mm}
\caption{\textbf{Cross-scenario positioning matrix.}
The axes are the scenario-fixed ceilings $\bar{K}$ and $\bar{F}$, both decreasing away from the origin. Each scenario sits at the inferred height of its ceilings, with shading deepening from MR (neither binding) to EM (both binding). The MG, GW, and EM cells denote open-ended prompts, sandboxed environments with state scripts, and physical deployment, and dashed arrows show the displacement of closed-form generation sub-families, of navigation on live websites, and of simulated manipulation. 
}
\label{fig:positioning_matrix}
\end{figure}

\subsection{Cross-Scenario Positioning}\label{sec:positioning_matrix}

The stage analyses locate where each scenario's loop is bound, and this subsection condenses them into a positioning matrix that accounts for the cross-scenario asymmetry. 
To serve as an axis, a stage factor must be \emph{scenario-exogenous}, fixed by task structure instead of algorithmic choice, and \emph{substitution-resistant}, bounded in a way that the paired factor of its stage cannot relax. 
Realized values, which each method sets through its budget and evaluator, fail the first condition, leaving the ceilings as the only admissible axes.

\pgraph{Axis Selection}\; We select the ceilings $\bar{K}$ and $\bar{F}$ as the two axes, the only stage factors satisfying both conditions. The horizontal axis $\bar{K}$ is fixed by interaction cost and recoverability, and no information-gain intervention lifts it. With $B$ shared as a comparison convention, not a measured quantity, $\bar{K}$ orders scenarios by $\mathbb{E}_\tau[c(\tau)]$. The vertical axis $\bar{F}$ is fixed by verifier availability, and no localization refinement raises it. We assess $\bar{F}$ on the task families that define a scenario's target, such as open-ended prompts in generation and open task sets on hardware in manipulation, excluding sub-families whose closed-form success predicates push the ceiling near one. Its value is the highest fidelity a reported evaluator attains on these families, an estimate that may run high, agreement measured on fixed outputs tending to fall under optimization. The other four factors each fail a condition: the information gain fails scenario-exogeneity, its ceiling $\sup_{\omega} \mathbb{E}[I_{\mathcal{A}_k}]$ being set by backbone coverage; the localization $L_\nu$ fails substitution-resistance in that a higher $F_\nu$ offsets its deficit when an unbiased anchor exists; and $S_\zeta$ and $G_\zeta$ have no scenario-fixed ceiling.

\begin{table*}[!t]
\centering
\footnotesize
{\renewcommand\tabularxcolumn[1]{m{#1}}
\caption{\textbf{Cross-cycle degradation modes.}
Each failure mode is characterized by its mechanism, the stage where it surfaces, the path that carries it, and where it is primarily mitigated, all inferred from within-run, fixed-loop, and out-of-loop evidence. 
}
\label{tab:coupling}
\vspace{-2mm}
\renewcommand{\arraystretch}{1}
\rowcolors{1}{themelight}{white}
\begin{tabularx}{\textwidth}{%
>{\raggedright\arraybackslash}m{2.1cm}
>{\raggedright\arraybackslash}m{7.0cm}
>{\raggedright\arraybackslash}m{1.5cm}
>{\raggedright\arraybackslash}m{2.3cm}
>{\raggedright\arraybackslash}X
}
\toprule
\rowcolor{gray!25}
\textbf{Failure mode} & \textbf{Cross-cycle mechanism} & \textbf{Surfaces at} & \textbf{Carried by} & \textbf{Primary mitigation locus}
\\
\midrule
\ding{172} Compounding reward hacking
& An FA fidelity deficit inside $V_\nu$'s calibration support, tolerable per cycle, re-amplifies into a directional bias
& UC
& Agent edge, into FA if $V_\nu$ reads $\mathcal{A}_k$
& Tighten $F_\nu$ at FA, or shrink the update step \\
\ding{173} Distribution shift
& Rollouts drift off $V_\nu$'s calibration support, where its errors are extrapolations
& FA
& Agent edge, environment path
& Recalibrate $V_\nu$ at FA, or resample in EG \\
\ding{174} Forgetting and plasticity loss
& A moving goal distribution makes writes to $\Phi_\theta$ interfere, erasing earlier skills and eroding plasticity~\cite{dohare2024plasticity}
& UC
& Environment path, goal edge
& Replay earlier goals in EG, or isolate or reset parameters \\
\ding{175} Curriculum collapse
& $\sigma_\omega$ contracts the goal support toward mastered regions, causing $\mathbb{E}[I_{\mathcal{A}_k}]$ to decay
& EG
& Goal edge, or agent edge into $\sigma_\omega$
& Inject exploration into $\sigma_\omega$, or score goals by difficulty \\
\ding{176} Model collapse
& Self-generated training data narrows $\pi_{\mathcal{A}_k}$ and erases its tails, driving $\mathbb{E}[I_{\mathcal{A}_k}]$ down under any sampler~\cite{shumailov2024collapse}
& EG
& Agent edge
& Accumulate real or external data instead of replacing it~\cite{gerstgrasser2024collapse} \\
\bottomrule
\end{tabularx}
}
\end{table*}

\pgraph{Quadrant Assignment}\; We place each scenario in a quadrant of \cref{fig:positioning_matrix} by the height of its two ceilings, calling a ceiling binding when it is low. Multimodal reasoning faces neither low ceiling, multimodal generation a low $\bar{F}$ on open-ended prompts, GUI/Web navigation a low $\bar{K}$ under latency-bound episodes in sandboxed environments whose end states per-task scripts inspect, and embodied manipulation both in physical deployment, its target. 
Three displacements, drawn dashed in the matrix, qualify this placement: (\emph{i}) most embodied loops learn in simulation, where success predicates raise $\bar{F}$ and group sampling eases $\bar{K}$, shifting the operating point toward the navigation cell; (\emph{ii}) generation contains verifiable sub-families such as OCR-scored prompts~\cite{flowgrpo,wanr1,ardrag,identitygrpo,multicrafter} and program-rendered charts~\cite{rsiumm}, raising $\bar{F}$ toward the reasoning cell; and (\emph{iii}) navigation on live websites or free-form goals lacks state scripts and falls to screenshot judges~\cite{zerogui,openwebvoyager,skillweaver}, lowering $\bar{F}$ toward the manipulation cell. 
The four scenarios thus occupy four quadrants, consistent with the bottom rows of \cref{tab:stage1_dims,tab:stage2_dims}, and the two off-diagonal cells, cheap but gameable generation and costly but verifiable navigation, are combinations that a single difficulty ordering would merge. In the absence of any study that estimates either ceiling under one definition in more than one scenario, the placement remains inferred from reported operating points.

\pgraph{Backbone Prior as a Confound}\; The backbone $\mathcal{A}_0$ confounds this placement by bounding what a loop can reach independently of task structure. In text-only models, self-verification sharpens a model toward outputs it already ranks highly~\cite{huang2025sharpening}, reinforcement learning with verifiable rewards narrows the problems solved at large sampling budgets~\cite{yue2025rlvr}, and gains from spurious rewards vary with the model family~\cite{shao2026spurious}. Backbones also differ across scenarios, from vision-language models pretrained on large question-answering corpora in reasoning to action policies with far less motor pretraining in manipulation. Part of the contrast that the matrix attributes to $\bar{K}$ and $\bar{F}$ may thus reflect backbone coverage, a factor that stays entangled with task structure.

\pgraph{Feasibility Filter}\; The matrix supplies the first of two checks on whether a mechanism transfers across scenarios. Scenario-fixed by construction, the two ceilings bound the rollouts and the fidelity open to any learning loop within a quadrant. A transferred mechanism further presupposes resources of its source scenario that neither ceiling records, such as output equivalence or checkable outcomes, and where the target lacks them, the mechanism is conditional on an adaptation or a sub-family that supplies them. The matrix thus bounds within-cycle feasibility without deciding portability, its verdicts remaining untested while each reported comparison stays confined to one scenario.

\subsection{Cross-Cycle Coupling}\label{sec:coupling}

Beyond the matrix's within-cycle bounds, the recurrence of \cref{eq:macro_evolution_flow} couples successive cycles. 
It feeds two cycle-$k$ outputs into Stage~1 of cycle $k{+}1$, the updated agent $\mathcal{A}_{k+1}$ that induces the next rollout policy and the extended history $\mathcal{H}_{k+1}$ that conditions the next goal draw, each defining a directed edge across the cycle boundary. 
Environmental drift, which \cref{def:se_env} holds fixed but live websites exhibit~\cite{openwebvoyager,opagent}, enters as a third path from outside the loop. 
These paths carry both self-reference and long-run degradation.

\pgraph{Two Coupling Edges}\; The two edges differ in carrier and conditionality. The \emph{agent edge} runs from Stage~3 of cycle $k$ to Stage~1 of cycle $k{+}1$, carrying the persisted update into $\mathcal{A}_{k+1}$ and hence into the rollout distribution of $\mathcal{D}_{k+1}$. Present in every autonomous learning loop, it supplies the closure on which $\mathcal{C}_4$ rests and extends to any further component that reads the current agent: to $\sigma_\omega$ under a trained proposer~\cite{evolmm,visplay}, to Stage~2 under an evaluator sharing $\Phi_\theta$~\cite{evolmm,mmzero}, rendering $F_\nu^{(k)}$ endogenous, and to Stage~3 under a generative model that records lessons or skills~\cite{searrobot,mobileagente}. The \emph{goal edge} runs from Stages~1 and~2 of cycle $k$ to Stage~1 of cycle $k{+}1$ through $\sigma_\omega(k{+}1, \mathcal{H}_{k+1})$, carrying the cycle-$k$ trajectories and their scores into the next goal draw. A stationary or scheduled sampler, reading no cycle-specific signal, leaves this edge inert, whereas an adaptive sampler activates it and superimposes a second feedback loop on the agent edge. Self-reference thus acts through the extensions of the agent edge and through the goal edge: a fixed-loop method retains only the agent edge into Stage~1, and a frozen snapshot satisfies self-reference without opening any of these paths.

\pgraph{Five Degradation Modes}\; We interpret long-run degradation through these paths under a \emph{mitigation-locality reading}, in which the stage where a degradation surfaces, the path that carries it, and the stage where it is best mitigated need not coincide, and \cref{tab:coupling} records all three for five modes. Compounding reward hacking and distribution shift travel along the agent edge as the in-support and extrapolation variants of Goodhart's law~\cite{manheim2018goodhart}, which reward-model overoptimization exhibits together~\cite{gao2023scaling}, with shift also entering along the environment path~\cite{digirl,openwebvoyager}. Forgetting and plasticity loss follow a moving goal distribution carried by the environment path or the goal edge~\cite{dohare2024plasticity}, and the two collapse modes contract the support of EG, curriculum collapse through the goal edge or the agent edge into $\sigma_\omega$~\cite{wu2025selfplay,visplay} and model collapse through the agent edge under any sampler~\cite{shumailov2024collapse}. The three loci diverge most for hacking and forgetting, which both surface at UC yet admit remedies upstream of the symptom, a tighter $F_\nu$ at FA~\cite{solace} and replay in EG. Long-cycle drift compounds the first three modes within one task family and forms no sixth mode. The agent-edge extension into UC, absent from the table, surfaces through erroneous writes, within a single pass in Mobile-Agent-E~\cite{mobileagente} and across later rollouts in SEAR~\cite{searrobot}. With existing loops running only a few rounds, these mechanisms are inferred from within-run and fixed-loop evidence, and the reading stops at the cycle boundary.

\pgraph{Scenario Exposure}\; The scenarios engage these paths unequally. In reasoning, a learned proposer extends the agent edge into $\sigma_\omega$ and pass-rate curricula open the goal edge~\cite{syncloop,tempozero}, both inviting curriculum collapse. Reasoning and generation both extend the agent edge into FA through self-referential verifiers, yet the outcome follows $\bar{F}$: under reasoning's high ceiling, $F_\nu^{(k)}$ drifts below $\bar{F}$, whereas under generation's low ceiling, compounding hacking sets in unless a frozen judge cuts that extension. In navigation and manipulation, task or environment changes drive shift and forgetting along the environment path and agent edge, and costly rollouts limit the EG resampling that counters them. Coupling thus extends the asymmetry across cycles.

\section{Self-Evolution in Multimodal Reasoning}
\label{sec:mr}

Multimodal reasoning occupies the quadrant of \cref{fig:positioning_matrix} in which neither ceiling is low, its rollouts being single decoding passes and its short answers admitting exact match against a pre-collected or computed reference.
Training against pre-collected keys under a stationary sampler forms a fixed loop, from which self-evolving methods depart along two branches.
Weight-updating loops mostly pose their own questions or train on unlabeled ones, letting the agent's outputs supply most answers and tying the evaluator to the model it trains, which makes the realized fidelity $F_\nu^{(k)}$, not $\bar{F}$, the binding factor.
Training-free loops keep a frozen VLM with the key or a program and let the agent write the skills or tools it reuses, several gating each write against a validation reference, moving the cost of self-reference to the frozen writer.
Reported loops run within one training run, over up to nine rounds of weight updates or sixteen of training-free writes, or over one pass of a benchmark stream.

\begin{table*}[p]
\centering
\scriptsize
\renewcommand{\arraystretch}{1}
\setlength{\tabcolsep}{3.5pt}
\caption{\textbf{Self-evolving methods for multimodal reasoning.}
Methods are grouped by whether the agent generates questions and visual inputs, and each loop component is tagged by its realizing party: \rl{A} agent, \rl{B} bootstrap, and \rl{E} environment or task generation.
Notation: \textsuperscript{f} frozen-snapshot, \textsuperscript{ema} moving-average, and \textsuperscript{s} same-model evaluator; M memory, T tool, and I instruction update; \textsuperscript{\dag} cross-round and \textsuperscript{\ddag} single-point fidelity audit; \textit{1 run} continuous updating within one run; -- not reported.
}
\label{tab:mr_taxonomy}
\vspace{-2mm}
\rowcolors{3}{themelight}{white}
\renewcommand{\tabularxcolumn}[1]{m{#1}}
\begin{tabularx}{\textwidth}{>{\raggedright\arraybackslash}m{2.45cm}>{\centering\arraybackslash}m{0.35cm}>{\raggedright\arraybackslash}X>{\centering\arraybackslash}m{0.55cm}>{\centering\arraybackslash}m{0.65cm}>{\raggedright\arraybackslash}X>{\centering\arraybackslash}m{0.35cm}>{\raggedright\arraybackslash}m{2.0cm}>{\centering\arraybackslash}m{0.8cm}>{\raggedright\arraybackslash}m{1.7cm}}
\toprule
\rowcolor{gray!25}
\rule{0pt}{2.6ex}& \multicolumn{3}{c}{\textbf{Experience Generation (EG)}} & \multicolumn{2}{c}{\textbf{Feedback Acquisition (FA)}} & \multicolumn{3}{c}{\textbf{Update Consolidation (UC)}} & \\
\noalign{\global\aboverulesep=0pt\global\belowrulesep=0pt}
\arrayrulecolor{gray!25}\cmidrule{1-1}\cmidrule{10-10}\arrayrulecolor{black}\cmidrule(lr){2-4}\cmidrule(lr){5-6}\cmidrule(lr){7-9}
\noalign{\global\aboverulesep=0.4ex\global\belowrulesep=0.65ex}
\rowcolor{gray!25}
\multirow{-2}{*}{\textbf{Method}}
& \rule[-1ex]{0pt}{2.8ex}$\boldsymbol{\sigma}$
& \textbf{Goal content} & $\boldsymbol{K_k}$
& $\boldsymbol{V}$
& \textbf{Scoring reference \& signal}
& $\boldsymbol{f}$
& \textbf{Update}
& \textbf{Rounds}
& \multirow{-2}{*}{\textbf{Backbone}} \\
\midrule
\rowcolor{white}\multicolumn{10}{@{}l}{\;\,\ding{60} \textit{Questions from datasets or fixed generators}} \\
MM-UPT~\cite{mmupt} {\tiny NeurIPS'25} & \rl{B} & unlabeled and pre-synthesized questions & -- & \rl{A} & consensus, majority vote & \rl{B} & GRPO & 1 run & Qwen2.5-VL \\
TTRV~\cite{ttrv} {\tiny CVPR'26} & \rl{B} & 20 unlabeled test inputs & -- & \rl{A} & consensus, answer frequency, entropy control & \rl{B} & GRPO & 1 run & InternVL, Qwen \\
TTRSD~\cite{ttrsd} {\tiny arXiv'26} & \rl{B} & 20 unlabeled inputs & -- & \rl{A} & consensus, agreement across three views & \rl{B} & token-selective GRPO & 1 run & InternVL3 and others \\
IQA-SelfEv.~\cite{iqaselfevolve} {\tiny ICLR'26} & \rl{B} & unlabeled image pairs & -- & \rl{A} & consensus, pairwise majority vote & \rl{B} & GRPO & 2 & Qwen2.5-VL \\
SelfJudge~\cite{selfjudge} {\tiny arXiv'26} & \rl{B} & unlabeled questions & 8 & \rl{A} & consensus with self-judgment\textsuperscript{f} & \rl{B} & GRPO & 1 run & Qwen2.5-VL \\
Sherlock~\cite{sherlock} {\tiny NeurIPS'25} & \rl{B} & 5K unlabeled questions per round & -- & \rl{A} & self-judgment, self-correction preference & \rl{B} & trajectory-level DPO & 2 & Llama-3.2-V \\
ISR-DPO~\cite{isrdpo} {\tiny AAAI'25} & \rl{B} & unlabeled video questions & -- & \rl{A} & self-judgment, self-retrospective preference & \rl{B} & iterative DPO & 9 & 7B video LMM \\
BUS~\cite{bus} {\tiny arXiv'26} & \rl{B} & unlabeled questions & 8 & \rl{A} & self-judgment, backward prediction over sampled paths & \rl{B} & GRPO & 1 run & Qwen2.5/3-VL \\
Agent0-VL~\cite{agent0vl} {\tiny ICML'26} & \rl{B} & tool-integrated questions & 4/8 & \rl{A} & self-judgment, tool-grounded self-critique & \rl{B} & joint GRPO & 3 & Qwen2.5-VL \\
TVP~\cite{tvp} {\tiny ICLR'26} & \rl{B} & 501 benchmark test queries & -- & \rl{A} & self-judgment, program quality score & \rl{B} & tool and example library (T+M) & 3 & GPT-4o, frozen \\
Vision-Zero~\cite{visionzero} {\tiny ICLR'26} & \rl{B} & arbitrary images & 8 & \rl{E}+\rl{A} & computed, game outcome & \rl{B} & self-play and RLVR & 1 run & Qwen2.5-VL \\
CwI~\cite{cwi} {\tiny arXiv'26} & \rl{B} & 30 constructive task families & -- & \rl{E} & computed, answer match & \rl{A} & validation-gated skill library (T+I) & up to 12 & 4 frozen VLMs \\
Spatial Memory Ag.~\cite{spatialmem} {\tiny arXiv'26} & \rl{B} & verifiable spatial environment & -- & \rl{E}+\rl{B} & computed and key, environment reward & \rl{A} & memory cards (M) & episode & 4 frozen VLMs \\
EvoSpatial~\cite{evospatial} {\tiny arXiv'26} & \rl{B} & spatial benchmark streams & -- & \rl{E}+\rl{A} & computed checks and self-judgment & \rl{A} & skill and tool library (T+I) & online & Gemma4-31B, frozen \\
Vision-SR1~\cite{visionsr1} {\tiny arXiv'25} & \rl{B} & 47K labeled questions & -- & \rl{B}+\rl{A} & key and self-judgment, perception re-answering & \rl{B} & multi-reward GRPO & 1 run & Qwen2.5-VL \\
M$^2$Note~\cite{m2note} {\tiny arXiv'26} & \rl{B} & benchmark batches & -- & \rl{B}+\rl{A} & key or self-judgment, rollback-gated & \rl{A} & mistake notebook (M) & online & Qwen3-VL, GPT-5.4, frozen \\
VisualEvolve~\cite{visualevolve} {\tiny arXiv'26} & \rl{B} & labeled development tasks & -- & \rl{B} & key, failure diagnosis & \rl{A} & validation-gated skill library (T+I) & 16 & 4 frozen VLMs \\
CoWorld~\cite{coworld} {\tiny arXiv'26} & \rl{B} & 120 labeled development tasks & -- & \rl{B} & key, answer scoring & \rl{A} & selected harness (T+I) & 3 & GPT-5.6, frozen \\
Dynamo~\cite{dynamo} {\tiny arXiv'26} & \rl{B} & labeled training subset & -- & \rl{B} & key, training-sample accuracy & \rl{A} & gated skill-tool library (T+I) & 3 & 5 frozen VLMs \\
PhysNote~\cite{physnote} {\tiny ICLRW'26} & \rl{B} & 500 labeled physics videos & -- & \rl{B} & key, answer match & \rl{A} & knowledge notes (M) & 1 pass & Qwen2.5-VL-72B, frozen \\
NeSy-Spatial~\cite{nesyspatial} {\tiny arXiv'26} & \rl{B} & spatial benchmark streams & -- & \rl{B} & key, revealed after answering & \rl{A} & skill library (T+I) & online & GPT-5.4, Gemini, frozen \\
MetaForge~\cite{metaforge} {\tiny arXiv'26} & \rl{B} & labeled multimodal tasks & -- & \rl{B} & key, composite tool-use reward & \rl{A} & GRPO + tool library (T) & 1 run & Qwen2.5/3-VL \\
SyncLoop~\cite{syncloop} {\tiny arXiv'25} & \rl{A} & generator-evolved diagram problems & -- & \rl{B} & key, constructed answers & \rl{B} & SFT and GRPO & 3 & Qwen2-VL \\
\midrule
\rowcolor{white}\multicolumn{10}{@{}l}{\;\,\ding{60} \textit{Agent-generated questions}} \\
EvoLMM~\cite{evolmm} {\tiny arXiv'25} & \rl{A} & unlabeled images & 5 & \rl{A} & consensus, soft agreement & \rl{B} & REINFORCE & 1 run & Qwen2.5-VL \\
VisPlay~\cite{visplay} {\tiny CVPR'26} & \rl{A} & unlabeled images & -- & \rl{A} & consensus, majority vote & \rl{B} & GRPO & 3\textsuperscript{\dag} & Qwen2.5-VL, MiMo \\
V-Zero~\cite{vzero} {\tiny arXiv'26} & \rl{A} & unlabeled images & 5 & \rl{A} & consensus, majority vote & \rl{B} & iterative GRPO & 2 & Qwen2.5-VL \\
iReasoner~\cite{ireasoner} {\tiny ACL Findings'26} & \rl{A} & unlabeled images & 5 & \rl{A} & consensus, step-level agreement & \rl{B} & GRPO & 1 run & Qwen2.5-VL \\
Video-Zero~\cite{videozero} {\tiny NeurIPS'26} & \rl{A} & unlabeled videos, temporal spans & -- & \rl{A} & consensus, vote with temporal alignment & \rl{B} & alternating GRPO & 5\textsuperscript{\dag} & Qwen3-VL \\
EvoVid~\cite{evovid} {\tiny arXiv'26} & \rl{A} & unlabeled videos, shuffle-sensitive questions & 10 & \rl{A} & consensus, vote with temporal IoU & \rl{B} & alternating GRPO & 3 & Qwen2.5/3-VL \\
RISE~\cite{rise} {\tiny arXiv'26} & \rl{A} & unlabeled images, skill-balanced & 10 & \rl{A} & consensus with supervisor\textsuperscript{s} & \rl{B} & alternating GRPO & 4\textsuperscript{\ddag} & Qwen2.5/3-VL \\
VISE~\cite{vtokens} {\tiny arXiv'26} & \rl{A} & unlabeled images, localization queries & -- & \rl{A} & consistency under image transformations & \rl{B} & REINFORCE & 1 run & Qwen3-VL and others \\
SVQ~\cite{svq} {\tiny arXiv'26} & \rl{A} & images & -- & \rl{A} & self-judgment\textsuperscript{f}, rewrite and filter & \rl{B} & SFT & 2 & Qwen2.5/3-VL \\
Trinity~\cite{trinity} {\tiny arXiv'26} & \rl{A} & images & -- & \rl{A} & self-judgment\textsuperscript{ema}, verifier scores & \rl{B} & joint GRPO & 1 run & Qwen2.5/3-VL \\
Marvo~\cite{marvo} {\tiny arXiv'26} & \rl{A} & unlabeled images, tool-use questions & 8 & \rl{A} & proposer answer, tool-use margin & \rl{B} & alternating GRPO & 3 & Qwen2.5/3-VL \\
VideoEvoHarness~\cite{videoevoharness} {\tiny arXiv'26} & \rl{A} & unlabeled videos, memory-grounded questions & -- & \rl{A} & proposer answer from video memory & \rl{A} & validated harness (T+I) & 3 & Qwen3.7 Plus, Gemini, frozen \\
VQS~\cite{pvse} {\tiny arXiv'26} & \rl{A} & parsed records, template questions & 8 & \rl{E}+\rl{A} & computed, program answers & \rl{B} & GRPO & 3\textsuperscript{\ddag} & Qwen3-VL \\
EVE~\cite{eve} {\tiny arXiv'26} & \rl{A} & executable visual transformations & 6 & \rl{E}+\rl{A} & computed, code execution & \rl{B} & alternating GRPO & 3 & Qwen2.5-VL \\
SpatialEvo~\cite{spatialevo} {\tiny arXiv'26} & \rl{A} & unlabeled 3D scenes, spatial questions & -- & \rl{E}+\rl{A} & computed, geometric environment & \rl{B} & GRPO & 4 & Qwen2.5-VL \\
DUEL~\cite{duel} {\tiny arXiv'26} & \rl{A} & unlabeled images, perturbed claims & -- & \rl{E}+\rl{A} & computed, claim-verification outcome & \rl{B} & GRPO & -- & Qwen2.5-VL \\
SAGE~\cite{sage} {\tiny arXiv'26} & \rl{A} & dual transformations of labeled items & -- & \rl{B}+\rl{E} & key and answer mapping, duality consistency & \rl{B} & GRPO & 1 run & Qwen2.5/3-VL \\
\bottomrule
\end{tabularx}
\vspace{-3mm}
\end{table*}

\begin{table*}[!t]
\ContinuedFloat
\centering
\scriptsize
\renewcommand{\arraystretch}{1}
\setlength{\tabcolsep}{3.5pt}
\caption[]{\textbf{Self-evolving methods for multimodal reasoning} (continued).}
\vspace{-2mm}
\rowcolors{3}{themelight}{white}
\renewcommand{\tabularxcolumn}[1]{m{#1}}
\begin{tabularx}{\textwidth}{>{\raggedright\arraybackslash}m{2.45cm}>{\centering\arraybackslash}m{0.35cm}>{\raggedright\arraybackslash}X>{\centering\arraybackslash}m{0.55cm}>{\centering\arraybackslash}m{0.65cm}>{\raggedright\arraybackslash}X>{\centering\arraybackslash}m{0.35cm}>{\raggedright\arraybackslash}m{2.0cm}>{\centering\arraybackslash}m{0.8cm}>{\raggedright\arraybackslash}m{1.7cm}}
\toprule
\rowcolor{gray!25}
\rule{0pt}{2.6ex}& \multicolumn{3}{c}{\textbf{Experience Generation (EG)}} & \multicolumn{2}{c}{\textbf{Feedback Acquisition (FA)}} & \multicolumn{3}{c}{\textbf{Update Consolidation (UC)}} & \\
\noalign{\global\aboverulesep=0pt\global\belowrulesep=0pt}
\arrayrulecolor{gray!25}\cmidrule{1-1}\cmidrule{10-10}\arrayrulecolor{black}\cmidrule(lr){2-4}\cmidrule(lr){5-6}\cmidrule(lr){7-9}
\noalign{\global\aboverulesep=0.4ex\global\belowrulesep=0.65ex}
\rowcolor{gray!25}
\multirow{-2}{*}{\textbf{Method}}
& \rule[-1ex]{0pt}{2.8ex}$\boldsymbol{\sigma}$
& \textbf{Goal content} & $\boldsymbol{K_k}$
& $\boldsymbol{V}$
& \textbf{Scoring reference \& signal}
& $\boldsymbol{f}$
& \textbf{Update}
& \textbf{Rounds}
& \multirow{-2}{*}{\textbf{Backbone}} \\
\midrule
\rowcolor{white}\multicolumn{10}{@{}l}{\;\,\ding{60} \textit{Agent-synthesized or retrieved images and videos}} \\
MM-Zero~\cite{mmzero} {\tiny EMNLP'26} & \rl{A} & code-rendered images & 4 & \rl{A} & consensus, majority vote & \rl{B} & GRPO & 3 & Qwen3-VL, MiMo \\
Active-Zero~\cite{activezero} {\tiny arXiv'26} & \rl{A} & retrieved open-world images & 8 & \rl{A} & consensus, majority vote & \rl{B} & GRPO & 3 & Qwen2.5-VL \\
ViCo~\cite{vico} {\tiny NeurIPS'26} & \rl{A} & edited scene graphs and chart tables & -- & \rl{E}+\rl{B} & computed or key & \rl{B} & GRPO and RWR & 5 & Qwen3-VL-8B \\
Tempo-Zero~\cite{tempozero} {\tiny arXiv'26} & \rl{A} & program-rendered videos & 8 & \rl{E}+\rl{A} & computed, execution & \rl{B} & GRPO & 5 & Qwen3-VL \\
PhysEvolve~\cite{physevolve} {\tiny arXiv'26} & \rl{A} & simulated physical scenes & -- & \rl{E}+\rl{A} & computed, physics engine & \rl{B} & GRPO & 1 run & Qwen3-VL-8B \\
\bottomrule
\end{tabularx}
\end{table*}

\subsection{EG Bound by Informative Goals}\label{sec:mr_eg}\label{sec:mr_patterns}

Experience generation in multimodal reasoning draws goals that pair a visual input (an image, a video clip, or frames of a 3D scene) with a question whose short final answer admits an equivalence check.
The task families center on visual question answering over mathematical, chart, and diagram inputs~\cite{lu2024mathvista,zhang2024mathverse,masry2022chartqa}, extending to pairwise quality judgments~\cite{iqaselfevolve}, temporally grounded video questions~\cite{videozero,evovid}, metric spatial questions over 3D scenes~\cite{spatialevo}, and physical dynamics in videos~\cite{physevolve,physnote,tempozero}.
These goals come from a dataset, a fixed generator, or generative roles of the agent, and in weight-updating loops the solver typically answers each goal in $K_k$ independent decoding passes, occasionally with tool calls, yielding a $\mathcal{D}_k$ that groups answers to the same question, whereas training-free loops run the frozen VLM on each goal with its current skills, tools, or notes.
\Cref{tab:mr_taxonomy} summarizes the self-evolving methods that satisfy the conditions of \cref{sec:defining_conditions}.

\pgraphhlY{Quantity and Quality Constraints}\; Experience generation is bound by quality at saturation (\cref{tab:stage1_dims}), the task structure leaving quantity slack.
A rollout is a single decoding pass over a static input, an episode with $T\!=\!1$ except where tool calls precede the answer~\cite{agent0vl,marvo,metaforge,dynamo,evospatial,coworld}, and leaves no side effect to revert, keeping both interaction cost and recoverability from lowering $\bar{K}$.
Information gain, in contrast, arises in single-answer episodes only from goals whose pass rate under the current solver lies strictly between zero and one.
A dataset or a fixed generator bounds the supply of such goals by how many lie near the edge of the current agent's competence, a number that depends on the backbone's pretraining and, by the mechanism that stalls self-play~\cite{wu2025selfplay}, shrinks as the agent masters these goals.

\pgraphhlP{Self-Posed Goals}\; Larger groups do not appear to substitute for informative goals under self-consistency rewards, which lose contrast as groups grow, and loops raise the information gain instead by letting the agent pose its goals or by reweighting goals by pass rate, at a cost paid in feedback.
Reported group sizes stay between four and ten rollouts per goal despite cheap decoding passes~\cite{vzero,agent0vl,mmzero,evolmm,activezero,visionzero,eve,rise,evovid,marvo,bus}, and EvoLMM finds its self-consistency reward best at four to six samples, reporting over-consensus collapse at twelve~\cite{evolmm}, consistent with the reward, not the cost, limiting $K_k$.
The agent enters $\sigma_\omega$ either as a generative role within $\Phi_\theta$ that synthesizes the question beside the solver, and in some loops the visual input, or through a rule that reweights dataset or generator goals by the solver's pass rate, alone in SyncLoop~\cite{syncloop} or on top of a learned proposer~\cite{sage,physevolve,tempozero}.
Proposer-solver loops after R-Zero~\cite{rzero} reward goals of intermediate solver agreement through terms that vanish with $p_g(1-p_g)$~\cite{eve,rise}, adding diversity terms against a narrowing of goals~\cite{evovid,visplay}.
These rewards are computed in FA yet reshape the goal distribution of EG.
In VisPlay, a trained Questioner lifts Qwen2.5-VL-3B from 30.61 to 44.16 after one round against 33.77 for an untrained Challenger~\cite{visplay}, a comparison that attributes about three quarters of the gain to learning the role.
The gain is paid in feedback, where a self-posed question lacks a key unless a program computes its answer, leaving the answer to the agent's own outputs.

\subsection{FA Bound by Realized Fidelity}\label{sec:mr_feedback}

Each cycle hands feedback acquisition groups of short answers to the same question, or the tool-calling episodes of a frozen agent, and the loops score each answer against one or more of three references (\cref{tab:mr_taxonomy}).
The agent's own outputs (\rl{A}), through a majority vote, a self-judgment, the proposer's answer, or consistency under image transformations, score self-posed or unlabeled questions~\cite{evolmm,visplay,mmupt}.
An answer computed by task generation (\rl{E}) scores questions whose answers programs, game rules, a geometric environment, or a physics engine determine~\cite{pvse,visionzero,spatialevo,physevolve}, and a pre-collected key (\rl{B}) scores labeled sets~\cite{visionsr1,visualevolve,dynamo}.
Weight-updating methods mostly turn scores into group-normalized rewards for GRPO-style updates or into preference pairs for DPO~\cite{vzero,visplay,sherlock,isrdpo}, whereas training-free methods use them to admit skills, tools, or notes into a library~\cite{cwi,dynamo,m2note}.

\pgraphhlY{Fidelity and Localization Constraints}\; Feedback acquisition is bound by realized fidelity (\cref{tab:stage2_dims}), its attainable fidelity being high and localization imposing little.
A short answer admits exact match against a reference fixed outside the agent and leaves no channel unchecked, and without a key the program-computed answers of VQS stay correct on 94.4\% of retained questions among 500 human-audited ones~\cite{pvse}, a lower bound on $\bar{F}$ under the conversion of \cref{sec:stage2}.
Consensus already falls below this bound within one round, with 24\% of retained majority-vote labels and 18\% of retained model-judge labels wrong in the same audit~\cite{pvse}.
Across rounds, externally estimated pseudo-label accuracy falls from 72.0 to 61.0 over three iterations in VisPlay~\cite{visplay} and from 64.5 to 37.5 over five in Video-Zero~\cite{videozero}, declines that bear on $F_\nu^{(k)}$ only jointly with rising question difficulty, to which VisPlay attributes its own.
Gaming resistance of the vote under long optimization remains untested here, whereas prolonged training against a majority-vote self-reward in text-only reasoning ends in reward hacking and abrupt collapse~\cite{shafayat2025selftrain}.
Single-answer episodes fix $L_\nu\!=\!1$, the step-level agreement of iReasoner refining credit only within the output~\cite{ireasoner}, and only the short tool-integrated episodes of Agent0-VL, scored as a whole~\cite{agent0vl}, leave credit coarser.

\pgraphhlP{Vote Preconditions and Evaluator Drift}\; With localization free, every deficit of feedback acquisition is a fidelity deficit, and self-derived rewards make that fidelity a property of the model being trained.
A vote recovers the model's belief and tracks $R$ only where answers admit an equivalence relation and the pretrained prior concentrates on the correct answer~\cite{wang2023selfconsistency,chen2023universal}, discrete answers supplying the first condition and the backbone, not the task structure, deciding the second.
Within these conditions the reward form matters, TTRV lowering accuracy on four of five VQA benchmarks under a hard majority-vote reward and raising it on all five under a frequency-proportional reward with entropy control~\cite{ttrv}.
The evaluators further differ in whether they move with the policy: votes over current samples, parameter-sharing supervisors~\cite{rise}, and self-judging models~\cite{isrdpo} keep the agent edge into FA, judges frozen at the initial model cut it~\cite{selfjudge,svq}, and the moving-average verifier of Trinity lags the policy by construction~\cite{trinity}.
A frozen evaluator still shares the prior of the policies descended from it, removing drift but not a fixed bias, and reports under differing numbers of rounds leave the designs uncompared.
Answers fixed outside the agent, by a key or a program, restore $\bar{F}$ only for labeled sets and program-computable families, the same restriction that experience generation imposes on the goals.

\subsection{UC Bound by Retention}\label{sec:mr_uc}

Update consolidation in multimodal reasoning persists the scored experience in two forms, weight updates and training-free writes (\cref{tab:mr_taxonomy}).
Weight-updating methods write $\Phi_\theta$ with an update rule fixed before the loop, through group-relative policy gradients, preference optimization, or fine-tuning, either once per round over two to nine reported rounds~\cite{visplay,videozero,pvse,spatialevo,vico,tempozero,isrdpo} or continuously within a single training run.
Training-free methods keep a frozen VLM and let the agent write text or code into skill and tool libraries~\cite{cwi,visualevolve,dynamo,evospatial,nesyspatial}, memory cards or notes~\cite{spatialmem,m2note,physnote}, or harnesses of tools, skills, and execution code~\cite{coworld,videoevoharness}, over benchmark streams or, in VisualEvolve, up to sixteen rounds, whereas a separate o4-mini abstracts the tool library of TVP~\cite{tvp}.
MetaForge alone combines the two, registering validated tools that its policy forges during GRPO into a persistent library~\cite{metaforge}.

\pgraphhlY{Retention and Generalization Constraints}\; Update consolidation is bound by retention (\cref{tab:stage3_dims}), inferred from non-monotone curves, not a measured loss, whereas generalization, though reported, is confounded with the backbone prior.
No method measures retention on earlier-round goals, and the closest evidence, per-round scores on a fixed benchmark suite, traces the increment $\mathcal{J}_{k+1}-\mathcal{J}_k$ of \cref{eq:eval_seq} over the evaluation distribution, not the trained goals.
These curves rise monotonically in Agent0-VL~\cite{agent0vl} but peak or dip mid-run elsewhere~\cite{activezero,visplay}, while single benchmarks reverse more often, MM-Vet falling from 48.62 to 38.07 over three VisPlay iterations as the average rises~\cite{visplay}.
Among weight-updating methods, only ISR-DPO and Marvo run beyond five rounds, the former gaining through its ninth iteration~\cite{isrdpo} and the latter degrading after about the fifth or sixth of eight~\cite{marvo}.
Parametric writes generalize more freely, reaching any goal that activates the updated weights and lifting standard benchmarks after training on synthesized questions~\cite{vzero,visplay,evolmm}.
These gains are consistent with elicitation of the backbone prior, most weight-updating methods resting on Qwen2.5-VL or Qwen3-VL, the family whose text-only counterparts gain even under spurious rewards~\cite{shao2026spurious}.
V-Zero offers a partial control, reaching 51.9 against 50.8 for supervised GRPO on the same Qwen2.5-VL-7B backbone~\cite{vzero}, and only an equal-budget non-evolving baseline would separate self-evolution from elicitation.
Non-parametric writes act only when retrieved, and their transfer is tested directly, EvoSpatial evaluating libraries frozen after one benchmark on 18 others~\cite{evospatial} and VideoEvoHarness moving an evolved harness to a second backbone~\cite{videoevoharness}.

\pgraphhlP{Reference-Gated Writes}\; Consolidation inherits its exposure from the first two stages, and the two forms of writing differ in how they guard against it.
Where the evaluator moves with the policy or a proposer supplies the goals, a parametric write is scored and driven by quantities that change with the update itself, and the update rule accepts every scored step without a fixed reference that could reveal a later round erasing an earlier gain.
Several training-free methods gate their writes against such a reference, admitting a skill or change only on a validation or held-out gain~\cite{cwi,videoevoharness}, promoting a skill-tool pair only on a strict gain over its training sample~\cite{dynamo}, rolling back note edits that do not raise batch accuracy~\cite{m2note}, or selecting snapshots on fixed validation or development sets~\cite{visualevolve,coworld}, whereas TVP checks a new tool only against the outputs of the programs it replaces~\cite{tvp}.
A guard on retention thus exists only where gated training-free writes keep a fixed reference.

\subsection{Limits and Preconditions}\label{sec:mr_positioning}

\pgraph{Bottleneck Attribution}\; The binding bottlenecks of feedback and consolidation arise from self-reference, not from task structure.
Both ceilings are slack for any learning loop, and a fixed loop that trains against pre-collected keys, as M-STAR does~\cite{mstar}, scores its rollouts at the fidelity of an exact match.
The weight-updating branch gives up this reference to pose and answer its own questions, its answers coming from the agent's own outputs, whose realized fidelity falls below $\bar{F}$, and its parametric writes lacking a fixed reference against which a later round could be checked.
The training-free branch keeps the key or a validation set and admits self-reference only through the update, shifting the cost to the writer, whose frozen backbone both diagnoses a bottleneck and writes its remedy, its capability bounding both, a limitation that Dynamo reports for itself~\cite{dynamo}.
Self-reference repays part of this cost in experience generation, where a learned proposer relieves the saturation of fixed goal supplies and accounts for most of the first-round gain of VisPlay~\cite{visplay}.
The few cross-round audits report declines confounded with rising question difficulty, and a large question-answering prior in the backbones may account for part of the gains in both branches, disqualifying the least constrained cell as a clean baseline for the other scenarios.
This attribution rests on the stage evidence and on the provenance of each method in \cref{tab:mr_taxonomy}, not on a controlled comparison with fixed loops.

\pgraph{Cross-Cycle Degradation}\; Across cycles, self-reference exposes multimodal reasoning to degradations that a fixed loop with keys avoids.
On the goal side, learned proposers extend the agent edge into $\sigma_\omega$ and pass-rate curricula open the goal edge~\cite{syncloop,sage,physevolve,tempozero}, both inviting curriculum collapse.
On the feedback side, self-derived rewards extend the agent edge into FA, inviting compounding reward hacking, and training on self-labeled answers invites model collapse along the edge into the rollout policy.
The extension of training-free writers into UC remains untested, and cross-round evidence reduces to the Vision-Zero plateau, where a self-derived signal stops informing updates~\cite{visionzero}, and the late-round decline of Marvo~\cite{marvo}, with per-round reversals of single benchmarks fitting hacking and forgetting alike.

\pgraph{Presupposed Resources}\; The mechanisms of multimodal reasoning presuppose resources that its short, discrete answers supply.
Majority votes and agreement-based proposer rewards presuppose answers that admit an equivalence relation~\cite{evolmm,eve,rise}, program-computed answers presuppose a program that determines the answer at task generation~\cite{pvse}, and reference-gated writes presuppose a key or validation set held outside the writer~\cite{cwi,dynamo}.

\section{Self-Evolution in Multimodal Generation}
\label{sec:mg}

Multimodal generation occupies the quadrant of \cref{fig:positioning_matrix} in which only the fidelity ceiling is low, an open-ended image or video having no reference output to be checked against outside sub-families such as object counts, rendered text, or program-drawn charts.
Training against learned preference scorers fixed before the loop forms a fixed loop that this ceiling already binds, such scorers being exploitable under optimization~\cite{imgownreward,iris}.
Weight-updating loops mostly derive the verifier from the agent, leaving the ceiling unchanged but letting the verifier's bias move with the policy and exposing the loop to compounding reward hacking.
Training-free loops mostly let a frozen backbone judge its own outputs and write its memory or skills, sharing one bias between judge and writer.
Reported loops run within one training run, over at most five rounds, or over one pass of a task stream.

\subsection{EG Bound by Checkable Prompts}\label{sec:mg_eg}

In multimodal generation, experience generation draws prompts as goals, paired with a source image in editing, and the outputs are images, videos, 3D scenes, charts, diagrams, or educational animations.
The task families center on compositional text-to-image prompts~\cite{unirl,silmm,internalgap}, extending to aesthetic prompt optimization~\cite{srlvlm,prism}, text rendering~\cite{imgownreward}, safety~\cite{safecodebook}, charts~\cite{rsiumm}, text-to-video generation~\cite{dualipo,physrepair}, image editing~\cite{mtopsd,jarvisevo}, and diagrams, 3D scenes, method illustrations, and animations built through tool chains~\cite{evodiagram,scenecraft,figagent,manimagent}.
These goals come from a fixed pool or from generative roles of the agent, and weight-updating loops sample $K_k$ outputs per prompt in one pass each, whereas training-free loops run a frozen agent through a workflow of tool calls, renders, and edits.
\Cref{tab:mg_taxonomy} summarizes the self-evolving methods that satisfy the conditions of \cref{sec:defining_conditions}.

\pgraphhlY{Quantity and Quality Constraints}\; Experience generation is bound by quality at the coverage of checkable prompts (\cref{tab:stage1_dims}), the task structure leaving quantity slack.
A rollout costs one sampling pass or an offline tool chain~\cite{symbomni} and leaves no side effect to revert, with reported group sizes of three to sixteen samples per prompt~\cite{unirl,iris,corlulm,solace,readitback,imgownreward}.
Information gain, in contrast, runs out at the edge of what a check can score, not at the edge of the agent's competence as in multimodal reasoning.
Fixed pools come from benchmark categories whose attributes a detector, OCR, or a VQA model scores, such as those of GenEval and T2I-CompBench~\cite{unirl,iris,internalgap,suder}, and a learning loop drives pass rates on such a pool toward one, where $p_g(1-p_g)$ approaches zero, as the fixed loop Flow-GRPO raises SD3.5-M from 0.63 to 0.95 on GenEval~\cite{flowgrpo}.
Open-ended prompts would still separate good outputs from poor ones under $R$, yet no check fixed before the loop scores them, leaving the loop unable to draw information from them.
Loops drawing every round from one prompt distribution report diminishing gains by about the third round, consistent with this saturation~\cite{silmm,safecodebook}.

\pgraphhlP{Checkable Goal Composition}\; A larger group adds samples of a prompt without extending the checkable pool, and loops raise $\mathbb{E}[I_{\mathcal{A}_k}]$ instead by letting the agent compose or reselect goals, at a cost paid in fidelity once the goals leave what a fixed check scores.
Built-in checks keep composed goals within checkable sub-families, fixing answers at prompt construction in UniRL~\cite{unirl}, decomposing each prompt into yes/no questions in SILMM~\cite{silmm}, and restricting the specifications that RSI in UMMs plans from its own failures to charts that program execution verifies~\cite{rsiumm}.
Goals beyond these sub-families fall to frozen external judges~\cite{octot2i,mate,rsigen}, whereas ASG poses understanding questions over unlabeled images, scored by the agreement of its own answers across framings at a fidelity that moves with the policy~\cite{asg}, and reselection stays within the pool, InternalGap regenerating the prompts that yielded no accepted image~\cite{internalgap}.
Controlled returns are reported only for built-in checks and external judges, RSI in UMMs reaching 59.0\% on charts after four rounds against 46.3\% for longer training on its original data and leading a fixed distribution of equally many verified charts in every round~\cite{rsiumm}, and RSIgen gaining 4.7, 3.1, and 2.2 points over curricula compiled from frozen sources across its three attempts~\cite{rsigen}.

\begin{table*}[p]
\centering
\scriptsize
\renewcommand{\arraystretch}{1}
\setlength{\tabcolsep}{3.5pt}
\caption{\textbf{Self-evolving methods for multimodal generation.}
Methods are grouped by whether and how the verifier changes as the policy is updated, and each loop component is tagged by its realizing party: \rl{A} agent, \rl{B} bootstrap, and \rl{E} environment or task generation.
Notation: \textsuperscript{f} frozen-snapshot, \textsuperscript{p} previous-iteration, \textsuperscript{c} co-trained, \textsuperscript{ema} moving-average, and \textsuperscript{s} same-model evaluator; M memory, T tool, and I instruction update; \textit{1 run} continuous updating within one run; \textit{1 pass} a single pass over the goals; -- not reported.
}
\label{tab:mg_taxonomy}
\vspace{-2mm}
\rowcolors{3}{themelight}{white}
\renewcommand{\tabularxcolumn}[1]{m{#1}}
\begin{tabularx}{\textwidth}{>{\raggedright\arraybackslash}m{2.45cm}>{\centering\arraybackslash}m{0.35cm}>{\raggedright\arraybackslash}X>{\centering\arraybackslash}m{0.45cm}>{\centering\arraybackslash}m{0.65cm}>{\raggedright\arraybackslash}X>{\centering\arraybackslash}m{0.35cm}>{\raggedright\arraybackslash}m{2.2cm}>{\centering\arraybackslash}m{0.8cm}>{\raggedright\arraybackslash}m{1.9cm}}
\toprule
\rowcolor{gray!25}
\rule{0pt}{2.6ex}& \multicolumn{3}{c}{\textbf{Experience Generation (EG)}} & \multicolumn{2}{c}{\textbf{Feedback Acquisition (FA)}} & \multicolumn{3}{c}{\textbf{Update Consolidation (UC)}} & \\
\noalign{\global\aboverulesep=0pt\global\belowrulesep=0pt}
\arrayrulecolor{gray!25}\cmidrule{1-1}\cmidrule{10-10}\arrayrulecolor{black}\cmidrule(lr){2-4}\cmidrule(lr){5-6}\cmidrule(lr){7-9}
\noalign{\global\aboverulesep=0.4ex\global\belowrulesep=0.65ex}
\rowcolor{gray!25}
\multirow{-2}{*}{\textbf{Method}}
& \rule[-1ex]{0pt}{2.8ex}$\boldsymbol{\sigma}$
& \textbf{Goal content} & $\boldsymbol{K_k}$
& $\boldsymbol{V}$
& \textbf{Scoring reference \& signal}
& $\boldsymbol{f}$
& \textbf{Update}
& \textbf{Rounds}
& \multirow{-2}{*}{\textbf{Backbone}} \\
\midrule
\rowcolor{white}\multicolumn{10}{@{}l}{\;\,\ding{60} \textit{Self-verifiers updated with the policy}} \\
SOLACE~\cite{solace} {\tiny CVPR'26} & \rl{B} & fixed prompt pool & 16 & \rl{A} & denoising confidence under re-noising probes & \rl{B} & Flow-GRPO & 1 run & SD3.5-M \\
IRIS~\cite{iris} {\tiny arXiv'25} & \rl{B} & 553 GenEval prompts & 8 & \rl{A} & negative self-certainty over text and image tokens & \rl{B} & GRPO & 1 run & Janus-Pro \\
CoRL~\cite{corlulm} {\tiny NeurIPS'25} & \rl{B} & COCO captions paired with images and QA & 8--16 & \rl{A}+\rl{B} & own token-embedding alignment, BLIP re-caption and LPIPS cycle consistency & \rl{B} & joint then task-specific GRPO & 2 runs & Janus-Pro-1B \\
UniRL~\cite{unirl} {\tiny arXiv'25} & \rl{B} & constructed GenEval-category prompts & 3 & \rl{A} & own answers against answers fixed at prompt construction & \rl{B} & SFT and GRPO & 1 run & Show-o, Janus \\
SUDER~\cite{suder} {\tiny arXiv'25} & \rl{B} & T2I-CompBench prompts, unpaired images & -- & \rl{A} & reverse-direction likelihood in both branches & \rl{B} & SimPO or GRPO & 1 run & Janus-Pro \\
SILMM~\cite{silmm} {\tiny CVPR'25} & \rl{B} & LMM-composed compositional prompts & -- & \rl{A} & own answers to self-decomposed questions & \rl{B} & DPO or KC-DPO & 3 & DreamLLM, SEED-LLaMA \\
InternalGap~\cite{internalgap} {\tiny ICLR'26} & \rl{A} & T2I-CompBench++ prompts, discarded ones regenerated & 10 & \rl{A} & own yes/no probability of alignment & \rl{B} & SFT and DPO & 1 run & Janus-Pro, Show-o \\
Read It Back~\cite{readitback} {\tiny NeurIPS'26} & \rl{B} & AlphaGRPO20k prompts & 16 & \rl{A} & own likelihood of the prompt given the image & \rl{B} & AWM online RL & 1 run & BAGEL \\
GvU~\cite{gvu} {\tiny arXiv'26} & \rl{B} & 50K compositional prompts & -- & \rl{A} & own token-level likelihood of the prompt & \rl{B} & GRPO & 1 run & X-Omni (LoRA) \\
ASG~\cite{asg} {\tiny arXiv'26} & \rl{A} & self-posed questions over unlabeled images & 3 & \rl{A} & answer agreement across seven framings, cycle captioning & \rl{B} & GRPO on three LoRA adapters & 1 run & BLIP3o, BAGEL, VARGPT \\
Safe Codebooks~\cite{safecodebook} {\tiny ICML'26} & \rl{B} & harmful-prompt sets, self-rewritten safe prompts & -- & \rl{A} & own judgment of unsafe content & \rl{B} & codebook projection, null-space fine-tuning & 5 & Janus, VILA-U, Emu3 \\
JarvisEvo~\cite{jarvisevo} {\tiny CVPR'26} & \rl{B} & photo-editing instructions over tool chains & -- & \rl{A} & own evaluator, aligned to human scores fixed before the loop & \rl{B} & SEPO RL and reflection fine-tuning & 1 run & Qwen3-VL-8B \\
\midrule
\rowcolor{white}\multicolumn{10}{@{}l}{\;\,\ding{60} \textit{Co-evolving or lagged verifiers}} \\
SRL-VLM~\cite{srlvlm} {\tiny ACL Findings'25} & \rl{B} & raw prompts, 8 rewrites each, frozen SD1.5 & 8 & \rl{A} & previous-iteration LVLM\textsuperscript{p}, aesthetics and alignment & \rl{B} & DPO & 2 & VILA-1.5-8B \\
Image as Its Own Reward~\cite{imgownreward} {\tiny CVPR'26} & \rl{B} & PickScore, OCR, and GenEval prompt sets & 16 & \rl{A} & PickScore or DINO model\textsuperscript{c}, Qwen-Image positives & \rl{B} & GRPO & 1 run & SD3 (LoRA) \\
Dual-IPO~\cite{dualipo} {\tiny ICLR'26} & \rl{B} & LLM-expanded caption pool & -- & \rl{A} & critic\textsuperscript{c} refined on its pseudo-labels and a human seed set & \rl{B} & Diffusion-DPO or KTO & 3 & CogVideoX, Wan-1.3B \\
ImageRSI~\cite{imagersi} {\tiny arXiv'26} & \rl{B} & 530 requests, up to two repairs each & 1--3 & \rl{A} & judge retrained on loop records, blinded A/B admission & \rl{A} & experience library (I), judge SFT and GRPO & 1 judge update & Qwen-Image-2.0, frozen \\
UniEvo-VL~\cite{unievo} {\tiny arXiv'26} & \rl{B} & image generation prompts & -- & \rl{A}+\rl{B} & moving-average teacher\textsuperscript{ema} with external critiques & \rl{B} & on-policy self-distillation & 1 run & Qwen-Image-2512 (LoRA) \\
\midrule
\rowcolor{white}\multicolumn{10}{@{}l}{\;\,\ding{60} \textit{Frozen self-verifiers}} \\
PRISM~\cite{prism} {\tiny arXiv'26} & \rl{B} & community prompts, 8 rewrites each, frozen SD3.5-M & 8 & \rl{A} & initial VLM\textsuperscript{f}, dimension scores & \rl{B} & GSPO & 2 & Qwen3-VL-8B \\
MT-OPSD~\cite{mtopsd} {\tiny arXiv'26} & \rl{B} & OmniEdit pairs, self-generated editing states & -- & \rl{A} & clean-conditioned checkpoint\textsuperscript{f}, gated promotion & \rl{B} & on-policy self-distillation & 1 run & three editing backbones \\
ManimAgent~\cite{manimagent} {\tiny arXiv'26} & \rl{B} & stream of 39 lesson tasks & -- & \rl{E}+\rl{A} & renderer errors, keyframe review\textsuperscript{s} & \rl{A} & gated success and failure memory (M) & 39 tasks & GPT-5.5, frozen \\
COMFYCLAW~\cite{comfyclaw} {\tiny arXiv'26} & \rl{B} & GenEval 2 and DPG-Bench prompts & -- & \rl{A} & region checklist\textsuperscript{s}, held-out gain gate & \rl{A} & skill library (T) & -- & Claude, Qwen, Gemma, frozen \\
SceneCraft~\cite{scenecraft} {\tiny ICML'24} & \rl{B} & synthetic scene queries & -- & \rl{A} & review of renders\textsuperscript{s} against parsed constraints & \rl{A} & constraint-function library (T) & batches & GPT-4V, frozen \\
EvoDesign~\cite{evodiagram} {\tiny arXiv'26} & \rl{B} & CanvasBench queries, alternative requests & -- & \rl{A} & refiner critique\textsuperscript{s} of each render & \rl{A} & three-tier design memory (M+I) & 5 & GPT-5.2, Gemini-3-Pro, frozen \\
SymbOmni~\cite{symbomni} {\tiny arXiv'26} & \rl{B} & ComfyUI generation and editing instructions & -- & \rl{E}+\rl{A} & compiler errors, evaluation agent\textsuperscript{s} & \rl{A} & symbolic concept memory (T+I) & online & Gemini 2.5 Flash, frozen \\
SIDiffAgent~\cite{sidiffagent} {\tiny arXiv'26} & \rl{B} & GenAI-Bench prompts with edit loops & -- & \rl{A} & evaluation agent\textsuperscript{s}, aesthetics and alignment & \rl{A} & trajectory memory (M) & 2 & Qwen2.5-VL-72B, frozen \\
Drawing Middleware~\cite{figagent} {\tiny arXiv'26} & \rl{B} & paper and method-illustration pairs & -- & \rl{A}+\rl{B} & evaluator agent\textsuperscript{s}, reference illustrations & \rl{A} & middleware library with rollback (T) & -- & GPT-5, frozen \\
MemoGen~\cite{memogen} {\tiny arXiv'26} & \rl{B} & benchmark task set rerun each round & -- & \rl{A}+\rl{B} & controller self-review, independent memory judge & \rl{A} & experience memory (M) & 3 & Qwen-Image, GPT-5.5, frozen \\
PhysRepair~\cite{physrepair} {\tiny arXiv'26} & \rl{B} & physics benchmark cases processed in sequence & -- & \rl{E}+\rl{A} & re-diagnosis\textsuperscript{s} with perception tools & \rl{A} & repair experience memory (M) & case stream & GPT-6-Astra, MiniMax H3, frozen \\
\midrule
\rowcolor{white}\multicolumn{10}{@{}l}{\;\,\ding{60} \textit{External or programmatic verifiers}} \\
RSI in UMMs~\cite{rsiumm} {\tiny arXiv'26} & \rl{A} & chart specifications planned from self-diagnosed failures & 2 & \rl{E} & execution of own programs against the specification & \rl{B} & fine-tuning of generation, understanding, programs & 4 & BAGEL-7B-MoT \\
MATE~\cite{mate} {\tiny arXiv'26} & \rl{A} & branches pose challenges from evolving input pools & -- & \rl{B} & frozen UnifiedReward scorer, SigLIP source consistency & \rl{B} & GRPO on both branches & 5 epochs & Janus-Pro-1B \\
\bottomrule
\end{tabularx}
\vspace{-3mm}
\end{table*}

\begin{table*}[!t]
\ContinuedFloat
\centering
\scriptsize
\renewcommand{\arraystretch}{1}
\setlength{\tabcolsep}{3.5pt}
\caption[]{\textbf{Self-evolving methods for multimodal generation} (continued).}
\vspace{-2mm}
\rowcolors{3}{themelight}{white}
\renewcommand{\tabularxcolumn}[1]{m{#1}}
\begin{tabularx}{\textwidth}{>{\raggedright\arraybackslash}m{2.45cm}>{\centering\arraybackslash}m{0.35cm}>{\raggedright\arraybackslash}X>{\centering\arraybackslash}m{0.45cm}>{\centering\arraybackslash}m{0.65cm}>{\raggedright\arraybackslash}X>{\centering\arraybackslash}m{0.35cm}>{\raggedright\arraybackslash}m{2.2cm}>{\centering\arraybackslash}m{0.8cm}>{\raggedright\arraybackslash}m{1.9cm}}
\toprule
\rowcolor{gray!25}
\rule{0pt}{2.6ex}& \multicolumn{3}{c}{\textbf{Experience Generation (EG)}} & \multicolumn{2}{c}{\textbf{Feedback Acquisition (FA)}} & \multicolumn{3}{c}{\textbf{Update Consolidation (UC)}} & \\
\noalign{\global\aboverulesep=0pt\global\belowrulesep=0pt}
\arrayrulecolor{gray!25}\cmidrule{1-1}\cmidrule{10-10}\arrayrulecolor{black}\cmidrule(lr){2-4}\cmidrule(lr){5-6}\cmidrule(lr){7-9}
\noalign{\global\aboverulesep=0.4ex\global\belowrulesep=0.65ex}
\rowcolor{gray!25}
\multirow{-2}{*}{\textbf{Method}}
& \rule[-1ex]{0pt}{2.8ex}$\boldsymbol{\sigma}$
& \textbf{Goal content} & $\boldsymbol{K_k}$
& $\boldsymbol{V}$
& \textbf{Scoring reference \& signal}
& $\boldsymbol{f}$
& \textbf{Update}
& \textbf{Rounds}
& \multirow{-2}{*}{\textbf{Backbone}} \\
\midrule
RSIgen~\cite{rsigen} {\tiny arXiv'26} & \rl{A} & curricula compiled from own probes & -- & \rl{B} & fixed external evaluator & \rl{B} & agent and generator updates & 3 attempts & BAGEL, Qwen-Image, Wan-Image \\
OctoT2I~\cite{octot2i} {\tiny CVPR'26} & \rl{A} & self-proposed prompts over concept dimensions & 5 & \rl{B} & frozen MLLM probability of alignment & \rl{A} & routing knowledge base (I) & 1 pass & Qwen2-0.5B router \\
GenEvolve~\cite{genevolve} {\tiny arXiv'26} & \rl{B} & prompts with judge-extracted experience & 6 & \rl{B} & Gemini KScore judge, program-sufficiency judge & \rl{A} & GRPO, self-distillation from own experience-conditioned targets & -- & Qwen3-VL-8B, Qwen-Image-Edit \\
\bottomrule
\end{tabularx}
\end{table*}

\subsection{FA Bound by Attainable Fidelity}\label{sec:mg_feedback}\label{sec:mg_anchors}

Each cycle hands feedback acquisition a group of outputs for the same prompt or the renders of one workflow, and the loops score each output as a whole (\cref{tab:mg_taxonomy}).
The scores come from the agent itself (\rl{A}), through the generator's confidence, its understanding branch, a lagged, snapshot, or co-trained judge, or the frozen backbone that runs the workflow, or from outside it, through an external judge (\rl{B}) or program execution (\rl{E}).
Weight-updating methods mostly turn scores into group-normalized rewards for GRPO-style updates or into preference pairs for DPO~\cite{solace,iris,corlulm,silmm,srlvlm}, whereas training-free methods receive verbal critiques, checklist scores, or pass/fail verdicts~\cite{scenecraft,evodiagram,comfyclaw,symbomni} that trigger repairs within an episode and gate persistent writes.

\pgraphhlY{Fidelity and Localization Constraints}\; Feedback acquisition is bound by the fidelity ceiling (\cref{tab:stage2_dims}), to which signal source, channel coverage, and gaming resistance each contribute, while localization imposes little.
An image admits no equivalence check against a fixed reference, leaving learned scorers and model judges, whose agreement with humans on open-ended content can turn negative, the frozen reviewer of ManimAgent correlating with blind ratings of 37 educational videos at $r\!=\!-0.17$~\cite{manimagent}.
Instruments fixed before the loop score parts of the objective, such as detectors, OCR, and PickScore in Flow-GRPO~\cite{flowgrpo} and HPS-v2.1, CLIP, and VideoAlign in DanceGRPO~\cite{dancegrpo}, and Flow-GRPO raises PickScore and OCR accuracy while human raters judge its images below the base model in quality~\cite{imgownreward}.
Under optimization, IRIS attributes the realism drop of its HPSv2-trained baseline to reward hacking~\cite{iris}.
Only sub-families with a closed-form predicate escape the ceiling, RSI in UMMs gaining more on charts from execution-verified data than from the model's own best-rated images~\cite{rsiumm}.
Localization stays near $L_\nu\!=\!1$, a single sampling pass forming one action and tool-orchestrated systems scoring renders with execution errors as their chief step-level signal~\cite{manimagent,symbomni}.

\pgraphhlP{Verifier Dependence on the Policy}\; With localization nearly free, every deficit of feedback acquisition is a fidelity deficit, and the methods differ in how far the verifier depends on the policy it scores.
A verifier that depends on the current policy keeps the agent edge into FA, letting $F_\nu^{(k)}$ move with each update and inviting compounding reward hacking, whereas weaker dependence slows or cuts that edge and leaves a bias fixed across cycles.
Generator confidence depends on the policy entirely, and SOLACE and IRIS disagree even on its sign, rewarding higher denoising confidence and lower self-certainty, respectively~\cite{solace,iris}.
Understanding-branch checks hold only the prompt fixed and substitute for an external verifier only if the judging parameters are separated from the judged ones, the two directions are symmetric, and understanding exceeds generation.
Shared weights violate the first condition, which KL penalties to a reference policy slow but do not remove~\cite{suder,unirl,asg,corlulm}, the non-inverse relation between text-to-image and image-to-text weakens the second behind reverse-likelihood rewards~\cite{suder,readitback,gvu}, and the third, the internal gap that InternalGap converts into supervision~\cite{internalgap}, remains untested within a run~\cite{suder}.
Weaker dependence comes from co-evolving or lagged verifiers~\cite{srlvlm,unievo,imgownreward,dualipo,imagersi}, from frozen self-verifiers~\cite{prism,mtopsd}, including frozen backbones that also write the update~\cite{manimagent,comfyclaw,scenecraft}, and from external or programmatic verifiers~\cite{mate,rsigen,rsiumm}, and no report compares these designs at a matched number of rounds.
Such designs thus trade a moving bias for a fixed one without raising $\bar{F}$.

\subsection{UC Bound by Retention}\label{sec:mg_uc}

Update consolidation in multimodal generation persists the scored experience in two forms, weight updates and training-free writes (\cref{tab:mg_taxonomy}).
Weight-updating methods write $\Phi_\theta$ with an update rule fixed before the loop, through group-relative policy gradients, preference optimization, fine-tuning, or self-distillation, either continuously within a single training run or over two to five reported rounds~\cite{srlvlm,prism,silmm,dualipo,rsiumm,safecodebook}.
Training-free methods keep the generator and the agent frozen and let the agent write memory~\cite{manimagent,memogen,sidiffagent,evodiagram,physrepair}, skill, constraint-function, concept, or middleware libraries~\cite{comfyclaw,scenecraft,symbomni,figagent}, or routing records~\cite{octot2i}, over a task stream or up to five rounds.
ImageRSI alone combines the two, admitting experience to a library while retraining its judge and not its generator~\cite{imagersi}.

\pgraphhlY{Retention and Generalization Constraints}\; Update consolidation is bound by retention (\cref{tab:stage3_dims}), documented within runs and in RSI in UMMs over at most four rounds, whereas generalization is mostly untested and not shown to fail.
Only RSI in UMMs measures retention on earlier-round goals, exact category text on an earlier vocabulary falling from 72\% to 60\% once a round replaces that vocabulary and recovering when later rounds keep half of the old specifications, while general understanding, which the chart goals do not exercise, erodes by 2.6 points on MMBench~\cite{rsiumm}.
Within single runs, iteration robustness fails at the grain of optimizer steps, SOLACE degenerating into textureless images under prolonged optimization~\cite{solace} and IRIS into meaningless images within hundreds of steps~\cite{iris}, and across rounds, per-round scores cannot separate saturation from erosion, as in the slight slip of DreamLLM under SILMM on DPG-Bench~\cite{silmm}.
Generalization is tested only where a loop holds out evaluation goals~\cite{iris,imagersi,evodiagram}, which MemoGen does not~\cite{memogen}, and transfer is partial for both forms of writing, a memory that SIDiffAgent builds on GenAI-Bench raising DrawBench from 0.860 to 0.873 against 0.901 for in-domain memory~\cite{sidiffagent} and the parametric writes of RSI in UMMs gaining only from 11.0\% to 11.9\% on a chart subset its authors judge too hard for open models~\cite{rsiumm}.
Understanding-branch gains also remain entangled with the prior that judge and generator share in a unified model.

\pgraphhlP{Self-Judged Writes}\; The exposure that consolidation faces comes from feedback acquisition, where the agent itself judges most writes, and the two forms of writing guard against it differently.
Where the verifier depends on the policy, a parametric write is scored by the policy itself or by a judge that follows it, and the update rule admits every rewarded step, held back only by regularizers such as KL penalties~\cite{suder,unirl,asg,solace,iris,corlulm}.
Training-free writes pass an admission test before they persist, yet the reviewer gating of ManimAgent~\cite{manimagent} and the no-loss check of COMFYCLAW on three held-out prompts that the agent synthesizes~\cite{comfyclaw} reuse the frozen model that judged the experience, protecting retention on that model's scale but not against its bias.
Only ImageRSI, through paired validation on separate calibration and holdout splits~\cite{imagersi}, and Drawing Middleware Evolution, through rollback against reference illustrations~\cite{figagent}, test writes outside that model.
Retention is thus protected only where a training-free write is tested outside the model that judged it, a condition that the key- or validation-gated writes of multimodal reasoning meet more often.

\subsection{Limits and Preconditions}\label{sec:mg_positioning}

\pgraph{Bottleneck Attribution}\; In multimodal generation, the fidelity ceiling binds before self-reference, which changes the form of the deficit and leaves the ceiling in place.
The fixed loops of the scenario, Flow-GRPO~\cite{flowgrpo} and DanceGRPO~\cite{dancegrpo} among them, already train against exploitable proxies~\cite{iris}, unlike the key-scored fixed loops of multimodal reasoning~\cite{mstar}.
Weight-updating loops replace a proxy whose bias stays put with a policy-dependent verifier whose bias moves with the policy~\cite{solace,iris}, and training-free loops let the frozen model that judges an output also write the update, judge and writer sharing one bias.
Goal selection is where self-reference earns back part of this cost, controlled returns being reported there, whereas The Image as Its Own Reward offers the only comparison of a moving verifier with a fixed one, by human preference after one run~\cite{imgownreward}.
Both the ceiling and its cost rest on stage evidence and the provenance in \cref{tab:mg_taxonomy}, not on a controlled comparison.

\pgraph{Cross-Cycle Degradation}\; Across cycles, self-reference opens paths along which degradation follows the agent, with only one degradation measured across rounds.
On the goal side, learned proposers extend the agent edge into $\sigma_\omega$, and the regeneration of failed prompts in InternalGap~\cite{internalgap} and the pruning of prompts by pass rate in OctoT2I~\cite{octot2i} open the goal edge, both inviting curriculum collapse.
The goal edge also carries this one measured degradation, the forgetting in RSI in UMMs following a vocabulary change that a rule reading the agent's OCR score triggers~\cite{rsiumm}.
On the feedback side, self-verifiers extend the agent edge into FA, inviting compounding reward hacking, observed only within single runs~\cite{solace,iris}, and training on self-generated images invites model collapse, which the narrowing of IRIS outputs under higher image self-certainty resembles without isolating~\cite{iris}.
Curriculum collapse, the extension into UC, and cross-round self-verifier fidelity remain untested.

\pgraph{Presupposed Resources}\; The mechanisms of multimodal generation presuppose resources that only part of its outputs supply.
Understanding-branch checks presuppose an output that the agent's own perception can process and an understanding branch stronger than its generation~\cite{suder,internalgap}, built-in checks and execution-verified goals presuppose a sub-family with a closed-form predicate~\cite{unirl,rsiumm}, and admission gates presuppose held-out splits or reference sets outside the judging model~\cite{imagersi,figagent}.

\section{Self-Evolution in GUI and Web Navigation}
\label{sec:gw}

GUI and Web navigation occupies the quadrant of \cref{fig:positioning_matrix} in which only the quantity ceiling is low, each rollout being a multi-step episode that pays environment latency at every step, whereas the inspectable end state of an application or a web page keeps the attainable fidelity high in sandboxed environments, where per-task scripts read it.
Much automated online learning here runs as fixed loops, whose tasks and verifiers come from external models or benchmark authors and never read the agent's record~\cite{zerogui,mobileguirl,openwebvoyager,guiowl,computerrl,arpo,craftgui,wmr1,atmem,mobileforge,stamp,autoscalemem,scaffold}.
Weight-updating loops allocate scarce rollouts by proposing goals or reweighting a fixed task set by its success record, and buy step-level credit with verifiers refit on the agent's trajectories, whose fidelity moves with the policy.
Training-free loops keep the task set fixed and reuse each episode through memory or skills that a frozen backbone writes and judges, sharing one bias between judge and writer.
Reported loops span one training run, a few rounds, or one pass over a task stream, mostly evaluated on the task sets they train on.

\begin{table*}[!t]
\centering
\scriptsize
\renewcommand{\arraystretch}{1}
\setlength{\tabcolsep}{3.5pt}
\caption{\textbf{Self-evolving methods for GUI and Web navigation.}
Methods are grouped by whether and how the goal sampler reads the agent's record, and each loop component is tagged by its realizing party: \rl{A} agent and \rl{B} bootstrap.
Notation: \textsuperscript{c} co-trained and \textsuperscript{s} same-model evaluator; M memory, T tool, and I instruction update; \textit{1 run} continuous updating within one run; \textit{1 pass} a single pass over the goals; -- not reported.
}
\label{tab:gw_taxonomy}
\vspace{-2mm}
\rowcolors{3}{themelight}{white}
\renewcommand{\tabularxcolumn}[1]{m{#1}}
\begin{tabularx}{\textwidth}{>{\raggedright\arraybackslash}m{2.45cm}>{\centering\arraybackslash}m{0.55cm}>{\raggedright\arraybackslash}X>{\centering\arraybackslash}m{0.55cm}>{\centering\arraybackslash}m{0.65cm}>{\raggedright\arraybackslash}X>{\centering\arraybackslash}m{0.35cm}>{\raggedright\arraybackslash}m{2.2cm}>{\centering\arraybackslash}m{0.95cm}>{\raggedright\arraybackslash}m{2.0cm}}
\toprule
\rowcolor{gray!25}
\rule{0pt}{2.6ex}& \multicolumn{3}{c}{\textbf{Experience Generation (EG)}} & \multicolumn{2}{c}{\textbf{Feedback Acquisition (FA)}} & \multicolumn{3}{c}{\textbf{Update Consolidation (UC)}} & \\
\noalign{\global\aboverulesep=0pt\global\belowrulesep=0pt}
\arrayrulecolor{gray!25}\cmidrule{1-1}\cmidrule{10-10}\arrayrulecolor{black}\cmidrule(lr){2-4}\cmidrule(lr){5-6}\cmidrule(lr){7-9}
\noalign{\global\aboverulesep=0.4ex\global\belowrulesep=0.65ex}
\rowcolor{gray!25}
\multirow{-2}{*}{\textbf{Method}}
& \rule[-1ex]{0pt}{2.8ex}$\boldsymbol{\sigma}$
& \textbf{Goal content} & $\boldsymbol{K_k}$
& $\boldsymbol{V}$
& \textbf{Scoring reference \& signal}
& $\boldsymbol{f}$
& \textbf{Update}
& \textbf{Rounds}
& \multirow{-2}{*}{\textbf{Backbone}} \\
\midrule
\rowcolor{white}\multicolumn{10}{@{}l}{\;\,\ding{60} \textit{Stationary sampling of fixed goals}} \\
Mobile-Agent-E~\cite{mobileagente} {\tiny arXiv'25} & \rl{B} & five-task scenario sequences & -- & \rl{A} & Experience Reflectors over action and error histories\textsuperscript{s} & \rl{A} & Tips and Shortcuts (M, T) & 1 pass & GPT-4o, frozen \\
Agent S~\cite{agents} {\tiny ICLR'25} & \rl{B} & exploration tasks, then OSWorld tasks & -- & \rl{A} & self-evaluator summaries of each task & \rl{A} & narrative and episodic memory (M) & task stream & GPT-4o, frozen \\
MobileSteward~\cite{mobilesteward} {\tiny KDD'25} & \rl{B} & cross-app instructions & -- & \rl{A} & StewardAgent evaluation\textsuperscript{s} & \rl{A} & expertise and guideline memory (M) & task stream & GPT-4o, frozen \\
AppAgentX~\cite{appagentx} {\tiny arXiv'25} & \rl{B} & DroidTask, AndroidWorld, and A3 tasks, five repetitions & -- & \rl{B} & benchmark success evaluation & \rl{A} & shortcut graph (M, T) & 5 & GPT-4o, frozen \\
WebCoach~\cite{webcoach} {\tiny ICLR'26} & \rl{B} & WebVoyager stream of 643 tasks & -- & \rl{A} & actor self-evaluation, condensed by Qwen3-8B & \rl{B} & cross-session episodic memory (M) & 1 pass & Qwen2.5-VL, Skywork-r1v3, frozen \\
EvoSkill-GUI~\cite{evoskillgui} {\tiny arXiv'26} & \rl{B} & AndroidWorld, MobileWorld, and OSWorld tasks & -- & \rl{A}+\rl{B} & isolated same-backbone critic, terminal task reward & \rl{A} & skill packages (M, T) & $\le$2 revisions per task & Claude, Qwen3.6, MAI-UI, frozen \\
MAGNET~\cite{magnet} {\tiny arXiv'26} & \rl{B} & AndroidWorld tasks, memory seeded from Amex & -- & \rl{B} & AndroidWorld success scripts & \rl{A} & procedural and grounding memory (M) & 3 & Qwen2.5-VL-32B, frozen \\
DigiRL~\cite{digirl} {\tiny NeurIPS'24} & \rl{B} & AitW instructions & -- & \rl{A}+\rl{B} & VLM evaluator with 2.8\% error, value functions filtering steps\textsuperscript{c} & \rl{B} & offline-to-online RL with replay & 1 run & AutoUI \\
MagicGUI-RMS~\cite{magicguirms} {\tiny arXiv'26} & \rl{B} & real-world mobile tasks, 2,154 episodes per round & -- & \rl{A}+\rl{B} & domain reward model on the agent backbone\textsuperscript{c}, GPT-4o arbitration & \rl{B} & SFT of agent and reward model & 2+2 & Qwen3-VL-8B \\
Agent S4~\cite{agents4} {\tiny OpenReview'26} & \rl{B} & synthetic OSWorld V2 training tasks & -- & \rl{B} & OSWorld V2 validation evaluators select parents & \rl{A} & evolved interfaces and usage knowledge (T, I) & 15 & Qwen3.8-27B, frozen \\
\midrule
\rowcolor{white}\multicolumn{10}{@{}l}{\;\,\ding{60} \textit{Adaptive sampling of fixed goals}} \\
DART-GUI~\cite{dartgui} {\tiny arXiv'25} & \rl{A} & OSWorld tasks, rollout counts and length caps set by success rate & 8 & \rl{B} & OSWorld evaluation scripts & \rl{B} & decoupled multi-turn RL & 1 run & UI-TARS-1.5-7B \\
MobileRL~\cite{mobilerl} {\tiny ICLR'26} & \rl{A} & AndroidWorld tasks down-weighted after consecutive zero-reward epochs & 16 & \rl{B} & rule-based AndroidWorld rewards, fine-tuned VLM for AndroidLab & \rl{B} & AdaGRPO with positive replay & 1 run & Qwen2.5-VL-7B, GLM-4.1V-9B \\
UI-Genie~\cite{uigenie} {\tiny NeurIPS'25} & \rl{A}+\rl{B} & failed tasks of earlier rounds, manually crafted complex scenarios & 10, top 5 & \rl{A} & reward model fit on continuation-rollout step labels\textsuperscript{c} & \rl{B} & SFT of agent and reward model & 3 & 3B, 7B, 72B MLLMs \\
\midrule
\rowcolor{white}\multicolumn{10}{@{}l}{\;\,\ding{60} \textit{Adaptive generation of new goals}} \\
SEAgent~\cite{seagent} {\tiny ICML'26} & \rl{A} & tasks written from a guidebook of judged trajectories & 8 & \rl{B} & World State Model trained on GPT-4o judgments, step-level & \rl{B} & GRPO and adversarial imitation & 3 & UI-TARS-7B-DPO \\
ACuRL~\cite{acurl} {\tiny arXiv'26} & \rl{A} & tasks generated from agent experience and judge feedback & -- & \rl{B} & CUAJudge outcome judgment from key screenshots & \rl{B} & RL & 3 & UI-TARS-1.5-7B, Qwen3-VL-8B \\
SCALE~\cite{scale} {\tiny CVPR'26} & \rl{A} & website actions whose outcomes the agent is least sure of & -- & \rl{A} & predicted against observed next screen\textsuperscript{s} & \rl{B} & fine-tuning of Selector and Predictor & 20--60 hops & Qwen2.5-VL-7B, InternVL2.5-8B \\
SkillWeaver~\cite{skillweaver} {\tiny arXiv'25} & \rl{A} & skills proposed beyond the current library & -- & \rl{A} & success check from trajectory and final screenshot\textsuperscript{s} & \rl{A} & executable API library (T) & 160 per site & GPT-4o, frozen \\
\bottomrule
\end{tabularx}
\end{table*}

\subsection{EG Bound by Rollout Cost}\label{sec:gw_eg}\label{sec:gw_goals}

In GUI and Web navigation, experience generation draws goals that pair a natural-language instruction with the initial state of an application, desktop, or website, and each rollout is an episode of screenshots, often with an accessibility tree, and actions such as taps, typing, and scrolls, ending in success, failure, or a step limit.
The task families span mobile apps in AitW, AndroidWorld, and AndroidLab~\cite{digirl,mobilerl,uigenie}, desktop software in OSWorld~\cite{dartgui,seagent,agents}, and web navigation in WebArena and VisualWebArena sandboxes and on live WebVoyager and Online-Mind2Web sites~\cite{skillweaver,scale,webcoach}.
These goals come from a benchmark task set, a task stream, or the agent's own proposals, and weight-updating loops run episode groups per goal in parallel instances, whereas training-free loops run one or a few episodes in sequence with a frozen backbone.
\Cref{tab:gw_taxonomy} summarizes the self-evolving methods that satisfy the conditions of \cref{sec:defining_conditions}.

\pgraphhlY{Quantity and Quality Constraints}\; Experience generation is bound by quantity at the episodes that parallel environment instances afford (\cref{tab:stage1_dims}), an episode cost that the single passes of multimodal reasoning and generation do not incur.
A rollout spans up to 25 screen-action steps~\cite{mobileguirl}, each paying rendering and environment latency on top of policy inference, 2.5 to 4.0 seconds per step on the virtual machines of UI-TARS-2~\cite{uitars2}, with DART-GUI reporting episodes that last tens of minutes~\cite{dartgui}.
Group sizes of eight to 64 rollouts per task~\cite{mobileguirl,dartgui,mobilerl,zerogui} are bought with parallel instances, up to 64 emulators in DigiRL~\cite{digirl}, whereas training-free loops run each task once or a few times in sequence~\cite{mobileagente,webcoach,appagentx,evoskillgui}.
Quality decides what each scarce episode returns, and on long-horizon tasks the loss falls mostly on goals beyond the agent's reach, where a binary outcome gives $p_g(1-p_g)\approx 0$, consistent with starting success rates of 11.3\% on the professional software of SEAgent~\cite{seagent} and 18.1\% on AndroidLab in UI-Genie~\cite{uigenie}.

\pgraphhlP{Reuse, Reweighting, and Goal Generation}\; A larger group costs environment instances, and loops raise $\mathbb{E}[I_{\mathcal{A}_k}]$ instead along three routes, reuse, reweighting, and goal generation, which the inspectable end state keeps verifiable.
Under stationary sampling, reuse raises the return of later episodes by conditioning them on memory and skills that the agent writes~\cite{mobileagente,agents,magnet,mobilesteward,appagentx,webcoach,evoskillgui,agents4} or by training the agent on steps that value functions or reward models fit on its rollouts select and label~\cite{digirl,magicguirms}, both lifting unreachable tasks toward informative success rates.
Reweighting by the success record keeps the verifier of a fixed task set, lowering the rollout count of tasks with high running success in DART-GUI~\cite{dartgui} and the sampling probability of tasks that earn no reward over consecutive epochs in MobileRL~\cite{mobilerl}, or returning earlier failures~\cite{uigenie}.
Goal generation needs a verifier for each new goal, which reads the outcome off the screen through a model or judge fixed before the loop~\cite{seagent,acurl} or a check that the agent itself runs~\cite{scale,skillweaver}.
Each route carries a controlled return, and the most direct test of substitution favors reuse, EvoSkill-GUI with skills outperforming pass@3 without them on MobileWorld at a comparable token budget~\cite{evoskillgui}, while removing the failure filter of MobileRL lowers its AndroidWorld success~\cite{mobilerl} and random task generation lowers ACuRL's third-round LibreOffice Impress score from 40.7\% to 36.4\%~\cite{acurl}.

\subsection{FA Bound by Localization}\label{sec:gw_feedback}

Each cycle hands feedback acquisition an episode of screens and actions, which most loops score once at its end and loops with learned step judges score per step (\cref{tab:gw_taxonomy}).
Scores from outside the agent (\rl{B}) come from per-task scripts and rule-based rewards written with the benchmark~\cite{dartgui,mobilerl}, the fixed evaluators of DigiRL and SEAgent~\cite{digirl,seagent}, and benchmark success~\cite{appagentx}.
Scores from the agent (\rl{A}) come from value functions and reward models fit on its trajectories~\cite{digirl,uigenie,magicguirms}, the Judger role of the base model that SCALE fine-tunes in its other roles~\cite{scale}, and the frozen backbone that acts and judges its own episode in training-free loops~\cite{mobileagente,agents,mobilesteward,webcoach,evoskillgui,skillweaver}.
Weight-updating methods turn scores into group-normalized advantages or rejection-sampled training targets~\cite{dartgui,mobilerl,uigenie,magicguirms}, whereas training-free methods turn them into success labels and diagnoses that gate memory and skill writes~\cite{agents,evoskillgui,webcoach}.

\pgraphhlY{Fidelity and Localization Constraints}\; Feedback acquisition is bound by localization (\cref{tab:stage2_dims}), the attainable fidelity being high wherever the goal fixes an end state, though signal sources reach that ceiling unevenly.
Task completion resides in the final screen, application database, or file system, which per-task scripts and rule-based rewards inspect directly~\cite{dartgui,mobilerl}, whereas model judges fall short, EvoSkill-GUI's same-backbone critic erring on 28 of 143 audited failures and passing at least four as successes~\cite{evoskillgui}.
Free-form goals and live websites lack scripts and fall to screenshot judges~\cite{zerogui,openwebvoyager,skillweaver}, which see the screen but not the backend state.
Under optimization, ACuRL audits its judge against on-policy trajectories over three rounds, with precision of 89.1\% to 98.0\%~\cite{acurl}, whereas judge benchmarks score fixed trajectories~\cite{agentrewardbench} or the downstream success of policies trained on each judge~\cite{mobilejudgebench}.
Localization binds every source alike, even an exact checker spreading one verdict over episodes of up to 50 actions~\cite{evoskillgui}, with $L_\nu$ falling as episodes lengthen.
Loops recover step-level signal at a credit-assignment cost through learned value functions and judges~\cite{digirl,seagent,uigenie,magicguirms} or a failure-diagnosing critic for training-free revision~\cite{evoskillgui}, step-level advantages speeding DigiRL's learning by about 12\% on its Web Shopping subset~\cite{digirl}.

\pgraphhlP{Localization Bought with Fidelity}\; With end states checkable, feedback acquisition falls short in localization and lacks checkers for free-form goals, and loops open the agent edge into FA to buy either.
A value function or reward model refit on the agent's trajectories follows the policy, letting $F_\nu^{(k)}$ move with each round and inviting compounding reward hacking, exploited errors re-entering the data on which it is refit~\cite{digirl,uigenie,magicguirms}.
MagicGUI-RMS slows this edge with a GPT-4o reward model that arbitrates its domain model~\cite{magicguirms}, yet both it and UI-Genie track their reward models only on fixed labeled sets, excluding the on-policy rollouts where exploitation would surface~\cite{uigenie}.
A frozen self-judge keeps its bias fixed but shares it with the writer~\cite{mobileagente,agents,mobilesteward,evoskillgui,skillweaver}, its errors cutting revision short, as the over-optimistic EvoSkill-GUI critic shows.
Verifiers fixed before the loop cut the edge, at the episode level in DART-GUI and MobileRL~\cite{dartgui,mobilerl} and at the step level in SEAgent's World State Model~\cite{seagent}, which localizes credit independently of the agent but exposes the loop to distribution shift once rollouts leave its training distribution.
Localization and free-form goal coverage thus cost either fidelity that follows the agent or, from a model fixed before the loop, distribution shift.

\subsection{UC Bound by Generalization}\label{sec:gw_uc}

Update consolidation in GUI and Web navigation persists the scored experience in two forms, weight updates and training-free writes, each method using one of them (\cref{tab:gw_taxonomy}).
Weight-updating methods write $\Phi_\theta$ with an update rule fixed before the loop, either continuously within a single training run~\cite{digirl,dartgui,mobilerl} or over two or three rounds through GRPO with adversarial imitation, reinforcement learning on generated tasks, or fine-tuning of the agent and its reward model~\cite{seagent,acurl,magicguirms,uigenie}, while SCALE fine-tunes its Selector and Predictor over repeated exploration rounds~\cite{scale}.
Training-free methods keep the backbone frozen and let the acting model write narrative, episodic, procedural, or grounding memory~\cite{agents,mobilesteward,magnet}, tips and executable shortcuts~\cite{mobileagente,appagentx}, or skill packages and executable APIs~\cite{evoskillgui,skillweaver}, over one pass of a task stream, a few attempts per task, or a fixed budget of practice iterations, whereas a separate condenser writes the memory of WebCoach~\cite{webcoach}.

\pgraphhlY{Retention and Generalization Constraints}\; Update consolidation is bound by generalization (\cref{tab:stage3_dims}), documented within one platform and mostly on the task sets that produced the experience, whereas general capability, where measured, holds.
No method measures retention on earlier-round goals, and the closest evidence concerns general capability, ACuRL keeping or raising its scores on untrained environments by 2 to 7 points~\cite{acurl}.
Elsewhere, replay sustains iteration robustness~\cite{digirl,mobilerl}, and per-round scores rise without a reported reversal~\cite{uigenie,magicguirms,scale}, though non-monotonically in Mobile-Agent-E~\cite{mobileagente}, tracking $\mathcal{J}_k$ on the current evaluation set, not $S_\zeta$.
Training-free libraries evolve on their evaluation tasks~\cite{evoskillgui,mobileagente,webcoach}, as in several fixed loops~\cite{uivoyager,uimem,traces2skills}, measuring adaptation to a known goal distribution.
Held-out applications show transfer within the platform, MagicGUI-RMS gaining about as much on unseen apps as in domain~\cite{magicguirms}, whereas ACuRL reaches a new software version or operating system only by training there~\cite{acurl}.
Zero-shot transfer across platforms appears only in the fixed loop GUI-HARVEST, whose frozen harness, evolved on OSWorld, adds 6.47 points for Qwen3-VL-32B-Instruct and 13.87 for GPT-5 on WindowsAgentArena~\cite{guiharvest}.
Cross-form comparisons are confounded by backbone, training-free loops mostly running frozen proprietary models and weight-updating loops open ones.

\pgraphhlP{Author Checks and Reader Dependence}\; What a write carries into the agent is fixed in feedback acquisition, and the two forms of writing differ in what checks a write before it persists and, for non-parametric writes, in who reads it afterward.
Parametric writes pass on the verifier's score alone, carrying its exposure unchanged, compounding hacking for agent-derived verifiers~\cite{digirl,uigenie,magicguirms,scale} and distribution shift for SEAgent's frozen World State Model~\cite{seagent}, while exact checkers leave little to inherit~\cite{dartgui,mobilerl}.
Non-parametric writes pass an admission test that mostly reuses the author's judgment, EvoSkill-GUI and MobileSteward admitting skills and memories largely on their own critics' verdicts~\cite{evoskillgui,mobilesteward}, whereas SkillWeaver unit-tests each synthesized API against execution feedback, a check outside the writer~\cite{skillweaver}.
Self-written skills can carry errors, such as the redundant or missing taps in the generated shortcuts of Mobile-Agent-E~\cite{mobileagente}, unlike the key- or validation-gated writes of multimodal reasoning.
Once admitted, a non-parametric write takes effect only when a later rollout reads it, its gain depending on the reader, as the memory that a separate condenser writes for WebCoach lowers Qwen2.5-VL-7B from 32.8\% to 31.1\% yet raises Skywork-r1v3-38B from 47.3\% to 61.4\%~\cite{webcoach}.
A non-parametric write is thus checked mostly by its author and valued by its reader.

\subsection{Limits and Preconditions}\label{sec:gw_positioning}

\pgraph{Bottleneck Attribution}\; In GUI and Web navigation, the quantity ceiling binds fixed and self-evolving loops alike, and self-reference leaves it in place, its cost lying in what each entry point puts at risk.
Fixed loops pay the same episode cost, scaling to thousands of virtual machines in ComputerRL~\cite{computerrl}, and already train against signals fixed before the loop, the verification functions of ComputerRL at high fidelity and, below it, the screen judges of ZeroGUI~\cite{zerogui} and MobileGUI-RL~\cite{mobileguirl}.
A success-conditioned rule over a task set with fixed checkers risks little, only the allocation following the agent~\cite{mobilerl}, whereas generated goals move verification from fixed checkers to learned or prompted judges~\cite{seagent,acurl,scale,skillweaver}, refit value functions and reward models buy localization with a fidelity that moves with the policy~\cite{digirl,uigenie,magicguirms}, and a frozen backbone that judges and writes carries one bias into both roles~\cite{agents,evoskillgui}.
The higher return of each scarce episode recovers part of these costs, controlled gains being reported for reuse, reweighting, and goal generation.
No controlled comparison with fixed loops isolates these costs, which we read off the stage evidence and the provenance in \cref{tab:gw_taxonomy}.

\pgraph{Cross-Cycle Degradation}\; Across cycles, self-reference opens paths along which degradation follows the agent, and the environment path adds a shift specific to the scenario.
On the goal side, success-conditioned rules and outcome-conditioned generators open the goal edge and agent-realized proposers extend the agent edge into $\sigma_\omega$~\cite{dartgui,mobilerl,uigenie,seagent,acurl,scale,skillweaver}, inviting curriculum collapse, which MobileRL risks by design in removing persistently failing tasks~\cite{mobilerl}, and forgetting once the goal distribution moves, which replay mitigates~\cite{digirl,mobilerl}.
On the feedback side, refit verifiers extend the agent edge into FA, inviting compounding reward hacking untested beyond the two or three reported rounds, and training on self-generated trajectories invites model collapse~\cite{uigenie,magicguirms,scale}, whereas frozen self-judges share a fixed bias with the writer and open no such path.
The extension into UC carries erroneous writes, which Mobile-Agent-E reports within a single pass~\cite{mobileagente}, and the environment path carries the drift of live websites between cycles, independently of the agent~\cite{openwebvoyager,opagent}.

\pgraph{Presupposed Resources}\; The mechanisms of GUI and Web navigation presuppose resources that its environments supply.
Repeated attempts at a goal, on which group sampling and reweighting rest, presuppose an environment that resets to a known state~\cite{dartgui,mobilerl}, checkers and screen judges presuppose an end state that a program or a model can inspect~\cite{dartgui,seagent}, and non-parametric reuse presupposes a reader able to exploit what the agent writes~\cite{webcoach}.

\section{Self-Evolution in Embodied Manipulation}
\label{sec:em}

Embodied manipulation occupies the quadrant of \cref{fig:positioning_matrix} in which both ceilings are low on hardware, each rollout costing robot time, a reset, and wear, and each outcome checkable only by a per-task classifier or a learned judge.
Simulation relaxes both, where reinforcement learning against simulator success already forms fixed loops~\cite{simplevlarl,riptvla,pirl} and most self-evolving loops close before any hardware evaluation.
The methods differ chiefly in feedback source: simulator predicates keep fidelity, agent-derived verifiers densify feedback at the cost of an agent-tied bias, and external judges buy fidelity with per-query latency.
Self-reference enters mostly through these verifiers in weight-updating loops and through memory, skills, and code that a frozen model writes and judges in training-free loops.
Reported loops run at most tens of rounds or a bounded practice budget, and hardware evidence rests on eight to thirty evaluation trials per task.

\begin{table*}[!t]
\centering
\scriptsize
\renewcommand{\arraystretch}{1}
\setlength{\tabcolsep}{3.5pt}
\caption{\textbf{Self-evolving methods for embodied manipulation.}
Methods are grouped by whether the environment, the agent, or an external model supplies the feedback that decides each update, and each loop component is tagged by its realizing party: \rl{A} agent, \rl{B} bootstrap, and \rl{E} environment.
Notation: \textsuperscript{f} frozen-snapshot, \textsuperscript{c} co-trained, and \textsuperscript{s} same-model evaluator; M memory, T tool, and I instruction update; Sim simulated, Real physical, and WM world-model loop, with $\rightarrow$ a later hardware evaluation; -- not reported.
}
\label{tab:em_taxonomy}
\vspace{-2mm}
\rowcolors{3}{themelight}{white}
\renewcommand{\tabularxcolumn}[1]{m{#1}}
\begin{tabularx}{\textwidth}{>{\raggedright\arraybackslash}m{2.45cm}>{\centering\arraybackslash}m{0.55cm}>{\raggedright\arraybackslash}X>{\centering\arraybackslash}m{0.45cm}>{\centering\arraybackslash}m{0.65cm}>{\raggedright\arraybackslash}X>{\centering\arraybackslash}m{0.35cm}>{\raggedright\arraybackslash}m{2.2cm}>{\centering\arraybackslash}m{0.8cm}>{\raggedright\arraybackslash}m{2.0cm}}
\toprule
\rowcolor{gray!25}
\rule{0pt}{2.6ex}& \multicolumn{3}{c}{\textbf{Experience Generation (EG)}} & \multicolumn{2}{c}{\textbf{Feedback Acquisition (FA)}} & \multicolumn{3}{c}{\textbf{Update Consolidation (UC)}} & \\
\noalign{\global\aboverulesep=0pt\global\belowrulesep=0pt}
\arrayrulecolor{gray!25}\cmidrule{1-1}\cmidrule{10-10}\arrayrulecolor{black}\cmidrule(lr){2-4}\cmidrule(lr){5-6}\cmidrule(lr){7-9}
\noalign{\global\aboverulesep=0.4ex\global\belowrulesep=0.65ex}
\rowcolor{gray!25}
\multirow{-2}{*}{\textbf{Method}}
& \rule[-1ex]{0pt}{2.8ex}$\boldsymbol{\sigma}$
& \textbf{Goal content} & $\boldsymbol{K_k}$
& $\boldsymbol{V}$
& \textbf{Scoring reference \& signal}
& $\boldsymbol{f}$
& \textbf{Update}
& \textbf{Rounds}
& \multirow{-2}{*}{\textbf{Backbone}} \\
\midrule
\rowcolor{white}\multicolumn{10}{@{}l}{\;\,\ding{60} \textit{Environment feedback}} \\
EmbodiedRSI~\cite{embodiedrsi} {\tiny arXiv'26} & \rl{A} & experiments chosen by value of information (Sim$\rightarrow$Real) & -- & \rl{E} & physical trial outcomes & \rl{A} & code, skills, hierarchical memory (T, M), PPO memory policy & -- & frozen VLA, Qwen3.5-122B \\
RoboBridge~\cite{robobridge} {\tiny arXiv'26} & \rl{B} & LIBERO-PRO training seeds (Sim$\rightarrow$Real) & -- & \rl{E}+\rl{B} & completion predicate, held-out validation gate & \rl{A} & versioned skill documents (M, I) & 2 & $\pi_{0.5}$, Codex, frozen \\
EmbodiedEvo~\cite{embodiedevo} {\tiny arXiv'26} & \rl{B} & repeated attempts with scene resets (Sim) & -- & \rl{E} & completion, errors, termination status & \rl{A} & executable skills, lessons (T, M) & -- & coding agent, frozen \\
RoboHarn-Evo~\cite{roboharnevo} {\tiny arXiv'26} & \rl{B} & interaction rollouts on RMBench (Sim, Real) & -- & \rl{E} & physical feedback on rollouts & \rl{A} & hierarchical physical knowledge (M) & 80 rollouts & GPT-5.5, GPT-6, frozen \\
BATON~\cite{batonem} {\tiny arXiv'26} & \rl{B} & subtasks of RoboMemArena tasks (Sim) & -- & \rl{E} & subtask outcomes, entry-state checks & \rl{A} & verified subtask units (M) & -- & GPT-5.6-sol, frozen VLA \\
Harness VLA~\cite{harnessvla} {\tiny arXiv'26} & \rl{B} & LIBERO-Pro, RoboCasa365, RoboTwin tasks (Sim) & -- & \rl{E} & primitive post-conditions, task success & \rl{A} & global memory of verified sequences (M) & -- & Codex or Claude Code, frozen VLA \\
ASPIRE~\cite{aspire} {\tiny arXiv'26} & \rl{B} & LIBERO-Pro, Robosuite, BEHAVIOR-1K tasks (Sim$\rightarrow$Real) & -- & \rl{E} & execution and collision feedback & \rl{A} & growing skill library (T) & -- & Claude Opus 4.6, frozen \\
ViSkill~\cite{viskill} {\tiny arXiv'26} & \rl{B} & PPO rollouts on ManiSkill PrimitiveSkill (Sim) & -- & \rl{A}+\rl{E} & binary success, skill-guided partial credit & \rl{A} & PPO with skill cards (M) & -- & Qwen2.5-VL-3B \\
Policy-Level RSI~\cite{plrsi} {\tiny arXiv'26} & \rl{A} & force perturbations ranked by a failure model refit each round (Sim) & -- & \rl{E} & simulator failure labels & \rl{B} & PPO with BC anchor, risk-gated specialist & 17 & MoE policy \\
MERL~\cite{merl} {\tiny arXiv'26} & \rl{B} & perturbed LIBERO-PRO tasks, trust-scheduled imagination (Sim, WM) & -- & \rl{E}+\rl{A} & terminal success, reward proxy refit on rollouts\textsuperscript{c} & \rl{B} & RL on real and imagined chunks & -- & OpenVLA-OFT, world model \\
Scale and Selection~\cite{scaleselection} {\tiny arXiv'26} & \rl{B} & fixed robosuite and LIBERO cases, screenshots of a 3D interface (Sim) & -- & \rl{E} & built-in success detectors, champion-challenger acceptance & \rl{A} & harness prompts, tools, control rules (I, T) & 30 & GPT-5.6 Sol, frozen \\
\midrule
\rowcolor{white}\multicolumn{10}{@{}l}{\;\,\ding{60} \textit{Agent-generated feedback}} \\
SC-VLA~\cite{scvla} {\tiny arXiv'26} & \rl{B} & residual RL on ManiSkill3 (Sim, Real) & -- & \rl{A} & predicted progress and displacement\textsuperscript{f} & \rl{B} & SAC residual policy & -- & flow-matching VLA, frozen base \\
Self-Improving EFMs~\cite{siefm} {\tiny NeurIPS'25} & \rl{B} & on-policy rollouts on robot stations (Sim, Real) & -- & \rl{A} & decrease in predicted steps-to-go\textsuperscript{f} & \rl{B} & REINFORCE & -- & PaLI-3B \\
RoboEvolve~\cite{roboevolve} {\tiny arXiv'26} & \rl{A} & tasks parsed from seed images, rising difficulty (WM) & -- & \rl{A} & planner as zero-shot critic\textsuperscript{c} & \rl{B} & GRPO and DPO for planner and simulator & -- & Qwen3-VL-4B, Wan2.2-5B \\
Motus2~\cite{motus2} {\tiny arXiv'26} & \rl{B} & imagined branches from robot-data prefixes (WM$\rightarrow$Real) & -- & \rl{A} & own value model on imagined futures\textsuperscript{f} & \rl{B} & DiffusionNFT on action parameters & -- & Wan2.2-based world model \\
EEAgent~\cite{eeagent} {\tiny arXiv'26} & \rl{B} & repeated trials on VIMA-Bench (Sim) & -- & \rl{A}+\rl{E} & binary success, failure localization\textsuperscript{s} & \rl{A} & long- and short-term reflections (M, I) & -- & GPT-4o, frozen \\
SHAPER~\cite{skillharness} {\tiny arXiv'26} & \rl{B} & VLABench and ESI-Bench episodes (Sim) & -- & \rl{A}+\rl{E} & binary success, round-level critique\textsuperscript{s} & \rl{A} & skill text, context code (I, T) & 4 & Qwen3.6-27B, frozen \\
SEAR~\cite{searrobot} {\tiny arXiv'26} & \rl{A} & self-designed practice (Sim, Real) & -- & \rl{A}+\rl{B} & own local checks, judge on validation seeds & \rl{A} & code skills, written memories (T, M) & -- & coding agents, $\pi_{0.5}$, frozen \\
PragmaBot~\cite{pragmabot} {\tiny arXiv'25} & \rl{B} & tabletop tasks (Real) & -- & \rl{A} & VLM success detector\textsuperscript{s} & \rl{A} & reflections in long-term memory (M) & task stream & GPT-4o, frozen \\
RATs~\cite{rats} {\tiny arXiv'26} & \rl{A} & self-proposed play tasks (Sim$\rightarrow$Real) & -- & \rl{A}+\rl{E} & goal and per-step verifiers\textsuperscript{s}, deterministic checks & \rl{A} & code skill library (T) & 50 & Gemini 3.1 Pro, frozen \\
RoboRSI~\cite{roborsi} {\tiny arXiv'26} & \rl{A} & LIBERO and LIBERO-PRO tasks not yet solved (Sim) & -- & \rl{A}+\rl{E} & reviewer postcondition checks\textsuperscript{s}, simulator success gate & \rl{A} & skill hierarchy, guidance (T, M) & 5 & GPT-5.6-SOL, frozen \\
\midrule
\rowcolor{white}\multicolumn{10}{@{}l}{\;\,\ding{60} \textit{External-model feedback}} \\
Reflection-VLA~\cite{reflectionvla} {\tiny arXiv'25} & \rl{A} & LIBERO-Adapt, simplified at low success (Sim) & -- & \rl{B}+\rl{E} & GPT-4o-composed dense reward, binary success & \rl{B} & PPO and SFT on own successes & -- & OpenVLA-7B, GPT-4o \\
EvoVLA~\cite{evovla} {\tiny ECCV'26} & \rl{B} & 8 parallel environments (Sim$\rightarrow$Real) & -- & \rl{B}+\rl{A} & Gemini stage dictionary scored by CLIP, pose curiosity & \rl{B} & PPO with dual critics & -- & OpenVLA-OFT \\
Find Something~\cite{findcantdo} {\tiny arXiv'26} & \rl{A} & feasible library tasks of low recent success (Real) & -- & \rl{B} & VLM success from pre- and post-images & \rl{B} & residual off-policy RL & -- & $\pi_{0.5}$, frozen base \\
Agentic-VLA~\cite{agenticvla} {\tiny ICML'26} & \rl{B} & LIBERO and RoboTwin tasks, LLM sub-goals (Sim) & 8 & \rl{B}+\rl{A} & VLAC progress, sub-goal weights by success & \rl{B} & GRPO, memory of adapted weights (M) & -- & OpenVLA-OFT \\
Robo-COP~\cite{coorch} {\tiny arXiv'26} & \rl{B} & deployment trials (Sim, Real) & -- & \rl{B} & text and video judges, windowed verification & \rl{A} & LoRA adapters, skills, task memory (M, T) & -- & $\pi_{0.5}$, Gemini 3.8 Flash \\
\bottomrule
\end{tabularx}
\end{table*}

\subsection{EG Bound by Physical Rollouts}\label{sec:em_eg}

In embodied manipulation, experience generation draws goals that pair a language instruction with a scene, and each rollout is an episode of camera observations and robot actions, emitted by a vision-language-action policy or by programs that a language model writes over perception and control primitives.
The task families center on tabletop and household manipulation in LIBERO and its variants, ManiSkill, RoboCasa, RoboTwin, Robosuite, VIMA-Bench, and VLABench~\cite{harnessvla,viskill,aspire,agenticvla,eeagent,skillharness}, extending to memory-dependent tasks in RoboMemArena~\cite{batonem} and physical tasks on robot arms~\cite{siefm,findcantdo,pragmabot,coorch}.
These goals come from a benchmark suite, a library of scene-feasible tasks, or the agent's own proposals, and weight-updating loops run groups of episodes in parallel simulators, training-free loops sequential attempts, and hardware loops one episode at a time, with RoboEvolve imagining episodes inside a learned video simulator~\cite{roboevolve}.
\Cref{tab:em_taxonomy} summarizes the self-evolving methods that satisfy the conditions of \cref{sec:defining_conditions}.

\pgraphhlY{Quantity and Quality Constraints}\; Experience generation is bound by both factors on hardware and chiefly by quality in simulation (\cref{tab:stage1_dims}).
On hardware, quantity binds through physical time, human support, and scene recoverability, each real-world experiment of Self-Improving EFMs running about 20 hours with human station resets~\cite{siefm} and 6.58\% of the 456 autonomous episodes of Find Something You Can't Do ending in a scene that only a human restores~\cite{findcantdo}.
Simulation lifts this bound with parallel environments and group sampling~\cite{simplevlarl,evovla,makoviychuk2021isaac}, and where it also supplies a success predicate, reinforcement learning needs no agent-derived verifier and appears chiefly in fixed loops~\cite{riptvla,crlvla,simplerecipe,wovr,rawdream}.
Quality binds on both sides, a long-horizon task beyond the policy's reach returning $p_g(1-p_g)\approx 0$, and on hardware a persistent workspace leaves only part of a task library feasible after each episode~\cite{findcantdo}, each unreachable or infeasible task costing a full physical rollout.

\pgraphhlP{Task Choice and World Models}\; A larger group is affordable only in simulation, which also supplies the success predicate, and loops pull two other levers: task choice raises $\mathbb{E}[I_{\mathcal{A}_k}]$ at a cost paid in verification, and a learned world model buys rollouts at a cost paid in fidelity.
Task choice stays within a library or its edits, Find Something You Can't Do sampling scene-feasible tasks of low recent success~\cite{findcantdo} and Reflection-VLA simplifying failing tasks through edited simulation definitions~\cite{reflectionvla}.
Self-proposed practice extends this choice~\cite{rats,searrobot,embodiedrsi} and needs a check for each new goal, supplied by simulator predicates, by validation seeds that a sealed judge grades in SEAR~\cite{searrobot}, or by verifiers of the agent's own model in RATs~\cite{rats}.
Controlled returns are reported for both, removing the performance-aware curriculum of Find Something You Can't Do markedly slowing its real-world improvement~\cite{findcantdo}, failure-ranked collection lowering the ManiSkill failure rate of Policy-Level RSI below random collection~\cite{plrsi}, and curious play lifting the RATs executor on LIBERO-PRO to 44.3\% against 36.3\% without play and 38.3\% with random play~\cite{rats}.
A learned world model makes rollouts cheap and recoverable but passes its errors to feedback, as in RoboEvolve, which co-trains its planner with a video simulator from unlabeled seed images~\cite{roboevolve}, and in World-VLA-Loop, which sets long-horizon suites aside once generated video degrades beyond about 300 frames~\cite{worldvlaloop}.

\subsection{FA Bound by Real-Robot Fidelity}\label{sec:em_feedback}

Each cycle hands feedback acquisition an episode of camera frames and actions, together with the simulator state in simulation, and \cref{tab:em_taxonomy} groups the methods by who supplies the feedback.
Scores from the environment (\rl{E}) come from simulator predicates, execution errors, and termination status~\cite{embodiedevo,aspire,harnessvla,batonem,plrsi,scaleselection}.
Scores from the agent (\rl{A}) come from progress heads and value models frozen before reinforcement learning~\cite{siefm,scvla,motus2}, from a co-trained critic or a reward proxy refit on the policy's rollouts~\cite{roboevolve,merl}, and from the frozen model that acts and then judges its own trials~\cite{eeagent,skillharness,searrobot,pragmabot,rats,roborsi}.
Scores from outside the agent (\rl{B}) come from VLM judges and critics~\cite{reflectionvla,evovla,agenticvla,findcantdo,coorch}.
Weight-updating methods turn scores into rewards or advantages for on- or off-policy reinforcement learning~\cite{evovla,agenticvla,siefm,scvla,motus2}, whereas training-free methods turn them into success labels and diagnoses that gate memory, skill, and code writes~\cite{eeagent,embodiedevo,aspire}.

\pgraphhlY{Fidelity and Localization Constraints}\; Feedback acquisition is bound on real robots by fidelity and by a localization whose step-level signals inherit the fidelity of their estimators (\cref{tab:stage2_dims}), the attainable fidelity being high in simulation and low on hardware.
A simulator predicate checks completion exactly, whereas on a real robot a judge must infer it from images with an accuracy reported between 69\% and 100\% across judges and test sets, from 100\% for the best of five VLM judges on 100 image pairs in Find Something You Can't Do~\cite{findcantdo} to 78.9\% for the judge of SOAR on 546 trajectories~\cite{soar} and 69\% to 75\% for that of VideoAgent on 36 videos~\cite{videoagent}.
Cameras show an outcome's appearance and not contact or force, which judges reading images cannot observe.
Under optimization, SEAR reports agents working around its sealed judge, for instance by reading object geometry from asset files~\cite{searrobot}, and policies trained against VLM judges learn to exploit them~\cite{dern2026vlmjudges}.
Localization comes only from learned estimators, such as progress heads, VLAC critics, and per-step verifiers, which return a signal at every step or sub-goal~\cite{siefm,scvla,agenticvla,rats}, whereas a simulator predicate or a real-robot judge returns one verdict per episode.

\pgraphhlP{Dense Feedback from Learned Verifiers}\; A simulator predicate stays fixed as the policy improves, whereas the learned verifiers that make real-robot feedback dense or cheap incur degradations set by whether they follow the policy.
Progress heads and value models frozen before reinforcement learning cut the agent edge~\cite{siefm,scvla,motus2} but meet distribution shift on states that the improving policy reaches outside its imitation data, the key challenge Self-Improving EFMs names for its reward model~\cite{siefm}.
Co-trained critics keep the edge and invite compounding reward hacking, which RoboEvolve counters by binarizing its physical reward terms, citing as stability monotone reward curves that hacking would equally produce~\cite{roboevolve}, and MERL by down-weighting low-trust imagined chunks, outperforming a matched control~\cite{merl}.
A frozen self-judge opens no path into FA but writes its misjudged trials into the memory or skills it later retrieves~\cite{eeagent,skillharness,searrobot,pragmabot,rats}, unless a sealed judge checks them on validation seeds, a judge that the agents of SEAR still circumvent~\cite{searrobot}.
External judges stay independent of the policy but cost 2.6 to 16.9\,s per query in Find Something You Can't Do~\cite{findcantdo}, bounding the rollouts they can score.
Dense feedback on hardware thus rests on fidelity audited within one task distribution, for judges on at most a few hundred samples~\cite{videoagent,findcantdo,soar} and for the progress head on target-task demonstrations~\cite{siefm}.

\subsection{UC Bound by Generalization}\label{sec:em_uc}

Update consolidation in embodied manipulation persists the scored experience in weight updates, training-free writes, or, in four methods, both (\cref{tab:em_taxonomy}).
Weight-updating methods write $\Phi_\theta$ with an update rule fixed before the loop, through reinforcement learning or fine-tuning on their own successes, RoboEvolve updating its video simulator alongside its planner~\cite{siefm,reflectionvla,evovla,scvla,findcantdo,plrsi,motus2,merl,roboevolve}.
Training-free methods keep the backbone frozen and let the agent's language model write reflections~\cite{eeagent,pragmabot}, skills, code, or harnesses~\cite{embodiedevo,searrobot,aspire,rats,roborsi,skillharness,robobridge,scaleselection}, or memory of verified behaviors and physical knowledge~\cite{batonem,harnessvla,roboharnevo}, over two to 50 rounds, a rollout budget, or one pass of a task stream.
ViSkill~\cite{viskill}, Agentic-VLA~\cite{agenticvla}, and Robo-COP~\cite{coorch} train the policy alongside skill cards, a memory of adapted weights, or admitted skills, and EmbodiedRSI keeps its VLA frozen while training the policy that manages its memory of code and skills~\cite{embodiedrsi}.

\pgraphhlY{Retention and Generalization Constraints}\; Update consolidation is bound by generalization (\cref{tab:stage3_dims}), success falling with the distance of the test from practice even where gains persist within one benchmark, whereas retention is measured only in fixed loops, self-evolving ones guarding against regression by gates.
The code policies of SEAR fall from 87.8\% on new initial states of practiced tasks to 33.3\% on new tasks and 28.7\% under a new simulator and action interface, and its trained VLAs from 73.2\% to 43.1\% and 10.2\%~\cite{searrobot}.
Gains of written artifacts persist within one benchmark, SHAPER adding 2.5 points on its in-domain VLABench split and 6.0 to 10.0 on splits with unseen targets or task forms~\cite{skillharness}, although its absolute success on the splits with unseen targets falls from 42.5\% to 26.0\% and 19.5\%.
Hardware transfer is tested on small sets, five tasks of 30 trials each in EmbodiedRSI~\cite{embodiedrsi}, eight scenes per task in RoboHarn-Evo~\cite{roboharnevo}, and 20 tests per task in Robo-COP, which doubles success on only one of three held-out tasks~\cite{coorch}.
Retention is measured only in the fixed loops CRL-VLA and Simple Recipe~\cite{crlvla,simplerecipe} and approximated in self-evolving loops only by gates against regression, the adoption test of Robo-COP and the risk gate of Policy-Level RSI~\cite{coorch,plrsi}.
As in navigation, backbones confound cross-form comparison: training-free loops mostly run frozen frontier models, weight-updating loops open ones.

\pgraphhlP{Tests outside the Writer}\; Feedback acquisition sets the exposure of consolidation, the two forms of writing differing in whether a test outside the writer checks a write before it persists.
Parametric writes pass on the verifier's score, inheriting the distribution shift of frozen progress heads~\cite{siefm,scvla,motus2} and the compounding hacking of co-trained critics~\cite{roboevolve,merl}, except where the adoption test of Robo-COP or the risk gate of Policy-Level RSI adds a test outside the update rule~\cite{coorch,plrsi}.
Non-parametric writes pass tests at different distances from the writer, from held-out seeds and fixed validation sets~\cite{robobridge,skillharness,scaleselection} through execution outcomes~\cite{embodiedevo,aspire,harnessvla,batonem} to the writer's own judgment~\cite{eeagent,searrobot,pragmabot,rats}.
A write that only its writer checks persists with its errors, one SEAR run saving a self-calibrated placement verifier and finishing at 4.6\% against 15.5\% and 23.6\% for two sibling runs without it~\cite{searrobot}.
A test outside the writer matters even when it checks only task success, the harness of Scale and Selection falling from 40\% to 0\% over 20 rounds once every revision is accepted, whereas champion-challenger selection raises held-out success from 51\% to 67\%~\cite{scaleselection}.
The farther the test sits from the writer, the more of the retained gain it protects.

\subsection{Limits and Preconditions}\label{sec:em_positioning}

\pgraph{Bottleneck Attribution}\; In embodied manipulation, both ceilings bind fixed and self-evolving loops alike on hardware, self-reference changing who supplies the goal, the score, or the write but not the cost or attainable fidelity of a physical rollout.
Fixed loops meet the same ceilings, training against simulator success~\cite{simplevlarl,riptvla} or letting a separate model write memory or revise a skill library~\cite{physmem,rpg,hymes,enpire}.
The cost of self-reference lies in what each entry point puts at risk: a success-conditioned choice over a fixed library risks little, only the allocation following the agent~\cite{findcantdo}, whereas self-proposed practice moves verification to checks the agent builds~\cite{searrobot}, a co-trained critic lets its bias follow the policy~\cite{roboevolve,merl}, and a frozen model that judges and writes carries one bias into both roles~\cite{eeagent,pragmabot}.
Task choice and practice offset part of these costs, and only there are controlled gains reported~\cite{findcantdo,rats}.
Beyond these ablations, the costs rest on stage evidence and the provenance in \cref{tab:em_taxonomy}.

\pgraph{Cross-Cycle Degradation}\; Across cycles, self-reference opens paths along which degradation follows the agent, and the environment path adds the sim-to-real gap.
On the goal side, success-conditioned task choice, simplification, and failure-ranked collection open the goal edge~\cite{findcantdo,reflectionvla,plrsi} and self-proposed practice extends the agent edge into $\sigma_\omega$~\cite{rats,searrobot,embodiedrsi,roboevolve}, both inviting curriculum collapse.
On the feedback side, co-trained critics extend the agent edge into FA, inviting compounding reward hacking~\cite{roboevolve,merl}, and training a planner on videos of its co-trained simulator invites model collapse~\cite{roboevolve}, whereas frozen progress heads stay off the agent edge yet admit distribution shift~\cite{siefm}.
The extension into UC carries erroneous writes~\cite{searrobot} and regression when every checkpoint is deployed, below the frozen policy on three of the four tasks whose first candidate its gate rejected~\cite{coorch}, while the environment path carries dynamics that loops closed in simulation never score.
Within reported budgets, regression and erroneous writes are documented, whereas curriculum collapse, compounding reward hacking, model collapse, and distribution shift remain inferred.

\pgraph{Presupposed Resources}\; The mechanisms of embodied manipulation presuppose resources that simulation supplies and hardware withholds.
Reinforcement learning against success presupposes simulator success predicates~\cite{simplevlarl}, held-out gates simulated or deployment trials~\cite{robobridge,coorch}, learned hardware feedback an audited judge~\cite{findcantdo,soar}, and written artifacts a test outside the writer~\cite{scaleselection,searrobot}.

\section{Cross-Scenario Asymmetry}
\label{sec:cross_scenario}

This section characterizes the cross-scenario asymmetry, in which the two directions of a scenario pair receive different kinds of portability verdict, a mechanism that applies in one scenario needing an adaptation, being blocked, or becoming redundant in the other while the reverse direction meets a different combination.
We interpret the asymmetry through two properties of each scenario, which factors are scarce there, exogenously or through resources that self-reference gives up, and which task-structural resources and conditions it holds, chiefly output equivalence, checkable outcomes, and outcome observability.
A mechanism is portable on a target that holds the resource it presupposes, conditional where that resource is held only on a sub-family or after an adaptation, blocked where the outcome it judges or must check is unobservable, and redundant where the factor it raises is not scarce.
The account concerns the kind of verdict each direction receives and leaves aside cost differences across scenarios, which the generation-verification gap would set.
Quadrant placements and cost comparisons are inferred from single-scenario evidence, and the verdicts remain untested.

\subsection{Exogenous and Endogenous Scarcity}\label{sec:cs_regime}

A factor is scarce in a scenario where its value limits what a loop gains, a condition that a binding ceiling implies but that a factor without a ceiling can also meet.
Scarcity is exogenous where it limits fixed loops as well and endogenous where self-reference gives up a resource that a fixed loop keeps, two reasons that comparing the self-evolving and fixed loops of each scenario separates (\cref{sec:mr_positioning,sec:mg_positioning,sec:gw_positioning,sec:em_positioning}).

\pgraph{Exogenous Scarcity}\; Exogenous scarcity has two sources, the saturation of $\mathbb{E}[I_{\mathcal{A}_k}]$ in every scenario and the low ceilings of three.
We infer the saturation from $p_g(1-p_g)$ falling near zero where goals are mastered or out of reach.
Generation saturates from above, the fixed loop Flow-GRPO raising the GenEval score of SD3.5-M from 0.63 to 0.95~\cite{flowgrpo}, and navigation and manipulation from below, with initial navigation success below 20\%~\cite{seagent,uigenie} and physical scenes that afford only part of a task library~\cite{findcantdo}.
Unbounded by any ceiling, $\mathbb{E}[I_{\mathcal{A}_k}]$ counts as scarce in every scenario.
Fixed loops meet the low ceilings as well, scoring open-ended images with learned preference models in generation~\cite{flowgrpo,dancegrpo,spro,onlinevpo,superflow,rrmedit}, paying per-step environment latency in navigation~\cite{computerrl}, and facing resets and success evaluation on hardware~\cite{findcantdo,physmem}.
Self-reference leaves these ceilings in place, changing who supplies the goal, the score, or the write and with it the form of their cost.

\pgraph{Endogenous Scarcity}\; Endogenous scarcity, the residual bottleneck we place in multimodal reasoning, where neither ceiling is low, arises once loops give up the answer keys that fixed loops keep.
Fixed loops filter rollouts by exact match against pre-collected keys, as in M-STAR~\cite{mstar}, whereas loops that pose or label their own questions without a program to compute answers label by consensus or self-judgment at a lower realized fidelity $F_\nu^{(k)}$.
On 500 shared questions, VQS finds 24\% of retained majority-vote labels and 18\% of retained model-judge labels wrong against 5.6\% of retained program-computed answers~\cite{pvse}.
Self-posed goals also give up scripts in SCALE~\cite{scale} and SEAR~\cite{searrobot} without an endogenous scarcity of fidelity, SCALE's web end states remaining open to inspection and SEAR's goals staying in simulation, short of the hardware on which both ceilings bind.

\begin{table*}[!t]
\centering
\scriptsize
\renewcommand{\arraystretch}{1}
\setlength{\tabcolsep}{3pt}
\caption{\textbf{Trade-offs and degradation of self-evolving mechanisms.}
\textit{Evidence} stops short of cross-round measurement for most degradation modes.
Notation: (T) task-structural resource, (M) capability of the agent's model, and artifact authored before the loop if unmarked; \emph{cut} path closed by a frozen evaluator.}
\label{tab:cs_substitutions}
\vspace{-2mm}
\rowcolors{3}{themelight}{white}
\renewcommand{\tabularxcolumn}[1]{m{#1}}
\begin{tabularx}{\textwidth}{%
>{\raggedright\arraybackslash}m{2.25cm}
>{\centering\arraybackslash}m{0.62cm}
>{\raggedright\arraybackslash}m{1.45cm}
>{\raggedright\arraybackslash}m{2.5cm}
>{\raggedright\arraybackslash}m{2.3cm}
>{\raggedright\arraybackslash}m{1.12cm}
>{\raggedright\arraybackslash}m{1.55cm}
>{\raggedright\arraybackslash}X
>{\raggedright\arraybackslash}m{2.2cm}}
\toprule
\rowcolor{gray!25}
\multicolumn{2}{c}{\rule{0pt}{2.8ex}\textbf{Mechanism}}
& \multicolumn{3}{c}{\textbf{Trade-off}}
& \multicolumn{3}{c}{\textbf{Degradation}}
& \\
\noalign{\global\aboverulesep=0pt\global\belowrulesep=0pt}
\arrayrulecolor{black}\cmidrule(lr){1-2}\cmidrule(lr){3-5}\cmidrule(lr){6-8}\arrayrulecolor{gray!25}\cmidrule{9-9}\arrayrulecolor{black}
\noalign{\global\aboverulesep=0.4ex\global\belowrulesep=0.65ex}
\rowcolor{gray!25}
\rule[-1ex]{0pt}{2.8ex}\textbf{Name}
& \textbf{Entry}
& \textbf{Raises}
& \textbf{Requires}
& \textbf{Pays}
& \textbf{Path}
& \textbf{Mode}
& \textbf{Evidence}
& \multirow{-2}{*}{\textbf{Instances}} \\
\midrule
\rowcolor{white}\multicolumn{9}{l}{\;\,\ding{60} \textit{Experience generation}} \\
Success-conditioned allocation & $\sigma_\omega$ & $\mathbb{E}[I_{\mathcal{A}_k}]$ & fixed task set with checkable outcomes (T) & coverage of tasks dropped by design~\cite{mobilerl} & goal & curriculum collapse & untested & \mbox{MR~\cite{syncloop}}, \mbox{MG~\cite{internalgap}}, \mbox{GW~\cite{dartgui,mobilerl}}, \mbox{EM~\cite{findcantdo,plrsi}} \\
Agent-posed goals & $\sigma_\omega$ & $\mathbb{E}[I_{\mathcal{A}_k}]$ & checkable outcomes or output equivalence (T) & pooled keys or scripts, label accuracy 72\% to 61\%~\cite{visplay} & agent & curriculum collapse & untested, countered by diversity terms~\cite{evovid,visplay} & \mbox{MR~\cite{visplay,evolmm}}, \mbox{MG~\cite{asg,rsigen}}, \mbox{GW~\cite{scale,skillweaver}}, \mbox{EM~\cite{rats,searrobot}} \\
Record-conditioned goal generation & $\sigma_\omega$ & $\mathbb{E}[I_{\mathcal{A}_k}]$ & checkable outcomes or output equivalence (T) & pooled keys or scripts & goal & curriculum collapse & untested & \mbox{GW~\cite{seagent,acurl}} \\
Program-checkable goal families & $\sigma_\omega$ & $\mathbb{E}[I_{\mathcal{A}_k}]$ at a program-checked $F_\nu$ & program, simulator, or scripted scene checks (T) & coverage outside the family & agent or goal & forgetting & 72\% to 60\% after a vocabulary switch~\cite{rsiumm} & \mbox{MR~\cite{pvse,spatialevo,visionzero}}, \mbox{MG~\cite{rsiumm}}, \mbox{EM~\cite{reflectionvla}} \\
Imagined rollouts in a world model & $V_\nu$ & $K_k$ beyond $\bar{K}$ via imagined rollouts & learnable short-horizon dynamics (T) & $F_\nu$ via model error, video degrades past 300 frames~\cite{worldvlaloop} & agent or cut & distribution shift, model collapse & inferred, imagined chunks weighted by trust~\cite{merl} & \mbox{EM~\cite{roboevolve,merl,motus2}} \\
\midrule
\rowcolor{white}\multicolumn{9}{l}{\;\,\ding{60} \textit{Feedback acquisition}} \\
Self-consistency vote & $V_\nu$ & $F_\nu$ without keys & output equivalence (T), prior on the true answer (M) & $K_k$ rollouts per goal, label errors~\cite{pvse} & agent & compounding hacking & text-only reasoning~\cite{shafayat2025selftrain} & \mbox{MR~\cite{evolmm,vzero,ttrv}}, \mbox{MG~\cite{asg}} \\
Self-judgment by the agent's model & $V_\nu$ & $F_\nu$ without checkers & generation-\allowbreak verification gap (M) & fidelity moving with the policy & agent & compounding hacking & untraced across rounds~\cite{suder} & \mbox{MR~\cite{isrdpo,rise}}, \mbox{MG~\cite{suder,internalgap}}, \mbox{GW~\cite{mobileagente,agents}}, \mbox{EM~\cite{roboevolve,pragmabot}} \\
Generator-confidence reward & $V_\nu$ & $F_\nu$ without checkers & generator's confidence (M) & fidelity tied to updated weights & agent & compounding hacking & within single runs~\cite{solace,iris} & \mbox{MG~\cite{solace,iris}} \\
Verifier refit on agent trajectories & $V_\nu$ & $L_\nu$, or $F_\nu$ on free-form goals & labeled seed or reference set & fidelity moving with the policy & agent & compounding hacking & audits on fixed labeled sets~\cite{uigenie,magicguirms} & \mbox{MG~\cite{imgownreward,imagersi}}, \mbox{GW~\cite{digirl,uigenie,magicguirms}}, \mbox{EM~\cite{merl}} \\
Frozen self-derived estimator & $V_\nu$ & $L_\nu$, or $F_\nu$ on free-form goals & demonstrations or $\mathcal{A}_0$ calibration & fixed bias off the training support & cut & distribution shift & named as the key challenge~\cite{siefm} & \mbox{MR~\cite{selfjudge,svq}}, \mbox{MG~\cite{prism,mtopsd}}, \mbox{EM~\cite{siefm,scvla,motus2}} \\
\midrule
\rowcolor{white}\multicolumn{9}{l}{\;\,\ding{60} \textit{Update consolidation}} \\
Episode reuse via memory or skills & $f_\zeta$ & $G_\zeta$, $\mathbb{E}[I_{\mathcal{A}_k}]$ on hard goals & writer's generation-\allowbreak verification gap (M) or an online check (T) & exposure to the writer's errors & agent & erroneous writes & wrong verifier saved to memory~\cite{searrobot} & \mbox{MR~\cite{cwi,dynamo}}, \mbox{MG~\cite{manimagent,comfyclaw}}, \mbox{GW~\cite{mobileagente,evoskillgui}}, \mbox{EM~\cite{embodiedevo,searrobot}} \\
\bottomrule
\end{tabularx}
\end{table*}

\subsection{Task-Structural Resources}\label{sec:cs_resources}

Whether a meta-operator that the agent realizes can raise a scarce factor varies across scenarios only with resources that the scenario holds independently of the agent.
We call \emph{task structure} the part of a scenario fixed before any choice of agent or meta-operator, namely the environment family and evaluation distribution of the macro-environment (\cref{def:se_env}) without its goal sampler, together with the interfaces through which a loop resets, replicates, or inspects its environments, a property being task-structural when replacing the agent and its three meta-operators leaves it unchanged.
\emph{Task-structural resources} decide whether a mechanism that presupposes them is defined or affordable, whereas capabilities of the agent's model and artifacts authored before the loop, the two other kinds a mechanism can presuppose, hold on every target.

\pgraph{Resources by Scenario}\; Three task-structural resources recur across the scenarios and are held unequally, output equivalence, checkable outcomes, and resettable rollouts.
Output equivalence, held by short discrete answers but not by images or action trajectories, defines majority voting without converting outputs into answers.
Checkable outcomes give each newly posed or reweighted goal its own reference outside the agent, coming from keys, generation programs, state scripts, or screen judges~\cite{mstar,pvse,dartgui,digirl,r3v,svrr1,selfquestion,verievol}.
Resettable, parallelizable rollouts let group sampling scale, at the price of emulator and virtual-machine instances where episodes run in software environments~\cite{digirl,uitars2}.
Reasoning holds all three, navigation checkable outcomes and bought rollouts, generation free rollouts with checks only on program-checkable sub-families~\cite{flowgrpo,rsiumm}, and manipulation checks and bought rollouts only in simulation~\cite{simplevlarl}, hardware episodes costing physical time~\cite{siefm,findcantdo}.
The verdicts turn chiefly on output equivalence and checkable outcomes, rollouts entering them only through $\bar{K}$, which decides whether world models are redundant.
The ceilings of \cref{fig:positioning_matrix} reflect these resources only in part, navigation keeping a low $\bar{K}$ under per-step latency despite resettable rollouts and sharing a high $\bar{F}$ with reasoning, which alone holds output equivalence.
Portability is therefore derived from the resources of each scenario, not read off its quadrant.

\pgraph{Observability and Online Checks}\; Two further conditions decide whether a held resource serves the loop, the observability of the outcome, which task structure fixes, and an online check built on the resource, which the loop's design supplies or omits.
The outcome is observable where it appears in the agent's observations or in a check available to the loop, a condition that every mechanism judging or checking it presupposes.
An \emph{online check}, a check built on a resource that runs inside the loop, is the only route by which the resource lowers the realized cost of a trade-off, and loops run one wherever answer keys, execution results, or simulator outcomes are available, training-free methods checking their writes against them~\cite{cwi,skillweaver,embodiedevo}.
Holding a resource does not ensure such a check, task streams, open prompts, and self-chosen practice tasks leaving none even in navigation, as in WebCoach's task stream~\cite{webcoach}.
The frozen acting model then judges its own episodes and writes the update, as in ManimAgent~\cite{manimagent}, Mobile-Agent-E~\cite{mobileagente}, and SEAR~\cite{searrobot}, and an error of judgment enters the update unchecked, one simulated SEAR run saving a self-built verifier wrong on 91\% of later rollouts.
A loop without an online check thus gives up a resource that its scenario holds.

\subsection{Trade-offs of Self-Evolving Mechanisms}\label{sec:cs_summary}

\cref{tab:cs_substitutions} recasts the mechanisms of the scenario chapters as trade-offs, in which an agent-realized meta-operator raises a factor by supplying what the agent records, proposes, judges, or writes in place of a component fixed before the loop.
The goal sampler supplies the agent's success record or its own proposals in place of a fixed goal distribution, the feedback function the agent's judgment in place of an answer key or a checker, and the evolution operator the agent's authorship in place of an update rule written in advance.
Each trade-off pays what the supplied output lacks relative to the replaced component and presupposes a resource that is task-structural, a capability of the agent's model, or an artifact authored before the loop.
Of its three elements, the factor raised, the resource presupposed, and the price paid, only the first two enter the portability verdicts.

\pgraph{Experience Generation}\; The mechanisms of experience generation raise $\mathbb{E}[I_{\mathcal{A}_k}]$, scarce in every scenario, and differ in what they give up to raise it, with a learned world model as the one exception.
Success-conditioned allocation reweights a fixed task set, giving up coverage of the tasks it down-weights and presupposing checkable outcomes on that set.
Created goals, posed by the agent or generated from its record, extend the factor beyond a fixed pool but give up the answer key or checking script of each pooled goal, presupposing in its place an outcome that something other than their generator can check or answers that a vote can compare.
Program-checkable goal families keep the check by creating goals only where a program answers or renders them, as in VQS~\cite{pvse} and RSI in UMMs~\cite{rsiumm}, giving up coverage of $\mathcal{G}$ outside the family.
A learned world model, the one mechanism that replaces a component of the environment and not a fixed component of a meta-operator, instead raises $K_k$, replacing the environment dynamics $P$ with imagined rollouts that $\bar{K}$ does not bound and presupposing dynamics learnable over short horizons, the trade-off that model-based reinforcement learning has analyzed from Dyna~\cite{sutton1991dyna} to MBPO~\cite{janner2019mbpo}.
It enters the loop through $V_\nu$, its imagined outcomes scored by a reward or value head~\cite{merl,motus2} or by the agent's own judge~\cite{roboevolve}, and model error thus reaches $F_\nu$.

\pgraph{Feedback Acquisition}\; Each feedback mechanism pays for the substituted judgment in fidelity, its gain depending on the evaluator it replaces, and the mechanisms fall into three groups by the kind of resource they presuppose, the distinction on which their verdicts turn.
The self-consistency vote raises $F_\nu$ without keys at a price of several rollouts per goal and presupposes output equivalence together with a prior concentrated on the true answer.
Self-judgment and generator-confidence rewards raise $F_\nu$ without checkers and presuppose only capabilities of the agent's model, a generation-verification gap or the generator's confidence, which hold on every target, the gap pricing these rewards without deciding where they port.
Verifiers refit on the agent's trajectories and frozen self-derived estimators raise $L_\nu$ besides $F_\nu$ and presuppose an artifact authored before the loop, a labeled seed, a reference set, or demonstrations.

\pgraph{Update Consolidation}\; Update consolidation contributes one trade-off, episode reuse through memory, skills, or harnesses, which raises $\mathbb{E}[I_{\mathcal{A}_k}]$ on goals initially out of reach and, where a written library carries beyond $\mathcal{G}_k$, also $G_\zeta$, paying in exposure to the writer's errors.
It presupposes either the writer's generation-verification gap, a capability held on every target, or an online check that gates the writes, as keys and validation sets do in training-free reasoning loops~\cite{cwi,dynamo,m2note}.
Weight updates under a rule fixed before the loop supply no agent authorship and enter the trade-offs only through the feedback they consume.

\subsection{Costs and Alternative Sources}\label{sec:cs_costs}

The evidence on the price paid, the trade-off element the verdicts leave out, serves the account in two ways, setting the cost of the conditional routes the verdicts open and bearing on whether a factor other than scarcity and resources could produce the asymmetry.
The comparisons below each stay inside one paper, bearing on a mechanism's gain and, where measured, its cost.
Degradation, the rest of the cost, is left aside, its evidence too uneven for cross-scenario comparison.

\pgraph{Generation-Verification Gap}\; The generation-verification gap prices a trade-off wherever the agent's own judgment serves as the reference, as a capability that holds on every target (\cref{sec:cs_summary}), and remains inferred in multimodal loops.
Introduced for language models, the gap is the improvement a model obtains by filtering or reweighting its own generations with its own verification~\cite{song2025mindgap}.
We extend the term by separating a task-level component, the difference in difficulty between verifying and generating on a task family, from a model-level component, the difference between the verification and generation accuracy of the agent's own model on the same goals.
The model-level component remains unestimated, multimodal audits reporting only the verification side~\cite{pvse,manimagent,evoskillgui}, and backbone priors would confound it even once measured (\cref{sec:positioning_matrix}).
We infer the task-level component to be large in GUI and Web navigation, where a final screen can be inspected without replaying the steps that reach it, and small in open-prompt generation, where judging whether an image satisfies a prompt calls on the compositional understanding that producing it requires, a difference that makes the cost of self-judgment vary across scenarios even where its verdict does not.

\pgraph{Evaluator Fidelity}\; The feedback ablations order evaluators by fidelity and thereby price the conditional routes that substitute a judge for a missing check.
Two ablations, both in generation, compare a verifier built on the agent's own judgment with the evaluator it replaces, a frozen self-rater against a program check in one and a co-trained reward model against a learned preference scorer in the other.
In RSI in UMMs, training on the better of two images as rated by the frozen starting model gains 0.5 points, against 8.6 points with execution-verified charts for the same requests and steps~\cite{rsiumm}.
The reward model that The Image as Its Own Reward~\cite{imgownreward} co-trains against reference images matches the PickScore of Flow-GRPO~\cite{flowgrpo} trained on PickScore and wins 70.0\% of the authors' human image-quality comparisons against it, a model that the same raters place below its base.
A verifier built on the agent's own judgment thus falls below a program check and can exceed a learned preference scorer, an ordering whose two ends come from different verifiers, one frozen and one co-trained, and which the other scenarios leave untested.
The ordering would be contradicted if such a verifier outperformed a keyed or programmatic check at a matched budget on fixed goals.

\pgraph{Goal-Mechanism Gains}\; Goal mechanisms gain in all four scenarios with one reported exception, consistent with a scarcity of $\mathbb{E}[I_{\mathcal{A}_k}]$ that gives this factor no part in the asymmetry.
Removing allocation hurts both scenarios that ablate it, MobileRL~\cite{mobilerl} losing 6.3 points of AndroidWorld success without its failure curriculum filter and Policy-Level RSI~\cite{plrsi} raising its ManiSkill failure rate from 2.34\% to 3.36\% under random scene selection.
Created goals lead untrained, random, or frozen-source generators in VisPlay~\cite{visplay}, RATs~\cite{rats}, ACuRL~\cite{acurl}, and RSIgen~\cite{rsigen}, and trail the NNetNav curriculum by 1.3 points in one SEAgent setting~\cite{seagent}.
Within a program-checkable family the error-driven chart distribution of RSI in UMMs leads a fixed distribution with the same number of verified charts by 2.8 to 4.1 points in every round~\cite{rsiumm}.
These comparisons come from the authors of each method and may overstate the gain, yet their agreement across the four scenarios leaves the verdicts on goal mechanisms to turn on the resources they presuppose.

\pgraph{Writer Dependence}\; The gain of reuse depends on the writer, a model-level dependence that travels with the agent to every target and adds no asymmetry of its own.
Authored writes outperform skill-free or raw alternatives in two navigation ablations, EvoSkill-GUI~\cite{evoskillgui} gaining 6.7 points on MobileWorld over pass@3 without skills and Agent S~\cite{agents} 7.7 points on an OSWorld subset over full trajectories in place of its self-evaluator's summaries.
SEAR~\cite{searrobot}, holding the scenario fixed and varying only the writer, finds the sign of this gain reversed, GPT-5.6-sol on RoboLab dropping from 46.9\% without a library to 42.4\% with its own, beyond the spread of three runs, and rising to 60.6\% with one that Claude Opus 5 wrote, a loss the authors trace to object-location recipes with few grasp checks.
Whether such content tracks the generation-verification gap of the writer's model is untested, and a library that gained the same whoever wrote it would contradict the dependence on the writer.

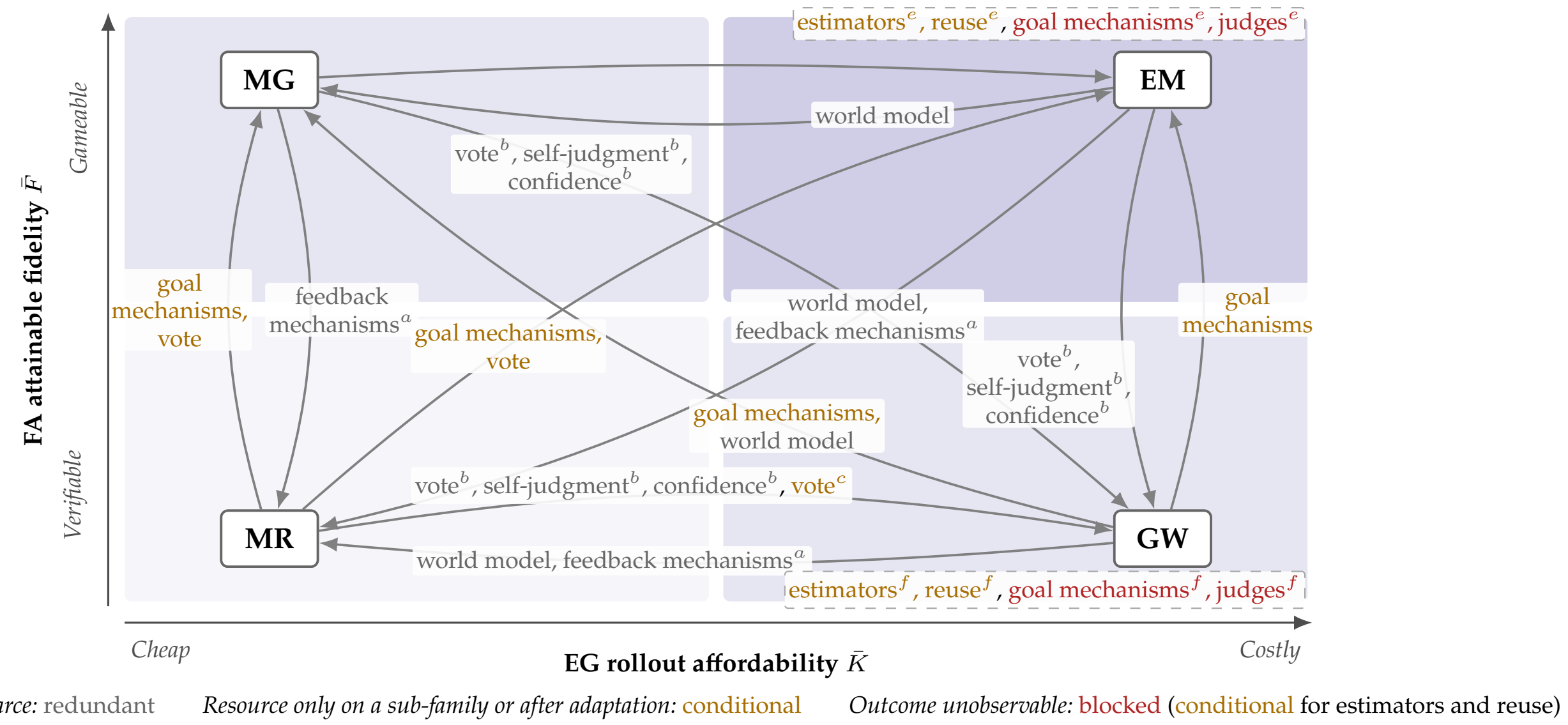
\begin{figure*}[!t]
\centering
\def\fR#1{\textcolor{black!60}{#1}}
\def\fC#1{\textcolor{vcondcolor!85!black}{#1}}
\def\fB#1{\textcolor{vblockcolor}{#1}}
\begin{tikzpicture}[
  font=\fontsize{7}{8.2}\selectfont, >={Latex[length=1.9mm,width=1.5mm]},
  scen/.style={draw=black!60, line width=0.7pt, rounded corners=2pt, fill=white,
               minimum width=1.0cm, minimum height=0.58cm, font=\fontsize{8.5}{9}\selectfont\bfseries},
  arr/.style={->, line width=0.7pt, draw=black!50},
  lab/.style={align=center, inner sep=1.2pt, fill=white, fill opacity=0.9, text opacity=1, rounded corners=1pt},
  axlab/.style={font=\fontsize{7.5}{8.5}\selectfont\bfseries},
  pole/.style={font=\fontsize{7}{8}\selectfont, text=black!70}
]
\definecolor{posaccent}{RGB}{99,91,177}

\fill[posaccent,opacity=0.06,rounded corners=2pt] (0.05,0.10) rectangle (6.10,3.05);
\fill[posaccent,opacity=0.16,rounded corners=2pt] (6.25,0.10) rectangle (12.30,3.05);
\fill[posaccent,opacity=0.16,rounded corners=2pt] (0.05,3.20) rectangle (6.10,6.15);
\fill[posaccent,opacity=0.30,rounded corners=2pt] (6.25,3.20) rectangle (12.30,6.15);

\node[scen] (MR) at (1.55,0.75) {MR};
\node[scen] (GW) at (10.8,0.75) {GW};
\node[scen] (MG) at (1.55,5.50) {MG};
\node[scen] (EM) at (10.8,5.50) {EM};

\draw[arr] (MR) to[bend left=16] (MG);
\draw[arr] (MG) to[bend left=16] (MR);
\draw[arr] (MR) to[bend left=9] (GW);
\draw[arr] (GW) to[bend left=5] (MR);
\draw[arr] (MG) to[bend left=3] (EM);
\draw[arr] (EM) to[bend left=9] (MG);
\draw[arr] (GW) to[bend right=16] (EM);
\draw[arr] (EM) to[bend right=16] (GW);
\draw[arr] (MR) to[bend left=14] coordinate[pos=0.27] (lMREM) (EM);
\draw[arr] (EM) to[bend left=14] coordinate[pos=0.36] (lEMMR) (MR);
\draw[arr] (MG) to[bend left=14] coordinate[pos=0.27] (lMGGW) (GW);
\draw[arr] (GW) to[bend left=14] coordinate[pos=0.36] (lGWMG) (MG);

\node[lab] at (0.62,3.12) {\fC{goal}\\\fC{mechanisms,}\\\fC{vote}};
\node[lab] at (2.3,3.12) {\fR{feedback}\\\fR{mechanisms$^{a}$}};
\node[lab] at (5.3,1.30) {\fR{vote$^{b}$, self-judgment$^{b}$, confidence$^{b}$}, \fC{vote$^{c}$}};
\node[lab] at (5.1,0.52) {\fR{world model, feedback mechanisms$^{a}$}};
\node[lab] at (7.9,5.14) {\fR{world model}};
\node[lab] at (11.68,3.12) {\fC{goal}\\\fC{mechanisms}};
\node[lab] at (9.62,2.35) {\fR{vote$^{b}$,}\\\fR{self-judgment$^{b}$,}\\\fR{confidence$^{b}$}};
\node[lab] at (lMREM) {\fC{goal mechanisms,}\\\fC{vote}};
\node[lab] at (lEMMR) {\fR{world model,}\\\fR{feedback mechanisms$^{a}$}};
\node[lab] at (lMGGW) {\fR{vote$^{b}$, self-judgment$^{b}$,}\\\fR{confidence$^{b}$}};
\node[lab] at (lGWMG) {\fC{goal mechanisms,}\\\fR{world model}};
\node[lab, draw=black!35, dashed, anchor=east] at (12.28,6.08) {\fC{estimators$^{e}$, reuse$^{e}$}, \fB{goal mechanisms$^{e}$, judges$^{e}$}};
\node[lab, draw=black!35, dashed, anchor=east] at (12.28,0.22) {\fC{estimators$^{f}$, reuse$^{f}$}, \fB{goal mechanisms$^{f}$, judges$^{f}$}};

\node[scen] at (MR) {MR}; \node[scen] at (GW) {GW};
\node[scen] at (MG) {MG}; \node[scen] at (EM) {EM};

\draw[->,line width=0.6pt,black!70] (0.05,-0.12) -- (12.35,-0.12);
\draw[->,line width=0.6pt,black!70] (-0.12,0.05) -- (-0.12,6.20);
\node[axlab,anchor=north] at (6.18,-0.30) {EG rollout affordability $\bar{K}$};
\node[pole,anchor=north west] at (0.00,-0.18) {\emph{Cheap}};
\node[pole,anchor=north east] at (12.35,-0.18) {\emph{Costly}};
\node[axlab,rotate=90,anchor=south] at (-0.62,3.12) {FA attainable fidelity $\bar{F}$};
\node[pole,rotate=90,anchor=south] at (-0.22,1.20) {\emph{Verifiable}};
\node[pole,rotate=90,anchor=south] at (-0.22,5.00) {\emph{Gameable}};

\node[anchor=center, inner sep=0pt] at (6.18,-1.0) {\emph{Factor not scarce:} \fR{redundant}\qquad\emph{Resource only on a sub-family or after adaptation:} \fC{conditional}\qquad\emph{Outcome unobservable:} \fB{blocked} (\fC{conditional} for estimators and reuse)};
\end{tikzpicture}
\vspace{-1mm}
\caption{\textbf{Directed portability on the positioning matrix.}
Each arrow lists the mechanisms portable or conditional in its source but redundant or lower on its target, and dashed boxes list the verdicts on GW and EM goals with unobservable outcomes.
Goal mechanisms: allocation, agent-posed and record-conditioned goals, and program-checkable families, the last inapplicable to unobservable outcomes.
Judges: vote and self-judgment. Estimators: refit verifier and frozen estimator. Feedback mechanisms: judges, estimators, and generator-confidence reward.
Superscripts follow \cref{tab:cs_pairs}.
}
\label{fig:cs_portability}
\end{figure*}

\begin{table}[!t]
\centering
\footnotesize
\caption{\textbf{Portability of self-evolving mechanisms (list in \cref{tab:cs_substitutions}) across target scenarios.}
\vPort\ portable, \vCond\ conditional, \vBlock\ blocked, and \vRed\ redundant, with \vCond\ also marking a verdict capped by an unobservable outcome; $^{\ast}$ reported instance.
A superscript limits an entry to the sub-family noted below and an unmarked entry covers the rest of its cell, EM entries referring to physical hardware unless marked $^{d}$.
Verdicts are untested preconditions, not transfer gains.
}
\label{tab:cs_pairs}
\vspace{-2mm}
\renewcommand{\arraystretch}{1}
\setlength{\tabcolsep}{1.2pt}
\begin{tabular}{@{}l*{2}{>{\centering\arraybackslash}p{0.84cm}}*{2}{>{\centering\arraybackslash}p{1.2cm}}@{}}
\toprule
\rowcolor{gray!25}
\textbf{Mechanism} & \textbf{MR} & \textbf{MG} & \textbf{GW} & \textbf{EM} \\
\midrule
\multicolumn{5}{@{}l}{\textit{Presupposes a task-structural resource}} \\
Success-conditioned allocation      & \vPort$^{\ast}$ & \vCond$^{\ast}$ & \vPort$^{\ast}$\,\vBlock$^{f}$ & \vCond$^{\ast}$\,\vBlock$^{e}$\,\vPort$^{\ast d}$ \\
Agent-posed goals                   & \vPort$^{\ast}$ & \vCond$^{\ast}$ & \vPort$^{\ast}$\,\vBlock$^{f}$ & \vCond\,\vBlock$^{e}$\,\vPort$^{\ast d}$ \\
Record-conditioned goal generation  & \vPort & \vCond & \vPort$^{\ast}$\,\vBlock$^{f}$ & \vCond\,\vBlock$^{e}$ \\
Program-checkable goal families     & \vPort$^{\ast}$ & \vCond$^{\ast}$ & \vPort & \vCond\,\vPort$^{\ast d}$ \\
Imagined rollouts in a world model  & \vRed & \vRed & \vCond & \vCond$^{\ast}$ \\
Self-consistency vote               & \vRed$^{a}$\,\vPort$^{\ast}$ & \vCond$^{\ast}$ & \vRed$^{b}$\,\vCond$^{c}$\,\vBlock$^{f}$ & \vCond\,\vBlock$^{e}$ \\
\midrule
\multicolumn{5}{@{}l}{\textit{Presupposes only the agent's model or a pre-loop artifact}} \\
Self-judgment by the agent's model  & \vRed$^{a}$\,\vPort$^{\ast}$ & \vPort$^{\ast}$ & \vRed$^{b}$\,\vPort$^{\ast c}$\,\vBlock$^{f}$ & \vPort$^{\ast}$\,\vBlock$^{e}$ \\
Generator-confidence reward         & \vRed$^{a}$\,\vPort & \vPort$^{\ast}$ & \vRed$^{b}$\,\vPort & \vPort \\
Verifier refit on agent trajectories& \vRed$^{a}$\,\vPort & \vPort$^{\ast}$ & \vPort$^{\ast}$\,\vCond$^{f}$ & \vPort\,\vCond$^{e}$\,\vPort$^{\ast d}$ \\
Frozen self-derived estimator       & \vRed$^{a}$\,\vPort$^{\ast}$ & \vPort$^{\ast}$ & \vPort\,\vCond$^{f}$ & \vPort$^{\ast}$\,\vCond$^{e}$ \\
Episode reuse via memory or skills  & \vPort$^{\ast}$ & \vPort$^{\ast}$ & \vPort$^{\ast}$\,\vCond$^{f}$ & \vPort$^{\ast}$\,\vCond$^{e}$ \\
\bottomrule
\end{tabular}
\par\smallskip
\begin{minipage}{\columnwidth}\scriptsize\raggedright
$^{a}$ key or program-computed answer, $^{b}$ outcome checked online, $^{c}$ screen-visible outcome without an online check, $^{d}$ simulation, $^{e}$ outcome neither in the agent's observations nor checked by instrumentation, with contact-rich assembly as a candidate, $^{f}$ background-state outcome that no reachable screen shows and no script checks.
\end{minipage}
\end{table}

\subsection{Directed Portability across Scenario Pairs}\label{sec:cs_transfer}

The two directions of each scenario pair receive verdicts of different kinds, a difference traced to which factors are scarce and which task-structural resources and conditions hold.
The direction from $X$ to $Y$ collects the mechanisms portable or conditional in $X$ that on $Y$ are redundant or rank lower in the order portable, conditional, blocked, and reversing the pair exchanges both the mechanisms carried and the conditions they meet. A mechanism whose verdict does not fall belongs to neither direction by construction, and the asymmetry accordingly concerns the kind of obstacle a direction meets, not how many mechanisms transfer.
\cref{tab:cs_pairs} assigns per-target verdicts by the rules below.

\pgraph{Verdict Rules}\; The rules apply in a fixed order, observability first, a mechanism that judges or checks an unobservable outcome being blocked whatever its other conditions, then scarcity, which decides redundancy, and resources last, which decide between portable and conditional.
A mechanism is \emph{redundant} where the factor it raises is not scarce, judged at the ceiling or, where the loop gives up a resource, at the realized value, as for self-labeled questions without keys and task streams without online checks.
$\mathbb{E}[I_{\mathcal{A}_k}]$ and $G_\zeta$, neither bounded by a ceiling, count as scarce everywhere (\cref{sec:cs_regime}), and $L_\nu$ as not scarce where a single action fixes $L_\nu\!=\!1$.
Otherwise a mechanism is \emph{portable} where the target holds the resource it presupposes and \emph{conditional} where an adaptation, such as converting outputs into answers or adding an external judge, or a sub-family supplies it, a missing resource admitting such a route in the surveyed targets wherever the outcome is observable.
Capabilities of the agent's model and artifacts authored before the loop count as held everywhere (\cref{sec:cs_resources}).
Where the outcome is unobservable, observability caps the verdict and every mechanism that judges or checks it is \emph{blocked}, even those presupposing only the agent's model, whereas low evaluator accuracy only raises the cost of a conditional route.
A portable verdict states that a precondition holds, not that a transfer improves the target loop.

\pgraph{Redundant Verdicts}\; Redundant verdicts fall on world models where rollouts are cheap and on feedback mechanisms where an external reference already supplies every factor they raise.
Reasoning and generation keep $\bar{K}$ high through a single decoding or sampling pass, leaving world models redundant there.
The vote, self-judgment, and generator-confidence rewards are redundant on reasoning questions with keys or program-computed answers and on navigation goals whose outcome a check inspects online.
Self-judging navigation loops such as Agent S~\cite{agents} and Mobile-Agent-E~\cite{mobileagente}, by contrast, run on task streams or practice goals whose screen-visible outcomes no check inspects, leaving $F_\nu$ scarce at the realized value and self-judgment portable.
Refit verifiers and frozen estimators are redundant only on these reasoning questions, where one action fixes $L_\nu\!=\!1$, multi-step navigation and manipulation keeping $L_\nu$ scarce.

\pgraph{Conditional Verdicts}\; Conditional verdicts arise where the target holds a presupposed task-structural resource only on a sub-family or through an adaptation.
Allocation and program-checkable families are conditional where checks cover only a sub-family, as in generation~\cite{internalgap,rsiumm} and on hardware, where an external judge grades a task library~\cite{findcantdo} or a scripted scene check covers a fixed family~\cite{kalashnikov2018qtopt}.
Created goals, checkable in reasoning and navigation, find a check in generation only through image questions, rendered charts, or external judges~\cite{asg,rsiumm,rsigen}.
On hardware, agent-posed goals stay in simulation~\cite{rats,searrobot}, and the external judge of before-and-after images under which SEAR practices given tasks offers a route where success shows as object motion, a case FailBench finds nearly solved~\cite{failbench}.
World models find learnable dynamics on hardware only over short horizons~\cite{worldvlaloop} and in navigation only over text-rendered web states, as in DynaWeb~\cite{dynaweb}.
Outside the discrete answers of reasoning, the vote finds output equivalence only through questions that turn outputs into comparable answers, as ASG poses about generated images~\cite{asg}, a conversion untested on final screens and camera frames.

\pgraph{Blocked Verdicts}\; Blocked verdicts arise where the outcome of a goal appears neither in the agent's observations nor in any check available to the loop, as for background-state goals in navigation that no reachable screen shows and no script inspects.
Contact-rich assembly is the candidate sub-family on hardware, where no judge in FailBench exceeds 0.60 balanced accuracy against 0.77 for the best judge overall, a result consistent with frames that omit the contact on which success depends, although the authors also name task-specific knowledge and precision~\cite{failbench}.
On such goals the vote and self-judgment have nothing to judge, recorded actions showing what the agent did and not whether the state changed, created goals have nothing to check them, and success-conditioned allocation has no success record to read, a block reaching model-only mechanisms as well as resource-dependent ones.
Refit verifiers and frozen estimators keep the progress that visible steps show, and episode reuse the writes that need no judgment of success, both becoming conditional there, whereas generator-confidence rewards, which never judge the outcome, stay portable at their cost in fidelity, a verdict certifying only that their precondition holds while their fidelity on outcomes that nothing observes stays unknown.
No reported instance falls in a blocked cell, leaving the rules consistent yet unsupported.

\pgraph{Asymmetry by Target}\; The kind of verdict that a direction meets is fixed chiefly by its target, and the two directions of a pair differ wherever their targets differ in scarcity, resources, or observability.
Checkable outcomes act on both sides of the pair of navigation and generation, the judges of generation becoming redundant on navigation goals that a check inspects online and the created goals of navigation becoming conditional on the open prompts of generation.
Reasoning and manipulation show the pattern most sharply, the vote and the goal mechanisms carried from reasoning becoming conditional on hardware and, program-checkable families aside, blocked where contact decides success, while the world models and estimators of manipulation become redundant on reasoning questions with keys or program-computed answers (\cref{tab:cs_pairs}).
More broadly, directions into reasoning meet only redundancy, directions into generation and hardware chiefly conditional goal mechanisms and votes, and where the outcome is unobservable, navigation and hardware block goal mechanisms and judges from every source, a verdict that depends on the target alone (\cref{fig:cs_portability}).

\pgraph{Structural Source}\; Taken together, we interpret the structural source of the asymmetry as two scenario properties that the directions of a pair meet in different combinations, the scarcity that $\bar{K}$ and $\bar{F}$ fix, read at the realized value where a loop gives up a resource, and the unequal task structure, comprising its resources and the observability of its outcomes.
Scarcity decides redundancy, task-structural resources decide between portable and conditional verdicts, and observability alone decides blocking.
Observability is also the one task-structural condition that bounds mechanisms presupposing only the agent's model.
Matrix positions anticipate most verdicts, failing where output equivalence or observability decides them, as for the vote and background-state goals in navigation.
Still untested, the verdicts would fall to a gain over an equal-budget control for an unadapted mechanism lacking its resource, one raising a non-scarce factor, or one judging unobservable outcomes. 
\section{Self-Evolving Multimodal Agent Evaluation}
\label{sec:evaluation}

\begin{table*}[!t]
\centering
\scriptsize
\renewcommand{\arraystretch}{1}
\providecommand{\noRole}{\textcolor{gray!45}{--}}
\setlength{\tabcolsep}{2.8pt}
\caption{\textbf{Task-specific benchmarks used with self-evolving multimodal agents.}
Pool, Env, and Checker mark benchmarks supplying a method's goals, hosting its own goals, or driving its updates.
$^{\diamond}$ discrete answers that admit voting without a key.
}
\label{tab:eval_tasks}
\vspace{-2mm}
\rowcolors{3}{themelight}{white}
\renewcommand{\tabularxcolumn}[1]{m{#1}}
\begin{tabularx}{\textwidth}{%
>{\raggedright\arraybackslash}m{3.1cm}
>{\raggedright\arraybackslash}X
>{\raggedright\arraybackslash}m{2.6cm}
>{\raggedright\arraybackslash}m{3.3cm}
>{\centering\arraybackslash}m{0.8cm}
>{\centering\arraybackslash}m{0.8cm}
>{\centering\arraybackslash}m{1.0cm}}
\toprule
\rowcolor{gray!25}
\textbf{Benchmark} & \textbf{Goals} & \textbf{Reset} & \textbf{Scorer} & \textbf{Pool} & \textbf{Env} & \textbf{Checker} \\
\midrule
\rowcolor{white}\multicolumn{7}{l}{\;\,\ding{60} \textit{Multimodal reasoning}} \\
MathVista~\cite{lu2024mathvista} & 6,141 math questions & none needed & answer key$^{\diamond}$ & \cYes & \noRole & \cYes \\
MathVerse~\cite{zhang2024mathverse} & 2,612 math problems, 6 versions & none needed & answer key$^{\diamond}$ & \cYes & \noRole & \noRole \\
MathVision~\cite{mathvision} & 3,040 math problems, 16 disciplines & none needed & answer key$^{\diamond}$ & \cYes & \noRole & \noRole \\
MMMU~\cite{mmmu} & 11.5K college questions, 30 subjects & none needed & answer key$^{\diamond}$ & \cYes & \noRole & \cYes \\
ChartQA~\cite{masry2022chartqa} & 32.7K chart questions & none needed & answer key$^{\diamond}$ & \cYes & \noRole & \cYes \\
Geometry3K~\cite{intergps} & 3,002 geometry problems & none needed & answer key$^{\diamond}$ & \cYes & \noRole & \cYes \\
PhysBench~\cite{physbench} & 10,002 physics image and video entries & none needed & answer key$^{\diamond}$ & \cYes & \noRole & \cYes \\
VSI-Bench~\cite{vsibench} & 5,000+ spatial video questions & none needed & answer key$^{\diamond}$ & \noRole & \noRole & \noRole \\
HallusionBench~\cite{hallusionbench} & 1,129 illusion questions & none needed & answer key & \noRole & \noRole & \noRole \\
MMSI-Bench~\cite{mmsibench} & 1,000 multi-image spatial questions & none needed & answer key$^{\diamond}$ & \cYes & \noRole & \cYes \\
MMVU~\cite{mmvu} & 3,000 expert video questions, 27 subjects & none needed & answer key & \noRole & \noRole & \noRole \\
\midrule
\rowcolor{white}\multicolumn{7}{l}{\;\,\ding{60} \textit{Multimodal generation}} \\
GenEval~\cite{geneval} & 553 object-composition prompts & none needed & object detector & \cYes & \noRole & \noRole \\
T2I-CompBench++~\cite{t2icompbench} & 8,000 compositional prompts & none needed & VQA, detector, MLLM & \cYes & \noRole & \noRole \\
DPG-Bench~\cite{ella} & 1K dense prompts & none needed & VQA model & \cYes & \noRole & \noRole \\
WISE~\cite{wise} & 1,000 knowledge prompts & none needed & VLM judge & \cYes & \noRole & \noRole \\
Text rendering~\cite{flowgrpo} & 1,018 test prompts & none needed & OCR accuracy & \cYes & \noRole & \cYes \\
VBench~\cite{vbench} & video prompts, 16 dimensions & none needed & per-dimension metrics & \noRole & \noRole & \cYes \\
Diagram and design sets & per-paper charts, diagrams, edits & none needed & per-paper judge or program & \cYes & \noRole & \noRole \\
\midrule
\rowcolor{white}\multicolumn{7}{l}{\;\,\ding{60} \textit{GUI and web navigation}} \\
OSWorld~\cite{osworld} & 369 desktop tasks & virtual machine & per-task script & \cYes & \cYes & \cYes \\
ScienceBoard~\cite{scienceboard} & 169 scientific-software tasks & OS environment & per-task check & \noRole & \noRole & \noRole \\
AndroidWorld~\cite{androidworld} & 116 task templates, 20 apps & emulator & per-task script & \cYes & \noRole & \cYes \\
AndroidLab~\cite{androidlab} & 138 tasks, 9 apps & emulator & per-task check & \cYes & \cYes & \noRole \\
AitW~\cite{aitw} & 30K instructions & none (offline) & none shipped & \cYes & \noRole & \noRole \\
AndroidControl~\cite{androidcontrol} & 14,548 tasks, 833 apps & none (offline) & demonstrated actions & \noRole & \noRole & \noRole \\
WebVoyager~\cite{webvoyager} & 643 tasks, 15 live sites & none (live) & VLM judge & \cYes & \cYes & \noRole \\
WebArena~\cite{webarena} & 812 tasks, 5 sites & self-hosted sites & per-task check & \noRole & \cYes & \noRole \\
VisualWebArena~\cite{visualwebarena} & visual tasks, 3 sites & self-hosted sites & per-task check & \noRole & \cYes & \noRole \\
\midrule
\rowcolor{white}\multicolumn{7}{l}{\;\,\ding{60} \textit{Embodied manipulation}} \\
LIBERO(-PRO)~\cite{libero,liberopro} & 130 tasks, perturbed variants & simulator & BDDL predicate & \cYes & \cYes & \cYes \\
ManiSkill3~\cite{tao2024maniskill3} & GPU-parallel tasks & GPU simulator & success predicate & \cYes & \noRole & \cYes \\
RoboCasa~\cite{robocasa} & 100 kitchen tasks & simulator & success predicate & \cYes & \noRole & \cYes \\
RoboTwin 2.0~\cite{robotwin2} & 50 dual-arm tasks & simulator & success check & \cYes & \noRole & \cYes \\
RoboLab~\cite{robolab} & 120 tasks & simulator & success predicate & \cYes & \noRole & \cYes \\
RoboDojo~\cite{robodojo} & 42 simulated, 18 real tasks & simulator, real reset & success check & \cYes & \noRole & \cYes \\
VLABench~\cite{vlabench} & 100 language-conditioned categories & simulator & binary success & \cYes & \noRole & \cYes \\
BEHAVIOR-1K~\cite{behavior1k} & 1,000 activities, 50 scenes & simulator & BDDL predicate & \cYes & \noRole & \cYes \\
DISCOVERSE~\cite{discoverse} & long-horizon tasks & simulator & task reward & \cYes & \noRole & \cYes \\
VIMA-Bench~\cite{vima} & generated multimodal prompts & simulator & success predicate & \cYes & \noRole & \cYes \\
Hardware task libraries & per-paper tasks & manual or self-reset & none shipped & \cYes & \noRole & \noRole \\
\bottomrule
\end{tabularx}
\end{table*}

This section reviews how self-evolving multimodal agents are evaluated on task-solving capability, on self-evolving capability, and through the judges and reward models that supply their feedback, and identifies the measurements whose absence leaves our account untestable.

\subsection{Task-Solving Capability}\label{sec:eval_tasks}

Task-solving capability, the expected return $\mathcal{J}_k$ (\cref{eq:eval_seq}) that an agent attains on the goals of a scenario with its components fixed, is measured on task-specific benchmarks that fix a goal set, any environment that rollouts need, and usually a scorer (\cref{tab:eval_tasks}).
A reported score estimates $\mathcal{J}_k$ only insofar as the goal set represents $\mu_{\mathcal{G}}$ and the scorer agrees with $R$, and audits find both conditions broken.
The goal set fails through memorization, Sphinx-X-MoE reaching 43.6\% on MMMU~\cite{mmmu} without images, 17.9 points above its language-model backbone~\cite{mmstar} and vision-language-action models above 90\% on LIBERO~\cite{libero} falling to 0.0\% under the perturbations of LIBERO-PRO~\cite{liberopro}.
The scorer fails through divergence and exploits, GenEval~\cite{geneval} deviating from human judgment by up to 17.7 points~\cite{geneval2} and BenchJack~\cite{benchjack} reaching near-perfect scores on most of ten agent benchmarks without solving a task.
A loop that trains with the scorer it is evaluated by, as Reflection-VLA~\cite{reflectionvla} does on LIBERO, can write memorized goals and scorer errors into the agent, a self-evolving agent thus being more exposed to these failures than the static models and agents that the audits examine.

\subsection{Self-Evolving Capability}\label{sec:eval_benchmarks}

Self-evolving capability, the ability to raise $\mathcal{J}_k$ over the cycles of a loop, is measured by self-evolution benchmarks that fix the goals, the environment, any feedback, and the schedule in which the agent attempts and is tested on goals (\cref{tab:eval_benchmarks}), whereas loops on task-specific benchmarks report $\mathcal{J}_k$ by round or only at the endpoints.
Text-only benchmarks apply the same protocol to memory, lifelong-learning, and harness-evolution streams and to professional workflows~\cite{evomemory,evomembench,continualagentmemory,lifelongagentbench,stulife,agentcl,pathbench,selfevocl,evoharnessbench,evobenchharness,harnessbenefit,evoagentbench,seaeval,finevobench,finevolvebench,gdpevo,beyondendpoint}, whereas benchmarks in which the agent remembers supplied histories~\cite{memoryagentbench,worldmemarena,longmemevalv2}, learns from supplied demonstrations or labeled streams~\cite{cgl,continualgui}, or is scored without updating itself~\cite{agenticmme} fall outside this measurement, leaving no update that follows from the agent's own outcomes.
Independent scores across rounds trace the increment $\Delta\mathcal{J}=\mathcal{J}_{k+1}-\mathcal{J}_k$, cumulative scores mix it with resampling, endpoint comparisons give only the total gain, and re-tests of earlier goals and held-out goals bear on retention $S_\zeta$ and generalization $G_\zeta$.
A gain counts as evidence of the capability only to the extent that controls separate the agent's experience from added computation, repeated attempts, writing itself, and run variance, as in MemGUI-Bench~\cite{memguibench}, where retries alone raise M3A by 14.9 points while removing the long-term memory of Agent-S2 at the same three attempts costs it 20.0 points.
Whether the check reaches the agent also shapes the gain, conditions that withhold it lowering gains or leaving them mixed~\cite{mmg2skillbench,gameboyworlds}, with late-round successes in PhysRSI-eval falling from 15 to 5 of 36 without the grader~\cite{physrsieval}, whereas the protocols marked $^{\dagger}$ return the check, a gain under them possibly including its exploitation, as exposing a simulator's success criteria to embodied agents yields gains that vanish once the criteria change slightly~\cite{criteriaexposure}.
These benchmarks measure mechanisms only on goals whose outcomes are checked, away from the goals where \cref{tab:cs_pairs} sets verdicts by missing checks or unobservable outcomes, and cover navigation and manipulation, on hardware only in SEAR's five tasks~\cite{searrobot}, with reasoning and generation covered only by Flex-RSI~\cite{flexrsi}, which omits navigation.

\subsection{Judges and Reward Models}\label{sec:eval_judges}

Where no answer key or program checks an output, a self-evolving agent realizes $V_\nu$ through a judge or a reward model, whose preferences $f_\zeta$ writes into $\mathcal{A}_{k+1}$ while the loop observes no $R$, moving the policy toward outputs that the scorer overrates.
A scorer that shares the updated parameters or is refit moves with the policy as well.
Evaluations compare these scorers with a reference outside the loop (\cref{tab:ch_judges}) and bear on the realized fidelity $F_\nu^{(k)}$ only insofar as their outputs and scorer resemble those in the loop.
On fixed outputs, agreement stays moderate and falls off the training distribution, the strongest judges reaching 62.9\%--72\% accuracy~\cite{vlrewardbench,mmrewardbench,vurb,agentrewardbench}, reward models dropping to near chance on pairs outside their training data~\cite{uigenie,androidflux}, and a log that reports success misleading judges on failed clips~\cite{dontreadlog}.
Optimization exposes further errors, reward hacking reaching 48.1\% for VQA policies~\cite{mmrewardhacking} and a judge passing repairs that humans mostly reject~\cite{dontreadlog}, although a fixed screen judge stays accurate on the policy trained against it~\cite{acurl} and fixed-output accuracy largely orders scorers by the downstream results of policies optimized against them~\cite{mobilejudgebench,mjbench}.
References that check outcomes reach open-ended outputs only through single-property interventions~\cite{isovrbench}, leaving such outputs scored against human labels, larger models, or other reward models~\cite{mmrewardhacking,tdpo,t2irewardhacking}.

\subsection{Measurement Gaps}\label{sec:eval_gaps}

We list a measurement as missing only where its absence leaves one of three claims untestable, that a method's gain comes from self-evolution, that a scenario sits where we place it on the positioning matrix, and that the portability verdicts hold, the last two specific to our account.
These measurements share the need for a reference the loop never receives, a reference returned to the agent entering its feedback $V_\nu$ and becoming exploitable by its updates.

\pgraph{Attribution of Gains}\; Attributing a gain to self-evolution requires, in every scenario, the increment across rounds against a non-evolving baseline at a matched budget and, where the scorer that checks updates also produces the reported score, a reference withheld from the loop.
Four self-evolution benchmarks run this baseline (\cref{tab:eval_benchmarks}), Flex-RSI among them with harness edits raising the test score in six of seven paired runs~\cite{flexrsi}, as does RATs~\cite{rats}, whereas the other comparisons reported in reasoning and generation, such as V-Zero against supervised GRPO~\cite{vzero}, state no matched budget.
On hardware the budget also counts the human resets that $\mathcal{C}_1$ admits, which Find Something You Can't Do records as scene recoveries~\cite{findcantdo}, lest an evolving arm draw on more of them than its baseline.
The withheld reference, such as the test split that Flex-RSI scores once after each run~\cite{flexrsi}, separates the gain from exploitation of the scorer, as for the protocols marked $^{\dagger}$, and, where training and reported goals coincide, from memorization.
Agreement on fixed outputs ranks scorers without bounding that exploitation, and the audits in \cref{tab:ch_judges} cover parameter updates and single-output repair, leaving exploitation through memory and tool writes untested.

\begin{table*}[!t]
\centering
\scriptsize
\renewcommand{\arraystretch}{1}
\setlength{\tabcolsep}{1.5pt}
\providecommand{\noRole}{\textcolor{gray!45}{--}}
\caption{\textbf{Self-evolution benchmarks.}
Notation: $^{\ast}$ external experience supplied; (req.) no non-pixel substitute; $^{\dagger}$ reported scorer; P parameter, M memory, T tool, code, or skill, and I instruction or harness write, \emph{any} open choice, parentheses the reference agents' writes; \cYes{} per-round scores, re-tests, or a matched baseline under $\Delta\mathcal{J}$, $S_\zeta$, and Budget, \cPart{} cumulative or endpoint scores or partial matching; $G_\zeta$ farthest held-out goals.
}
\label{tab:eval_benchmarks}
\vspace{-2mm}
\rowcolors{3}{themelight}{white}
\renewcommand{\tabularxcolumn}[1]{m{#1}}
\begin{tabularx}{\textwidth}{%
@{\hspace{2pt}}>{\raggedright\arraybackslash}m{2.85cm}
>{\raggedright\arraybackslash}X
>{\raggedright\arraybackslash}m{1.75cm}
>{\raggedright\arraybackslash}m{1.55cm}
>{\raggedright\arraybackslash}m{1.40cm}
>{\raggedright\arraybackslash}m{1.85cm}
>{\centering\arraybackslash}m{0.56cm}
>{\centering\arraybackslash}m{0.37cm}
>{\raggedright\arraybackslash}m{1.45cm}
>{\raggedright\arraybackslash}m{1.75cm}
>{\centering\arraybackslash}m{0.84cm}@{\hspace{2pt}}}
\toprule
\rowcolor{gray!25}
\textbf{Benchmark} & \textbf{Tasks} & \textbf{Observation} & \textbf{Feedback} & \textbf{Writes} & \textbf{Schedule} & $\boldsymbol{\Delta\mathcal J}$ & $\boldsymbol{S_\zeta}$ & $\boldsymbol{G_\zeta}$ & \textbf{Cost} & \textbf{Budget} \\
\midrule
\rowcolor{white}\multicolumn{11}{l}{\;\,\ding{60} \textit{GUI and web navigation}} \\
MemGUI-Bench \cite{memguibench} & 128 tasks, 26 Android apps & screenshots or UI tree & LLM judge \rl{B}$^{\dagger}$ & any (M, T) & retry, 3 attempts & \cPart & \noRole & \noRole & tokens, time, API cost & \noRole \\
MMG2Skill-Bench$^{\ast}$ \cite{mmg2skillbench} & 130 tasks, desktop and games & screenshots (req.), text & self-diagnosis \rl{A} & T & retry, 5 attempts & \cYes & \noRole & \noRole & model calls, tokens & \cYes \\
\midrule
\rowcolor{white}\multicolumn{11}{l}{\;\,\ding{60} \textit{Embodied manipulation}} \\
PhysRSI-eval \cite{physrsieval} & 18 tasks, RoboDojo & images, state & RoboDojo score \rl{B}$^{\dagger}$ & M, T & retry, $\le$17 rounds & \cPart & \noRole & layouts & physical steps & \cYes \\
RobotAgentBench \cite{robotagentbench} & 116 tasks, Isaac Sim & images, state, map & execution feedback & T & iterations, 10 rounds & \cYes & \noRole & \noRole & \noRole & \noRole \\
SEAR \cite{searrobot} & 330 simulated, 5 hardware tasks & images or vision tools & sealed judge \rl{B}$^{\dagger}$ & P, M, T & task stream & \cPart & \noRole & new simulator & tokens, env steps, GPU time & \cYes \\
\midrule
\rowcolor{white}\multicolumn{11}{l}{\;\,\ding{60} \textit{Across reasoning, generation, and manipulation}} \\
Flex-RSI~\cite{flexrsi} & 24 tasks: robot simulation, video and spatial QA, image and video generation & images, video, simulator state & development labels \rl{B}, withheld validation gate & I, T & rounds under a fixed budget & \cPart & \noRole & hidden test split & tokens & \cYes \\
\midrule
\rowcolor{white}\multicolumn{11}{l}{\;\,\ding{60} \textit{Outside the four scenarios}} \\
PokeGym \cite{pokegym} & 114 tasks, Pok\'emon Z-A & frames (req.) & composite score \rl{B}+\rl{E}$^{\dagger}$ & any (T, I) & retry, 10 rounds & \cYes & \noRole & \noRole & tokens, API cost & \noRole \\
GameVerse$^{\ast}$ \cite{gameverse} & 15 games & video, frames (req.) & failure video \rl{E} & I & retry, 2 attempts & \cPart & \noRole & \noRole & \noRole & \noRole \\
OmniGameArena \cite{omnigamearena} & 2 of 12 UE5 games & frames (req.) & progress score \rl{B}$^{\dagger}$ & M, I & iterations, 10 rounds & \cYes & \noRole & variants & \noRole & \noRole \\
SelfImprove-Bench \cite{selfimprovebench} & 6 games & frames (req.) in 2, text in 4 & score, success \rl{E}$^{\dagger}$ & any (M, T, I) & task stream, replays & \cPart & \cYes & other games & \noRole & \noRole \\
EmbodiedCLBench \cite{embodiedclbench} & 268 tasks, offline state graphs & images, actions as text & unstated & M, T & source to target tasks & \noRole & \noRole & compositions & \noRole & \noRole \\
GameBoyWorlds \cite{gameboyworlds} & 500 tasks, 10 games & frames (req.) & intrinsic or self-judged \rl{A} & any (P, M, I) & source to target games & \noRole & \noRole & new games & \noRole & \noRole \\
MC-EvoBench \cite{mcevobench} & 80 tasks, Minecraft & frames, planner state & stage verifiers \rl{B}$^{\dagger}$ & M & source to reset target & \noRole & \noRole & related tasks & model calls, env steps & \cPart \\
\bottomrule
\end{tabularx}
\end{table*}

\pgraph{Scenario Placement}\; Placing a scenario depends only on whether each ceiling is low or high, and a misplacement would merely move verdicts into or out of the redundant class.
Rollout cost in a common unit would test the split on $\bar{K}$, yet the reported units do not convert, ranging from tokens in MemGUI-Bench to hours of robot time in Self-Improving EFMs~\cite{siefm}.
The split on $\bar{F}$ remains inferred, the evaluations of \cref{tab:ch_judges} defining each reference within one scenario and values taken from fixed outputs erring in either direction.
A higher $\bar{F}$ than we place in generation or on hardware would make the feedback mechanisms there redundant, and for generation the references on optimized outputs reduce to human labels in a simulated repair loop~\cite{dontreadlog} and, under retraining, other reward models~\cite{selfverifiercapture}.

\pgraph{Verdict Testing}\; Testing a verdict of \cref{tab:cs_pairs} means running the matched-budget comparison in the target scenario with the mechanism moved unadapted, on goals outside the self-evolution benchmarks wherever the verdict turns on missing checks or unobservable outcomes.
A conditional verdict also needs the adapted arm, and a blocked verdict an outcome check held outside the loop, one available to the loop making the outcome observable.
Where a loop gives up a resource, the verdict rests on realized $F_\nu$, whose test needs the policy optimized against a scorer that changes with it, a joint condition that \cref{tab:ch_judges} checks against a hidden key only in text~\cite{selfplayjudge}.
The reported transfers fall short of such a test, ASG~\cite{asg} bringing voting to generation without an unadapted arm, failure-conditioned curricula moving from the text-only WebRL~\cite{webrl} to screen-based loops~\cite{seagent,acurl} only within navigation, and SEAR starting its hardware tasks from an empty library instead of the one written in simulation.
The verdicts thus remain untested under a shared protocol.

\begin{table*}[!t]
\centering
\scriptsize
\renewcommand{\arraystretch}{1}
\setlength{\tabcolsep}{3.7pt}
\providecommand{\noRole}{\textcolor{gray!45}{--}}
\caption{\textbf{Evaluations of multimodal judges and reward models.}
Fixed, Optimized, and Retrained mark scoring of fixed outputs, of outputs trained or selected against the scorer, and across retraining of the scorer.
Notation: $^{\ddagger}$ outcome-checking reference; \cPart{} downstream results only.
}
\label{tab:ch_judges}
\vspace{-2mm}
\rowcolors{3}{themelight}{white}
\renewcommand{\tabularxcolumn}[1]{m{#1}}
\begin{tabularx}{\textwidth}{%
>{\raggedright\arraybackslash}m{3.73cm}
>{\raggedright\arraybackslash}m{3.16cm}
>{\raggedright\arraybackslash}m{2.92cm}
>{\raggedright\arraybackslash}X
>{\centering\arraybackslash}m{0.70cm}
>{\centering\arraybackslash}m{1.24cm}
>{\centering\arraybackslash}m{1.14cm}}
\toprule
\rowcolor{gray!25}
\textbf{Evaluation} & \textbf{Scorer} & \textbf{Judged unit} & \textbf{Reference} & \textbf{Fixed} & \textbf{Optimized} & \textbf{Retrained} \\
\midrule
\rowcolor{white}\multicolumn{7}{l}{\;\,\ding{60} \textit{Multimodal reasoning}} \\
MLLM-as-a-Judge~\cite{mllmjudge} & MLLM judges & vision-language responses & human scores and rankings & \cYes & \noRole & \noRole \\
VL-RewardBench~\cite{vlrewardbench} & VLM judges, reward models & vision-language responses & human-verified pairs & \cYes & \cPart & \noRole \\
Multimodal RewardBench~\cite{mmrewardbench} & VLM judges & vision-language responses & expert-labeled triplets & \cYes & \noRole & \noRole \\
VisualProcessBench~\cite{visualprm} & PRMs, MLLMs & reasoning steps & human step-level labels & \cYes & \cPart & \noRole \\
VURB~\cite{vurb} & VLM judges, reward models & long-trace video answers & answer correctness$^{\ddagger}$ & \cYes & \noRole & \noRole \\
Multimodal reward hacking~\cite{mmrewardhacking} & rule and judge rewards & VQA answers & 235B VLM oracle & \noRole & \cYes & \noRole \\
Self-improving VLM judge~\cite{selfimprovevlmjudge} & self-trained VLM judge & benchmark responses & benchmark preference labels & \cYes & \noRole & \cYes \\
\midrule
\rowcolor{white}\multicolumn{7}{l}{\;\,\ding{60} \textit{Multimodal generation}} \\
MJ-Bench~\cite{mjbench} & CLIP-based scorers, VLMs & text-to-image outputs & human pairs and rankings & \cYes & \cPart & \noRole \\
IsoVRBench~\cite{isovrbench} & video reward models & generated videos & single-property interventions$^{\ddagger}$ & \cYes & \cPart & \noRole \\
TDPO~\cite{tdpo} & reward models & text-to-image outputs & held-out reward models & \noRole & \cYes & \noRole \\
T2I reward hacking~\cite{t2irewardhacking} & preference, detector, MLLM & text-to-image outputs & artifact labels, held-out metrics & \cYes & \cYes & \noRole \\
Don't Read the Log~\cite{dontreadlog} & MLLM judges & generated video clips & human labels on contact sheets & \cYes & \cYes & \noRole \\
Self-Verifier Capture~\cite{selfverifiercapture} & the policy's model & generated images & reward-model panel & \cYes & \cYes & \cYes \\
\midrule
\rowcolor{white}\multicolumn{7}{l}{\;\,\ding{60} \textit{GUI and web navigation}} \\
AgentRewardBench~\cite{agentrewardbench} & LLM and VLM judges & web trajectories & expert success annotations & \cYes & \noRole & \noRole \\
Mobile judge benchmark~\cite{mobilejudgebench} & VLM judges & mobile trajectories & human labels, task checker$^{\ddagger}$ & \cYes & \cPart & \noRole \\
AndroidFlux~\cite{androidflux} & reward models, likelihoods & recovery actions & agent-panel rollout success$^{\ddagger}$ & \cYes & \noRole & \noRole \\
ACuRL judge audit~\cite{acurl} & VLM judge & computer-use trajectories & rule checker$^{\ddagger}$, human labels & \cYes & \cYes & \noRole \\
\midrule
\rowcolor{white}\multicolumn{7}{l}{\;\,\ding{60} \textit{Embodied manipulation}} \\
FailBench~\cite{failbench} & VLM judges & robot attempts & source success labels & \cYes & \noRole & \noRole \\
SOLE-R1~\cite{solerl} & progress reward models & per-step progress & task success, frame order$^{\ddagger}$ & \cYes & \cYes & \noRole \\
\midrule
\rowcolor{white}\multicolumn{7}{l}{\;\,\ding{60} \textit{Text only, for comparison}} \\
Self-play judge hacking~\cite{selfplayjudge} & the policy's model & math answers & hidden exact-match key$^{\ddagger}$ & \noRole & \cYes & \cYes \\
Judge-hacking pace~\cite{judgehackingpace} & frozen LLM judges & math answers & answer key$^{\ddagger}$ & \cYes & \cYes & \noRole \\
Self-Rewarding LMs~\cite{yuan2024self} & the policy's model & chat responses & human rankings & \cYes & \cPart & \cYes \\
\bottomrule
\end{tabularx}
\end{table*}

\section{Open Challenges}
\label{sec:open_challenges}

This section discusses four open problems: reference construction, self-derived verification, and cross-substrate transfer, which inform our account of the cross-scenario asymmetry, along with safety under autonomy, which bears on the operational definition of self-evolution.

\subsection{Reference Construction}\label{sec:ch_reference}

Reference construction, the building of a substitute reference for an outcome absent from both the agent's observations and every check available to the loop, decides whether the blocked verdicts can be lifted, as for background-state goals in navigation and possibly contact-rich assembly on hardware (\cref{sec:cs_transfer}), and the four reported routes each succeed only in part.
Converting outputs into answers~\cite{asg} needs an equivalence relation that screens and action trajectories lack, and a program written together with the goal covers only goal families with a closed-form success condition, EvoCUA resolving by hand the synthesized tasks on which its two checks disagree~\cite{evocua}.
A learned world model adds its own prediction error, its dynamics learned on hardware only over short horizons~\cite{worldvlaloop} and in navigation only over text-rendered states~\cite{dynaweb}, while a vision-language judge reads camera frames that do not record contact forces~\cite{failbench}.
Whether a loop can construct such a reference without a human inside the loop thus remains open, an affirmative answer turning the blocked verdicts into conditional ones that depend on how the reference is built and a negative one making the block a property of the outcome itself.

\subsection{Self-Derived Verification}\label{sec:ch_verification}

Self-derived verification, the scoring of outcomes or steps by a verifier derived from the agent's own model or trajectories, is adopted wherever no check scores an outcome or a check returns one verdict per long episode, and its reliability turns on how far the verifier depends on the policy.
Verifiers that change with the policy drift or inherit its errors, confidence rewards collapsing in ablated configurations of SOLACE~\cite{solace} and IRIS~\cite{iris} and step reward models trained on the agent's trajectories inheriting its mistakes~\cite{uigenie,magicguirms}.
Lagged, co-trained, and frozen verifiers, such as the moving-average teacher of UniEvo-VL~\cite{unievo}, the initial model in PRISM~\cite{prism}, and progress heads frozen before reinforcement learning~\cite{siefm,scvla}, depend on the policy less or not at all, trading a moving bias for one fixed by their source, a trade that no report compares across degrees of dependence.
Whether any degree of dependence keeps a verifier consistent with $R$ once the policy leaves the support on which the verifier was formed remains open, an affirmative answer making reliability a choice of dependence and supplying step-level credit to long episodes, and a negative one turning the low $\bar{F}$ of generation and hardware and the scarcity of $L_\nu$ in navigation and manipulation into structural limits.

\subsection{Cross-Substrate Transfer}\label{sec:ch_transfer}

Cross-substrate transfer, the carrying of what a loop writes, whether a skill library, a harness, or a policy, to another reader, platform, or embodiment once the missing resources are supplied, underlies the conditional verdicts for hardware targets, and the reported evidence leaves it unsettled.
Across platforms, zero-shot transfer is reported only for the fixed loop GUI-HARVEST, whose harness evolved on OSWorld raises GPT-5 by 13.87 points on WindowsAgentArena~\cite{guiharvest}.
Across simulators and action interfaces, the code policies of SEAR lose little while its VLA policies lose most of their success, the former falling from 33.3\% to 28.7\% and the latter from 43.1\% to 10.2\%~\cite{searrobot}.
Across readers, the same WebCoach memory lowers a 7B agent and raises larger ones~\cite{webcoach}, in line with the writer dependence of \cref{sec:cs_costs}.
No report carries a write across the simulation-to-hardware boundary, SEAR starting its hardware tasks from an empty library (\cref{sec:eval_gaps}).
Whether a write takes effect across readers, platforms, and embodiments, the simulation-to-hardware boundary included, without new experience accordingly remains open, an affirmative answer loosening the conditional verdicts for hardware targets and a negative one making the substrate itself a missing resource there.

\subsection{Safety under Autonomy}\label{sec:ch_safety}

Autonomy ($\mathcal{C}_1$) excludes human veto from the loop, leaving the loop itself to preserve safety across modifications, as the first of the three laws proposed for self-evolving agents requires~\cite{fang2025comprehensive}, and the reported evidence shows safety eroding along each path of self-modification examined.
Degradation is reported along the model, memory, tool, and workflow paths, on text and code except for one screen-based case, SEAgent, whose unsafe completion rate on RiOSWorld rises from 39.5\% to 55.3\%~\cite{shao2026misevolve}.
Experience accumulated only on benign tasks in text-rendered web and embodied environments raises attack success on later high-risk tasks~\cite{zhao2026safetyrisks}, and safety-oriented loops such as FailBank~\cite{cui2026failbank} draw their proposals from a teacher fixed before the loop, leaving untouched the loops in which the agent supplies its own feedback.
Whether a loop can preserve a safety property across cycles while no human input reaches its goals, feedback, updates, or actions remains open, an affirmative answer showing autonomy and safety need not trade off under $\mathcal{C}_1$ and a negative one making the trade-off a consequence of the definition, common to all scenarios.

\section{Conclusion}
\label{sec:conclusion}

\pgraph{Summary}
This survey investigates the structural source of cross-scenario asymmetry in mechanism portability across four scenarios. 
Through the three-stage self-evolving loop, we place the four scenarios in a positioning matrix spanned by ceilings on rollout count and feedback fidelity, and use fixed loops to separate scenario-imposed limits from those incurred by self-evolving design.
We then recast each mechanism as a trade-off that raises a scarce factor by presupposing a task-structural resource, and attribute the asymmetry to the different combinations of scarce factors and resources that the two directions of a scenario pair encounter.
Inferred from single-scenario evidence, the resulting portability verdicts are testable preconditions, not empirical guarantees.
\cref{sec:evaluation} identifies the measurements that would test them, and \cref{sec:open_challenges} discusses the problems that existing mechanisms leave open.

\pgraph{Limitations}
Although this survey characterizes the cross-scenario asymmetry through an explicit definition, framework, and set of verdict rules, it has the following limitations.
First, inclusion requires the agent to realize at least one of the goal sampler, the feedback function, or the evolution operator, a criterion that confines the scenario patterns to the self-evolving subset and admits fixed loops only as controls. The provenance assignments and the grouping axes of the scenario taxonomies were made by a single coder, the axes after the binding constraints had been identified, and the consolidation profiles partly reflect which scenarios report retention and generalization at all.
Second, the quadrant placement rests on single-scenario estimates of rollout cost and feedback fidelity that do not separate the pretraining coverage of the backbone from the task structure, and the verdicts are derived from rules, not observed in transfer experiments.
Third, the three-stage decomposition is one of several possible stage boundaries, and credit assignment, treated here as a refinement of feedback acquisition, could instead form a separate stage.
Finally, the surveyed methods center on image, video, screen, and robot-camera inputs, leaving loops with audio, tactile, or force feedback unexamined, and later work, particularly in embodied manipulation, may revise the placements and verdicts.

\bibliographystyle{IEEEtran}
\bibliography{ref}

\end{document}